\documentclass[journal]{IEEEtran}
\usepackage{amsthm}
\usepackage{amsmath}
\usepackage{amssymb}
\usepackage{graphicx}
\usepackage{esint}
\usepackage{algorithmic}
\usepackage{algorithm}
\usepackage{comment}
\usepackage{subfigure}
\usepackage{subfig}
\usepackage{mathtools}
\usepackage{float}
\usepackage{wrapfig}
\usepackage{amsfonts}
\usepackage{bbm}
\usepackage{cite}
\usepackage{multirow}
\usepackage{lipsum}
\usepackage{tabularx}
\usepackage{longtable}
\usepackage{threeparttable}
\usepackage{booktabs}
\usepackage{graphics,color}

\usepackage{epstopdf}

\usepackage{color}
\usepackage{lscape}

\usepackage{caption}
\usepackage{multicol}
\usepackage{xcolor}
\usepackage{xpatch}

\theoremstyle{definition}
\newtheorem{theorem}{\textbf{Theorem}}
\newtheorem*{mytheorem}{\textbf{\textit{Theorem}}}
\newtheorem{corollary}{\textbf{Corollary}}
\newtheorem{definition}{\textbf{Definition}}
\newtheorem{lemma}{\textbf{Lemma}}

\newtheorem{rem}{\textbf{Remark}}
\begin{document}
%
\title{Incremental Online Learning of Randomized Neural Network with Forward Regularization}
\author{Junda Wang, Minghui Hu, Ning Li, Abdulaziz Al-Ali, Ponnuthurai Nagaratnam Suganthan,~\IEEEmembership{Fellow,~IEEE}
\thanks{This work was supported by the National Natural Science Foundation of China under Grant 62273230 and 62203302, the State Scholarship Fund of China Scholarship Council under Grant 202206230182. \textit{(Corresponding author: Ning Li.)}}
\thanks{J. D. Wang and N. Li are with the Key Laboratory of System Control and Information Processing, Ministry of Education of China, with the Department of Automation, Shanghai Jiao Tong University, Shanghai 200240, China, and also with the Shanghai Engineering Research Center of Intelligent Control and Management, Shanghai 200240, China (e-mail: $\{$jundawang, ning$\_$li$\}$@sjtu.edu.cn).}

\thanks{Minghui Hu is with the School of Electrical and Electronic, Nanyang Technological University (email: e200008@e.ntu.edu.sg).}
\thanks{A. Al-Ali and P. N. Suganthan are with the KINDI  Computing Research Center, College of Engineering, Qatar University (email: $\{$a.alali, p.n.suganthan$\}$ @qu.edu.qa).}}

%

\markboth{}%
{Shell \MakeLowercase{\textit{et al.}}: Bare Demo of IEEEtran.cls for Journals}
%


\maketitle

\begin{abstract}
Online learning of deep neural networks suffers from challenges such as hysteretic non-incremental updating, increasing memory usage, past retrospective retraining, and catastrophic forgetting. To alleviate these drawbacks and achieve progressive immediate decision-making, we propose a novel \textit{Incremental Online Learning} (IOL) process of Randomized Neural Networks (Randomized NN), a framework facilitating continuous improvements to Randomized NN performance in restrictive online scenarios. Within the framework, we further introduce IOL \textit{with ridge regularization} (-R) and IOL \textit{with forward regularization} (-F). -R generates stepwise incremental updates without retrospective retraining and avoids catastrophic forgetting. Moreover, we substituted -R with -F as it enhanced precognition learning ability using semi-supervision and realized better online regrets to offline global experts compared to -R during IOL. The algorithms of IOL for Randomized NN with -R/-F on non-stationary batch stream were derived respectively, featuring recursive weight updates and variable learning rates. Additionally, we conducted a detailed analysis and theoretically derived relative cumulative regret bounds of the Randomized NN learners with -R/-F in IOL under adversarial assumptions using a novel methodology and presented several corollaries, from which we observed the superiority on online learning acceleration and regret bounds of employing -F in IOL. Finally, our proposed methods were rigorously examined across regression and classification tasks on diverse datasets, which distinctly validated the efficacy of IOL frameworks of Randomized NN and the advantages of forward regularization.
 


\end{abstract}


\IEEEpeerreviewmaketitle

\section{Introduction}

\IEEEPARstart{O}{nline} learning paradigm significantly enables continual updates of neural networks, thereby progressively improving the completion quality of model training on online tasks. The approaches have gained popularity in scenarios that require immediate learning, particularly where data is available only as non-stationary sequential streams. Key applications of this paradigm include live recommendation systems, financial transaction processing, and resource dispatch systems. Compared to an offline expert scheme, the online learner is expected to promptly deliver competitive real-time decision-making behaviors and advance performance under online conditional limitations.\\
\indent Despite remarkable successes in difficult tasks, deep neural network (DNN) is still faced with certain \textit{challenges} when transferring to online environments \cite{1, 3, a0, a1, a2}: (1) \emph{Non-convex optimization.} $T$-time iterative gradient descent (GD) is commonly employed to train DNN due to the non-convexity of loss, which results in at least $\mathcal{O}(T \cdot {f_{DNN}})$ time complexity to converge to an approximate optimum on current batch data if possible. Besides, to respond to immediate requests, this toilsome process needs to be completed rapidly well in advance of upcoming data, and hurdles such as DNN volume and intricate parameter settings further exacerbate the above difficulty. (2) \emph{Resource consumption.} Continuous online non-convex optimization usually demands increasing large computational memory to support retrievals of past accumulative data, and this precondition becomes even more fragile in the actual contexts of huge information and when privacy protection is a necessity. (3) \emph{Catastrophic forgetting and distribution drifting.} Even if the entire task distribution remains consistent, previous knowledge erosion frequently occurs when the network focuses on new batch, and data distribution unpredictably drifts in the learning process of new observations \cite{29,30}. The two dilemmas provoke catastrophic forgetting of what has been learned and performance degradation, which demands a trade-off between plasticity and stability during online learning. Retrospective retraining is a commonly sought solution to preserve generalizations. 

Several recent works have been proposed to alleviate the issues mentioned above. Regarding to the first issue, a faster spatio-temporal GD was applied in the back-propagation of online DNN\cite{1}. Improvements involved accelerating the optimization process of recurrent neural network in online regression using first-order algorithm, with over two times shorter training span \cite{3}. However, time complexity of update step was at least of the same order as stochastic GD, still causing much delays in training and online responses, particularly when utilizing deep schemes. For the rest, multiple online recurrent nets were trained and the best candidate was selected, adapting to distribution drifting dynamics in traffic system or employing an extra recall network to generate pseudo rehearsals of past events\cite{2,4,6}. Nonetheless its bulky stacked structures consumed much time and computation resources, resulting in belated responses and labored incremental updates on taxing online tasks, while suffering from forgetting. Furthermore, for online learning, the notion of no constraint on training time and update efficiency is unrealistic, because data streams do not wait until the network accomplishes training before revealing the next batch chunk that requires learning, decision-making, or responses \cite{5}. Therefore, the problem of obtaining an applicable network with incremental updates for accurate immediate decisions under online scenario restraints without revisiting or forgetting the past is still study-worthy.

\begin{figure}[!htbp]
  \centering
  \includegraphics[width=\columnwidth,trim=4mm 8mm 8mm 4mm, clip]{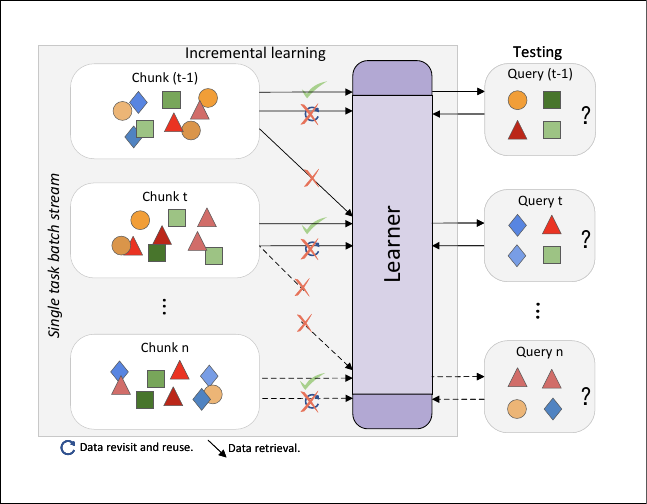}
  \caption{Task stream is learned incrementally. The proposed IOL frameworks investigate IL processes with -R/-F for deep structures under restricted conditions, such as limited retrieval and reuse.}
  \label{fig 00}
\end{figure}

One desirable concept known as incremental learning (IL) allows for: efficient resource utilization by avoiding retraining from scratch when new data arrives; and decreasing memory usage by restricting the amount of data needed to be stored, which is particularly crucial under privacy constraints \cite{30}. IL holds promise to treat the existing challenges, but catastrophic forgetting is still a critical problem in IL \cite{30,33}. Note the IL emphasized in this paper targets online progressive network updates for a single task session, distinct from task-IL and class-IL (e.g. lifelong learning) \cite{31,32}. According to the descriptive metrics in \cite{7,8,9,48}, IL is characterized by the following strict \textit{restrictions}: 
(1) stepping utilization of new data without past retrieval, and progressive relay updating rather than retraining; (2) small memory usage; (3) maintaining a contemporary model, namely the online learner, as the best approximation to target at any time, and generating prompt query responses of growing quality over time; (4) starting from scratch without pretraining and prior knowledge of tasks in strict cases. 
Our novel IL scheme will not only endeavor to adhere to all these restrictions while realizing continuous improvements to inner Randomized neural networks (Randomized NN), but also allow past chunks to be discarded and help online learners (networks) break free from the dilemmas of retrospective retraining and catastrophic forgetting of IL. It is further compatible with rapid recursive updates on multiple learners. We refer to the enhanced IL scheme for existing \textit{challenges} under online task \textit{restrictions} as incremental online learning (IOL) framework. Concept chart is shown in Fig. \ref{fig 00}.

Due to the harsh definition of incrementalism, only a few studies associated integrating the model into a real IOL framework with partial above theoretical criteria satisfied. In \cite{7}, IL on topology stream achieved node recursive updating and instant response based on a self-organizing network. An online classification task was tackled by incrementally solving constrained optimization to avoid retraining costs \cite{9}. These studies were inclined to use conventional methods with limited realization and learning capability rather than deep structures to carry out the IL scheme, and did not delve into theoretical analysis of online processes or learning regrets in IL. In this work, we strive to present an IOL framework that involves promising deep structures for overcoming challenges such as non-incremental updating, retrospective retraining, and catastrophic forgetting in complicated tasks with online restrictions.

As a prevailing Randomized NN, the ensemble deep random vector functional link network (edRVFL) presents a prospective way of contributing to real IOL frameworks of deep structures. This can be attributed to its universal approximation ability, extendibility, and closed-form solution of convex target, as the incrementalism for network updates can be completed by the recursive convertible solutions of ever-changing optimization targets based on Sherman's rule during IOL. For the ridge regularization form, recent works implemented IL of RVFL-based variants on sequential tasks \cite{b0,b1,b2,b3,b4,b5}. However, most did not conduct in-depth theoretical analysis for IL process on batch stream, or consider existing \textit{challenges} under online \textit{restrictions}. Furthermore, to endow the online deep learners with efficient foresight learning from future data, based on the IOL pattern with the extra forward regularization, there is an expectation of using semi-supervised strategy to enhance precognition and accelerate updating course as well as lower online regrets by leveraging coming unlabeled data batch in network online learning practice\cite{10,11,12}. 

We revisit aforesaid challenges and briefly outline the advantages of IOL for edRVFL in addressing them. (1) \emph{Immediate incremental update without retrospective retraining.} Based on differentiable convex targets and Sherman's rule, proposed IOL frameworks with ridge (-R) and forward regularization (-F) achieve non-iterative but recursive one-shot incremental updates for each learner on every batch chunk. This maintains an up-to-date model of rapid responses and increasing performance without retrospective retraining. (2) \emph{Low memory usage.} IOL stores the learned knowledge in weights, updates them incrementally, and allows previous data to be discarded. This frees IOL processes from cumulative data usage. (3) \emph{Forgetting-free and no relative solution drifting.} Passive -R/-F update algorithms within IOL frameworks are innately forgetting-free and have no solution (weight) drifting compared to a synchronous offline expert, which learns on past global data of batch stream (see Sec. III A). Detailed research on -R/-F algorithms integrated with IOL for edRVFL is given in this paper. Moreover, -F designed for regret reduction in IOL processes when current label invisible is advocated to replace -R for its superiority in theoretical and experimental analysis. Contributions are summarized as follows.
\begin{enumerate}
    \item To achieve progressive immediate decision-making, IOL framework incorporating and promoting IL's advantages is proposed to facilitate continuous evolvement of deep Randomized NN under online restrictions. It is expected to fulfill incremental updating without retrospective retraining and avoid catastrophic forgetting on online batch streams.
    \item Considering existing challenges, edRVFL-R/F within IOL are derived with variable learning rates and stepwise recursive weight updates. -F propels advancing learning of the IOL process using semi-supervision.
    \item We conduct in-depth theoretical analysis and derive several corollaries and cumulative regret bounds for edRVFL-R/F within IOL on batch streams, which suggests the superiority on online learning acceleration and lower regrets of employing -F style into IOL process.
    \item Extensive experiments on diverse datasets of regression and classification tasks and ablation tests are conducted to validate the efficacy of proposed edRVFL-R/F within IOL frameworks. Results demonstrate the advantages and lower regrets of using -F style. 
\end{enumerate}





The remainder is organized as follows. Section II and III introduce related works and preliminaries. In Section IV, we derive the IOL frameworks for edRVFL-R/F. Regret bounds are also analyzed. Section V conducts extensive experiments on various tasks. Section VI provides conclusions, and the Supplementary section records additional information.

\section{Related Work}

\textbf{Randomized NN.} Pioneer works on Randomized NN could be found in \cite{13,40}, where a semi-stochastic RVFL comprised two hidden layers with only output weights being trained. Over the decades, comprehensive research has been implemented on Randomized NN, and hatched several representative models: edRVFL\cite{14}, ELM\cite{18}, and BLS \cite{45}. Readers can refer to \cite{44,41,42,43} for the latest updates. Randomness of weights could be interpreted as high dimensional embeddings that disrupted rotation invariance in MLP and thereby enhanced the universal approximation capability of Randomized NN \cite{6,15,16}. Strong scalability and fast training also empowered models to stand out and excel popular DNNs in conventional offline tasks such as image recognition and representative learning \cite{19,34,35,39,37,38}. Although some online learning versions of Randomized NN have emerged recently in \cite{b0,27,46}, they still lacked general theoretical framework derivation, regret analysis, and improvement with extra regularization for online learning process on batch streams. Our work aiming to fill this gap is explained in Section IV.

\textbf{Incremental learning.} IL in this paper represents the dynamic process of the model's continuous evolving with progressing decision-making quality on the online batch stream of a single task. Batches appear over time. To cope with existing \textit{challenges} and \textit{restrictions}, we emphasize the IOL framework, an enhanced IL under restrictive conditions. Note here IL is different from the IL concepts in \cite{47,45}.

Previous literature has studied the following challenges: (1) \textit{Catastrophic forgetting
and distribution drifting} \cite{49,50,51,52,53}; (2) \textit{retrospective retraining} \cite{48,54}; (3) \textit{memory usage} \cite{55}. Though these challenges have been independently studied, their joint therapy has been mostly underexplored. We consider all the above issues in proposed -R/-F algorithms within IOL frameworks and are committed to addressing them, moreover, -F is employed to expedite IOL process.

\textbf{Online regret.} Compared to flourishing variants of Randomized NN, few studies involved in-depth analysis of the IL process or regret. Our work investigates the theoretical patterns of Randomized NN's IOL and its regrets. IOL process can also be regarded as continuous online convex optimization (OCO) with variable gradients to Randomized NN on non-stationary batch streams. The dynamic regret in IOL is defined as the difference between total cost the learner with algorithm $\mathcal{A}$ has incurred and that of the best expert $\mathcal{E}$'s decision in hindsight, which follows $\mathcal{R}_t^\mathcal{A} = \mathcal{L}_t^\mathcal{A} - \mathop {min}\limits_{\theta  \in \Omega } \mathcal{L}_t^\mathcal{E}(\theta )$\cite{21}. The regret analysis for OCO problem can be found in \cite{56,57,58,59,60,61}.
\section{Preliminary}
Given an online non-stationary batch data stream of a single task in sequence $(X,Y) = \{ ({X_t},{Y_t})|{X_t} \in {\mathbb{R}^{b \cdot k}},{Y_t} \in {\mathbb{R}^{b \cdot m}},t = 0,...,T\}\sim \mathcal{X} \times \mathcal{Y}$ distribution, where $({X_t},{Y_t})$ $ = \{ (x_t^i,y_t^i)\} _{i = 1}^b $, the $b$, $k$, $m$ denote number of batch samples, dimension of each sample, and number of classes respectively.
\subsection{Offline edRVFL network}
As an extension of RVFL, edRVFL stacked several RVFL layers vertically, reconnected origin input as residuals, and employed ensemble strategy to boost performance on hard tasks \cite{14}. Given an offline edRVFL scheme with $L$ hidden layers, $N$ nodes per layer, projection ${f_{edRVFL}}:\mathcal{X} \to \mathcal{Y}$, and one batch data $({X_t},{Y_t})$, the extracted feature $H$ of the $1$-st hidden layer is defined as:
\begin{equation} \label{1}
{H_{1,t}} = {g_1}({X_t} \cdot {W_1}).
\end{equation}
For layer $1 < l \leqslant L$ it is defined as:
\begin{equation} \label{2}
{H_{l,t}} = {g_l}([{H_{l - 1,t}}|{X_t}] \cdot {W_l}),
\end{equation}
where $g$ is the activation function (e.g. ReLU, Sigmoid, Swish), innate randomized weights ${W_1} \in {\mathbb{R}^{k \cdot N}}$ and ${W_l} \in {\mathbb{R}^{(k + N) \cdot N}}$, and layer feature ${H_{l,t}} \in {\mathbb{R}^{b \cdot N}}$. Due to the differentiable convex target, offline edRVFL can be trained to the optimum in $L$ snapshots without GD therapy by:
\begin{equation} \label{3}
{\beta _{l,t + 1}} = argmi{n_\beta }{\text{ }}\left\| {[{H_{l,t}}|{X_t}] \cdot \beta  - {Y_t}} \right\|_2^2 + \lambda \left\| \beta  \right\|_2^2,
\end{equation}
and the updated solution for future prediction:
\begin{equation} \label{4}
\begin{split}
primal&: {\beta _{l,t + 1}} = {(D_{l,t}^T{D_{l,t}} + \lambda I)^{ - 1}}D_{l,t}^T{Y_t}\\
dual&: {\beta _{l,t + 1}} = D_{l,t}^T{({D_{l,t}}D_{l,t}^T + \lambda I)^{ - 1}}{Y_t},
\end{split}
\end{equation}
where ${D_{l,t}} = [{H_{l,t}}|{X_t}]$, ${\beta _{l,t+1}} \in {\mathbb{R}^{(k + N) \cdot m}}$ denotes the learnable output weight matrix. Final predictions can be obtained by ensemble strategy, normally by aggregating averaged or median values of sub-learners for regression and the ones after SoftMax for classification task, which can be denoted as $En(\{ {D_{l,t}}{\beta _{l,t + 1}}\} _{l = 1}^L,{Y_t})$. For training, $\mathcal{O}({f_{edRVFL}})$ has minima of $\mathcal{O}(L \cdot {(N + k)^3})$ and $\mathcal{O}(L \cdot {b^3})$ time complexity compared to $\inf \mathcal{O}({f_{MLP}}) = \mathcal{O}({T_{it}} \cdot L \cdot {N^2})$ with ${T_{it}}$ iterations every epoch. However, edRVFL spends much less actual time as MLP needs multistep gradient computations each layer.

Theorems of universal approximation property and convergence analysis of Randomized NN could be found in \cite{100,17,16}. Using ensemble policy made edRVFL more robust and cost-effective than pure deep designs, and outperformed DNNs on challenging tasks. In this paper, we study the prevailing edRVFL with -R/-F algorithms on batch streams and conduct theoretical analysis of IOL framework.

\subsection{Incremental learning (IL) scheme}
Compared to offline learning model (i.e. expert) which makes decisions globally, the online learning model (i.e. learner) updates using stepwise trials of a dynamic stream. The alone learner of linear model works on single sample stream of one task in this subsection, and future data is temporarily invisible to the learner. Narrowing the relative error gap (i.e. regret ${\mathcal{R}_t}$) with experts as much as possible is the goal of learners. To pave our work on batch IOL of Randomized NN, some lemmas in \cite{20} are introduced first, where the task stream is denoted by $\{ (x_t^ \cdot ,y_t^ \cdot )\} _{t = 1}^T \subseteq (X,Y)$. Here superscript $ \cdot $ indicates arbitrary selection. Assume $m = 1$ afterwards for simplicity as classification can be divided into $m$ regressions.

\begin{definition} {\bf Bregman divergence}\label{define 2.1}
is used to measure relative projection distance between hyper-parameter sets or distributions. For a real-valued differentiable convex projection $G:\theta  \in \Theta  \to \mathbb{R}$, Bregman divergence $\Delta $ is defined as:
\begin{equation} \label{5}
{\Delta _G}(\tilde \theta ,\theta ): = G(\tilde \theta ) - G(\theta ) - {(\tilde \theta  - \theta )^T} \cdot {\nabla _\theta }G(\theta ),
\end{equation}
where ${\nabla _\theta }$ denotes the gradient on vector $\theta $.
\end{definition}

The immediate incurred loss on single sample ${x_t}$ (resp. batch $({X_t},{Y_t})$) is denoted by ${\ell _t}(\theta ) = \frac{1}{2}||x_t^T\theta  - {y_t}||_2^2$ (resp. ${L_t}(\theta ) = \frac{1}{2}||{X_t}\theta  - {Y_t}||_2^2$), ${\ell _{1..t}}(\theta ) = \sum\nolimits_{q = 1}^t {{\ell _q}} $ (resp. ${L_{1..t}}(\theta ) = \sum\nolimits_{q = 1}^t {{L_q}} $) is the cumulative loss on $t$  samples (resp. batches), and similarly define the forward predictive loss to ${\hat \ell _{t + 1}}(\theta ) = \frac{1}{2}||x_{t + 1}^T(\theta  - {\theta _0})||_2^2$ (resp. ${\hat L_{t + 1}}(\theta ) = \frac{1}{2}||{X_{t + 1}}(\theta  - {\theta _0})||_2^2$).

\begin{lemma} {\bf Offline learning} \label{lemma 2.1}
refers to the learning process of offline expert on global dataset. Assume ${\ell _t}$ and ${U_0}(\theta )$ are differentiable and convex, subscript $0$ denotes initial setup of prior knowledge, and there always exists a solution in $\Theta $:
\begin{equation} \label{6}
{\theta _{T + 1}} = argmi{n_\theta }{\text{ }}{U_{T + 1}}(\theta ),
\end{equation}
where ${U_{T + 1}}(\theta ) = {\Delta _{{U_0}}}(\theta ,{\theta _0}) + {\ell _{1..T}}(\theta )$, ${\theta _{T + 1}}$ represents the updated model parameter for future predictions after the last $T$-th batch knowledge acquisition completed, and function $U_0$ in Bregman divergence can be a $l_2$ term, ${U_0}(\theta ) = \frac{1}{2}{\theta ^T}\eta _0^{ - 1}\theta $ with symmetric positive definite matrix $\eta _0$ for example.
\end{lemma}
\begin{lemma} {\bf Incremental offline learning} \label{lemma 2.2}
directly transfers the offline pattern to online such as Lemma \ref{lemma 2.1} repeatedly retrains whole after knowing $t$-th single sample (time point). So it requires much retrievals and memory storage for cumulative data, resulting in retrospective retraining. For $0 \leqslant t \leqslant T$, following equations based on single sample stream are optimized:
\begin{align}\label{7}
\theta _{t + 1}^{\{ r,f\} } &= argmi{n_\theta }{\text{ }}U_{t + 1}^{\{ r,f\} }(\theta )\\
U_{t + 1}^r(\theta ) &= {\Delta _{{U_0}}}(\theta ,{\theta _0}) + {\ell _{1..t}}(\theta )\notag \\
U_{t + 1}^f(\theta ) &= {\Delta _{{U_0}}}(\theta ,{\theta _0}) + {\ell _{1..t}}(\theta ) + {{\hat \ell }_{t + 1}}(\theta ),\notag
\end{align}
where superscript ${\mathcal{A}}=\{ r,f\}$ is used to identify algorithm styles of -R or -F, $\hat \ell $ is the estimated loss generated by current $\theta $ on upcoming sample. It shows that forward learning adopts guessing of future examples before being labeled to incur loss and effects learning when updates present parameters. Note $\theta _1^{\{ r,f\} } = argmi{n_\theta }{\text{ }}U_{ 1}^{\{ r,f\} }(\theta ) = \theta _0^{\{ r,f\} }$ when $t = 0$. Pseudo algorithmic flows were posted in \textit{Algorithm 1} and \textit{2} of \cite{21}.
\end{lemma}

\begin{lemma} {\bf IL.} \label{lemma 2.3}
For $0 \leqslant t \leqslant T$, with previous ${\ell _{1..t - 1}}$ concentrated into Bregman divergence $\Delta {U_t}$, it realizes IL processes by stepwise optimizing the following equations at $t$ point and removes past rehearsals ostensibly. Note superscripts signify -R/-F styles and the setup of a single stream still holds.
\begin{equation} \label{8}
\begin{split}
\theta _{t + 1}^r &= argmi{n_\theta }{\text{ }}{\Delta _{U_t^r}}(\theta ,\theta _t^r) + {\ell _t}(\theta )\\
\theta _{t + 1}^f &= argmi{n_\theta }{\text{ }}{\Delta _{U_t^f}}(\theta ,\theta _t^f) + {\ell _t}(\theta ) + {{\hat \ell }_{t + 1}}(\theta ) - {{\hat \ell }_t}(\theta )
\end{split}
\end{equation}

One can refer to subsection 4.1.4 and 5.4 for traditional algorithms in \cite{20}. In previous research, several defects could be: (1) they were situated on plain linear regression of limited extendibility on single sample rather than batch stream, which degraded performance on hard tasks and was inapplicable with deducing novel IOL patterns and batch error bounds; (2) variable learning rate $\eta $ was hard to determine if only taking $\frac{{\partial U_{t + 1}^{\{ r,f\} }}}{{\partial \theta }}$ when ${U_{t + 1}}(\theta )$ forms were complex, and its recursive update policy for IL was not given; (3) no realistic implementation of IL process based on neural networks especially using -F, nor was algorithm flow derived; (4) no consideration of existing challenges and restrictions of online learning. To improve these issues, IOL processes with -R/-F for deep Randomized NN on batch stream are proposed, which is expected to break through the dilemmas of existing challenges and restrictions.

\end{lemma}
\subsection{Regret bounds of IL processes}
\begin{lemma} {\bf Online-to-offline relative regrets} \label{lemma 2.4}
are defined as additional cumulative loss (regrets) of online learner over that of the offline expert, which can quantify the gap to the best expert and be seen as the price of hiding future samples from the learner \cite{21}. Assume universal relative loss bounds hold for any arbitrary sequence. For the IL process based on Lemma \ref{lemma 2.3}, relative cumulative regret bound of online learner using ridge w.r.t offline expert under adversarial setup can be written as:
\begin{equation} \label{9}
\begin{split}
\sum\limits_{t = 1}^T {{\ell _t}(\theta _t^r) - \mathop {\min }\limits_\theta  } (\frac{1}{2}\lambda {(\theta  - {\theta _0})^2} + \sum\limits_{t = 1}^T {{\ell _t}} (\theta )) \\
\leqslant 2Y_m^2kIn(\frac{{TX_m^2}}{\lambda } + 1),
\end{split}
\end{equation}
where ${X_m} = \mathop {\max }\limits_{1 \leqslant t \leqslant T} \{ ||{x_t}|{|_\infty }\} $, ${Y_m} = \mathop {\max }\limits_{1 \leqslant t \leqslant T} \{ |{y_t}|,|x_t^T{\theta _t}|\} $, $\lambda $ is the penalty factor, and prediction satisfies $x_t^T{\theta _t} \in [ - {Y_m},{Y_m}]$ for history value restraints. The relative cumulative regret bound of online learner using forward style w.r.t offline expert under adversarial setup can be written as:
\begin{equation} \label{10}
\begin{split}
\sum\limits_{t = 1}^T {{\ell _t}(\theta _t^f) - \mathop {\min }\limits_\theta  } (\frac{1}{2}\lambda {\theta ^2} + \sum\limits_{t = 1}^T {{\ell _t}} (\theta )) \\
\leqslant \frac{1}{2}Y_m^2kIn(\frac{{TX_m^2}}{\lambda } + 1),
\end{split}
\end{equation}
where ${X_m} = \mathop {\max }\limits_{1 \leqslant t \leqslant T} \{ ||{x_t}|{|_\infty }\} $, ${Y_m} = \mathop {\max }\limits_{1 \leqslant t \leqslant T} \{ |{y_t}|\} $. Here it also retains ${y_t} \in [ - {Y_m},{Y_m}]$. Note the regret bound obtained in adoption of -F is 4 times better than that of using -R, which motivates us to advocate for the replacement of -R with -F in the IL process of any network. Unfortunately, rarely works investigated IOL framework of Randomized NN, especially with -F algorithm, and methods in \cite{20} could not be used to analyze and derive cumulative regret bounds of IOL processes for edRVFL-R/F on batch streams.

\end{lemma}
\section{Methodology}
\begin{figure*}[t]
  \centering
  \includegraphics[width=\textwidth, trim=8mm 7mm 10mm 7mm, clip]{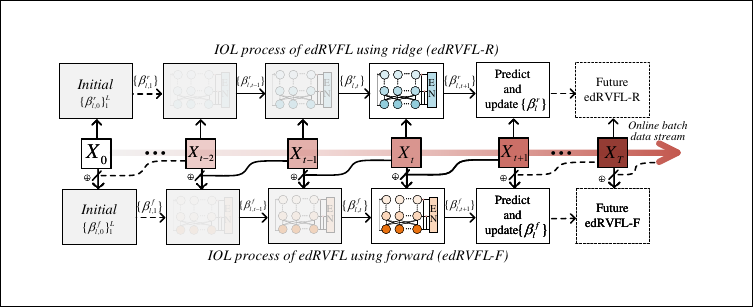}
  \caption{The IOL processes of edRVFL-R/F on batch stream. Stream over time is shown in varied rufous arrow. Extracted features inside edRVFL multilayers are painted respectively in varied brightness blue and yellow to distinguish algorithmic styles of -R and -F. Clustered trainable learners inside edRVFL are progressively updated and uncertainty is gradually removed (shown by increasing sharpness) to present improving performance as data batches come. The past chunks can be discarded without retrievals in the processes. Splines denote the participation of foresight data in -F style.}
  \label{fig 1}
\end{figure*}

In this section, firstly we discuss why -R/-F within IOL frameworks without past retrieval on non-stationary batch stream can mitigate relative \textit{distribution drifting} and \textit{catastrophic forgetting} on learnable weights, also known as the dynamic optimization solutions, compared to a synchronous offline expert. Then the algorithms of IOL frameworks for edRVFL-R/F are derived with variable learning rate and stepwise recursive update policy, which avoid \textit{retrospective retraining} and \textit{large memory usage} for accumulative data. Methods used for deriving IOL algorithms can also be generalized to other Randomized NNs. Finally, intensive theoretical analysis is conducted on IOL processes with -R/-F and related novel regret bounds, and some remarks are presented to demonstrate the advantages of -F. IOL processes of edRVFL-R/F are shown in Fig. \ref{fig 1}.

\subsection{IOL frameworks and discussions on weight drifting}
Given online batch stream $(X,Y) = \{ ({X_t},{Y_t})\} _{t = 0}^T$ and the edRVFL structure, for $0 \leqslant t \leqslant T$, $X$ is projected to $\{ {D_{l,t}} = [{H_{l,t}}|{X_t}]\} _{l = 1,t = 0}^{L,T}$ feature stream using (\ref{1}) and (\ref{2}) in clustered sub-learners. Because the output layer of edRVFL can be regarded as a continuous linear projection cluster $\{ {D_{l,t}} \cdot \beta _{l,t}^{\{ r,f\} } \to {Y_t}\} _{l = 1}^L$ over $t$ time series, the following techniques including loss $L$ are all based on the randomized feature stream of edRVFL $(D,Y) = \{ ({D_{l,t}},{Y_t})\} _{l = 1,t = 0}^{L,T}$.
\begin{lemma} {\bf IOL.} \label{lemma 2.3again}
 For $0 \leqslant t \leqslant T$, with previous ${L _{1..t - 1}}$ concentrated into Bregman divergence $\Delta {U_t}$, IOL processes are realized by continuously optimizing the following equations at every $t$ point on batch stream and removing past rehearsals ostensibly. We describe IOL frameworks for sub-learners of Randomized NN here, and superscripts signify -R/-F styles.
\begin{align} 
\beta _{l,t + 1}^r &= argmi{n_\beta }{\text{ }}{\Delta _{U_{l,t}^r}}(\beta ,\beta _{l,t}^r) + {L_{l,t}}(\beta )\nonumber\\
\beta _{l,t + 1}^f &= argmi{n_\beta }{\text{ }}{\Delta _{U_{l,t}^f}}(\beta ,\beta _{l,t}^f) + {L_{l,t}}(\beta ) \label{8again}\\
&\phantom{{}= \hspace{1.52cm}}+ {\hat L_{l,t + 1}}(\beta )\phantom{{}= \hspace{-0.2cm}} - {\hat L_{l,t}}(\beta )\nonumber
\end{align}

We emphasize that IOL is an enhanced IL as it learns on batch stream with existing \textit{challenges} and \textit{restrictions} considered, and improves inapplicable theories.
\end{lemma}
\begin{theorem}\label{theorem 1}
By using -R/-F within IOL frameworks of edRVFL, for $0 \leqslant t \leqslant T$, trainable weights in edRVFL-R/F can hold approximate optimal solutions as the ones of an up-to-date synchronous offline expert, which revisits and globally learns all previously observed data at each time point. This indicates no relative weight distribution drifting between synchronous offline experts and the IOL processes for edRVFL-R/F. 
\end{theorem}
\noindent{\bf Proof.} Please see the Supplementary Materials A. The closed-form solution retained during IOL has strong learning robustness, with approximate accuracy comparable to well-behaved experts regardless of random stream orderings. 

The IOL frameworks can achieve approximate performance as the best offline experts, and exhibit effective resistance against challenges such as \textit{catastrophic forgetting} and \textit{distribution drifting} of data and concept. It reaches a trade-off between plasticity and stability under online scenario restrictions.
\subsection{Derivations of edRVFL-R/F algorithms within IOL}
In this subsection, practical algorithms for IOL processes of edRVFL-R/F on batch stream are derived considering existing challenges and restraints. Procedures shown in proofs can be generalized to other algorithms within IOL frameworks if conforming to Lemma \ref{lemma 2.3again}.

\begin{theorem}\label{theorem 2}
For the IOL process of edRVFL-R on batch stream, as shown in the Fig. \ref{fig 1}, recursive updates of trainable weights follow $\beta _{l,t + 1}^r = \beta _{l,t}^r - \eta _{l,t}^r(D_{l,t}^T{D_{l,t}}\beta _{l,t}^r - D_{l,t}^T{Y_t})$ with variable learning rate $\eta _{l,t}^r = {({(\eta _{l,0}^r)^{ - 1}} + \sum\limits_{q = 1}^t {D_{l,q}^T{D_{l,q}}} )^{ - 1}}$, where $\eta _{l,0}^r = {({\lambda _l} \cdot I)^{ - 1}}$. This updating policy indicates that incremental updates to weights do not require the involvement of \textit{retrospective retraining} and \textit{high memory usage} of past data.
\end{theorem}
\noindent{\bf Proof.} Please see the Supplementary Materials B.

\begin{algorithm}[htb]
\caption{Batch IOL process of edRVFL using ridge.}\label{alg1}
\begin{algorithmic}
\REQUIRE online batch data stream $(X,Y) = \{ ({X_t},{Y_t})\} _{t = 0}^T$, $L$\\  \hspace{0.62cm} layers, $N$ nodes, ${\lambda _l}$ factor
\ENSURE updated weights by IOL $\{ \beta _{l,T + 1}^r\} _{l = 1}^L$
\vspace{0.3em}
\hrule
\vspace{0.3em}
\STATE {\textsc{INITIALIZE}}: random weights $\{ {W_l}\} _{l = 1}^L$, learning rate $\{ \eta _{l,0}^r\} _{l = 1}^L$, trainable weights $\{ \beta _{l,0}^r\} _{l = 1}^L$ (e.g. $\{ \beta _{l,0}^r\} _1^L = \{ \vec 0\} $)
\STATE
\STATE {\textsc{INCREMENTAL TRAIN}}
\STATE {\textsc{For}} $0 \leqslant t \leqslant T$, do:
\STATE \hspace{0.5cm}Observe $({X_t},{Y_t})$
\STATE \hspace{0.5cm}{\textsc{For}} $1 \leqslant l \leqslant L$, do:
\STATE \hspace{1.0cm}Compute ${D_{l,t}}$ by (\ref{1})-(\ref{2}), and predict with $\beta _{l,t}^r$  
\STATE \hspace{1.0cm}Incur immediate learner loss by ${L_{l,t}}(\beta _{l,t}^r)$
\STATE \hspace{1.0cm}Update $\eta _{l,t}^r$ by Sherman–Morrison–Woodbury law:
\STATE \hspace{1.0cm}$\eta _{l,t}^r = \eta _{l,t -1}^r - \eta _{l,t-1}^rD_{l,t}^T{(I + {D_{l,t}}\eta _{l,t-1}^rD_{l,t }^T)^{ - 1}}$
\STATE \hspace{1.9cm}${D_{l,t }}\eta _{l,t-1}^r$ $\#\eta _{l,t}^r: = \eta _{l,0}^r= \eta _{l,ini}^r$ when $t=0$.
\STATE \hspace{1.0cm}Update $\beta _{l,t + 1}^r$ by (\ref{17})
\STATE \hspace{0.5cm}{\textsc{End For}}
\STATE \hspace{0.5cm}Deduce train results by $En(\{ {D_{l,t}}\cdot\beta _{l,t+1}^r\} _{l = 1}^L,{Y_t})$
\STATE \hspace{0.5cm}$\#$Also can reason on other or previous data.
\STATE
\STATE \hspace{0.5cm}{\textsc{ONLINE TEST}}
\STATE \hspace{0.5cm}Compute $\{ {D_{l,te}}\} _{l = 1}^L$ on test data $({X_{te}},{Y_{te}})$ by (\ref{1})-(\ref{2})
\STATE \hspace{0.5cm}Deduce test results by $En(\{ {D_{l,te}}\cdot\beta _{l,t + 1}^r\} _{l = 1}^L,{Y_{te}})$
\STATE \hspace{0.5cm}$\#t + 1 = 0$ for the initial test and $T+1$ for the final.
\STATE {\textsc{End For}}

\end{algorithmic}
\end{algorithm}
Based on Theorem \ref{theorem 2}, IOL process of edRVFL-R is presented in \textbf{Algorithm \ref{alg1}} with initialization match considered. Batch comes at $t=0$. It is worth noting that $\{ \beta _{l,0}^r\} _1^L$ initialized by zeros indicates learning from blank sketch, and $\beta _{l,t + 1}^r$ still learns from $\beta _{l,t}^r$ containing any prior knowledge, with $\eta _{l,t}^r$ given accordingly. Loss of learners is inferred at each moment.

In some actual scenarios, the next batch ${X_{t + 1}}$ can be accessible in advance at trial $t$ because of tedious labeling works, and we expect to exploit unlabeled data to improve IOL process with -R by using semi-supervision strategy. Combining such an estimated penalty of future batch with online learning equals to substituting -R with -F in IOL framework, which is expected to enhance the precognition of future data, facilitate learning process, and generate better regret bounds.

\begin{theorem}\label{theorem 3}
For the IOL process of edRVFL-F on batch stream, as shown in the Fig. \ref{fig 1}, stepwise updates of learnable weights follow $\beta _{l,t + 1}^f = \beta _{l,t}^f - \eta _{l,t + 1}^f(D_{l,t + 1}^T{D_{l,t + 1}}\beta _{l,t}^f - D_{l,t}^T{Y_t}) + \eta _{l,t + 1}^f(D_{l,t + 1}^T{D_{l,t + 1}} - D_{l,t}^T{D_{l,t}})\beta _{l,0}^f$ with variable learning rate $\eta _{l,t + 1}^f = {({(\eta _{l,0}^f)^{ - 1}} + \sum\limits_{q = 1}^{t + 1} {D_{l,q}^T{D_{l,q}}} )^{ - 1}}$, where $\eta _{l,0}^f =$ $ {({\lambda _l} \cdot I)^{ - 1}}$. Specially, $\beta _{l,t + 1}^f = \beta _{l,t}^f - \eta _{l,t + 1}^f(D_{l,t + 1}^T{D_{l,t + 1}}\beta _{l,t}^f - D_{l,t}^T{Y_t})$ given $\beta _{l,0}^f = 0$. This recursive policy indicates that incremental weight updates do not require the involvement of \textit{retrospective retraining} and \textit{high memory usage} of past data at $t$ time. The optimization processes of IOL with -R/-F can be demonstrated in Fig. \ref{fig 2}.
\end{theorem}
\noindent{\bf Proof.} Please see the Supplementary Materials C. It shows that edRVFL-F maintains different yet more prophetic learning rates compared to edRVFL-R during IOL.

The IOL process of edRVFL-F is presented in \textbf{Algorithm \ref{alg2}} with initialization configurations considered. To ensure fair comparisons, the \textit{same initial setup} with edRVFL-R is kept here. It is noteworthy that the forward style adopts different variable learning rate and updating policy unlike Theorem \ref{theorem 2}. At time $t$, the learning rates of IOL processes with -R and -F are denoted as $\eta _{l,t}^r$ and $\eta _{l,t + 1}^f$ respectively, where the latter can generate better regret bound during IOL process.
\begin{algorithm}[htbp]
\caption{Batch IOL process of edRVFL using forward.}\label{alg2}
\begin{algorithmic}
\REQUIRE online batch data stream $(X,Y) = \{ ({X_t},{Y_t})\} _{t = 0}^T$, $L$\\  \hspace{0.62cm} layers, $N$ nodes, ${\lambda _l}$ factor
\ENSURE updated weights by IOL $\{ \beta _{l,T + 1}^f\} _{l = 1}^L$
\vspace{0.3em}
\hrule
\vspace{0.3em}
\STATE {\textsc{INITIALIZE}}: random weights $\{ {W_l}\} _{l = 1}^L$, learning rate $\{ \eta _{l,0}^f\} _{l = 1}^L$, trainable weights $\{ \beta _{l,0}^f\} _{l = 1}^L$ (e.g. $\{ \beta _{l,0}^f\} _1^L = \{ \vec 0\} $)
\STATE
\STATE {\textsc{INCREMENTAL TRAIN}}
\STATE {\textsc{For}} $0 \leqslant t \leqslant T$, do: \STATE \hspace{0.5cm}$\#T+1$ batch randomly selected for same steps as ridge. 
\STATE \hspace{0.5cm}Observe $({X_{t +1},{Y_t})}$ $\#$${X_t}$ already observed.
\STATE \hspace{0.5cm}{\textsc{For}} $1 \leqslant l \leqslant L$, do:
\STATE \hspace{1.0cm}Compute ${D_{l,t+1}}$ by (\ref{1})-(\ref{2}), and guess with $\beta _{l,t}^f$
\STATE \hspace{1.0cm}$\#{X_{t +1}}$ attended $t$-th optimization in -F term.
\STATE \hspace{1.0cm}Predict ${D_{l,t}}$ of ${X_t}$ with $\beta _{l,t}^f$ $\#$Prediction on current.
\STATE \hspace{1.0cm}$\#{X_t}$ attended $t$-th optimization in loss and -R terms.
\STATE \hspace{1.0cm}Incur immediate learner loss by ${L_{l,t}}(\beta _{l,t}^f)$

\STATE \hspace{1.0cm}Update $\eta _{l,t+1}^f$ by Sherman–Morrison–Woodbury law:
\STATE \hspace{1.0cm}$\eta _{l,t+1}^f = \eta _{l,t }^f - \eta _{l,t}^fD_{l,t+1}^T{(I + {D_{l,t+1}}\eta _{l,t}^fD_{l,t+1}^T)^{ - 1}}$
\STATE \hspace{2.25cm}${D_{l,t +1}}\eta _{l,t}^f$ $\#$Advancing learning rate.
\STATE \hspace{1.0cm}Update $\beta _{l,t + 1}^f$ by (\ref{21.1})
\STATE \hspace{0.5cm}{\textsc{End For}}
\STATE \hspace{0.5cm}Predict ${D_{l,t+1}}$ of ${X_{t +1}}$ with $\{ \beta _{l,t + 1}^f\} _{l = 1}^L$ 
\STATE \hspace{0.5cm}$\#{X_{t +1}}$ can be predicted rather than guessed now.
\STATE \hspace{0.5cm}Deduce train results by $En(\{ {D_{l,t}}\cdot\beta _{l,t+1}^f\} _{l = 1}^L,{Y_t})$
\STATE \hspace{0.5cm}$\#$Also can reason on other or previous data.
\STATE
\STATE \hspace{0.5cm}{\textsc{ONLINE TEST}}
\STATE \hspace{0.5cm}Compute $\{ {D_{l,te}}\} _{l = 1}^L$ on test data $({X_{te}},{Y_{te}})$ by (\ref{1})-(\ref{2})
\STATE \hspace{0.5cm}Deduce test results by $En(\{ {D_{l,te}}\cdot\beta _{l,t + 1}^f\} _{l = 1}^L,{Y_{te}})$
\STATE \hspace{0.5cm}$\#t + 1 = 0$ for the initial test and $T+1$ for the final.
\STATE {\textsc{End For}}

\end{algorithmic}
\end{algorithm}
\subsection{Regret analysis of batch IOL of Randomized learner-R/F}
In this subsection, regret bounds of IOL processes of edRVFL-R/F on batch stream are proved under adversarial assumptions. The introduction of adversary arises from the concerns of network security in Randomized NN and activation functions. Previous theoretical analyses on regret bounds of naive linear model on online single data can refer to \cite{20} and \cite{21}. However, the significant disparity in regret bounds for batch streams renders the original methods ineffective. Here we propose a novel method to derive the regret bounds on new situations. Before that, lemmas that compare cumulative regrets of online learning with respect to the total loss of offline experts are extended from single to batch stream.

\begin{figure}[tbp]
  \centering
  \includegraphics[width=\columnwidth, trim=8mm 8mm 7mm 7mm, clip]{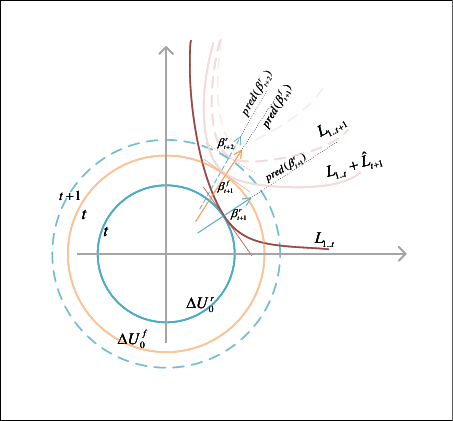}
  \caption{One online learner inside edRVFL uses -R and -F to update respectively. The learner using ridge (blue solid marked) try to match ${L_{1..t}}$ (dark red) and evolve into $\beta _{t + 1}^r$ (blue arrow) on batch $t$, and same logic as the $\beta _{t + 2}^r$ (blue dashed) at time $t+1$ for updating and prediction $pred(\beta _{t + 2}^r)$ over ${L_{1..t+1}}$ (light red dashed). The learner with forward term (yellow solid) try to meet ${L_{1..t}} + {\hat L_{t + 1}}$ (light red solid) including estimated cost at time $t$. It is suggested that the -F style can rectify the IOL process.}
  \label{fig 2}
\end{figure}

\begin{lemma} {\bf Online-to-offline (relative) cumulative regrets for batch stream and ridge regularization.} \label{lemma 3.2}
Assume that the $l$-th sub-learner inside edRVFL incrementally evolves on extracted feature stream of $T$ batches using -R, and $\beta _{l, \cdot }^r$ belongs to the offline expert solution set $\mathcal{B}$. One possible solution can be given by the global offline expert which learns on the entire dataset, as described in Lemma \ref{lemma 2.1}. The following formula compares the regrets generated in IOL process with -R to those of the offline expert model.
\begin{equation} \label{22}
\begin{split}
\sum\limits_{t = 1}^T {{L_{l,t}}(\beta _{l,t}^r)}  - ({\Delta _{U_{l,0}^r}}(\beta _{l, \cdot }^r,\beta _{l,0}^r) + {L_{l,1..T}}(\beta _{l, \cdot }^r)) \\
= \sum\limits_{t = 1}^T {{\Delta _{U_{l,t + 1}^r}}(\beta _{l,t}^r,\beta _{l,t + 1}^r)}  - {\Delta _{U_{l,T + 1}^r}}(\beta _{l, \cdot }^r,\beta _{l,T + 1}^r)
\end{split}
\end{equation}
\end{lemma}

The optimal status of the offline expert can be compared if set $\beta _{l, \cdot }^r = \beta _{l, * }^r$, namely the global optimum solution. To study the regret generation paradigm in IOL with -R and derive regret bounds, we exported Lemma \ref{lemma 3.2} to batch stream scenarios based on Lemma 4.2 \cite{20}. (\ref{22}) transforms the relative cumulative regrets between IOL of sub-learner-R and offline expert into the cumulative sum of Bregman divergence.

\begin{theorem}\label{theorem 4}
For the IOL process of the $l$-th online learner inside edRVFL-R working on batch stream, during $0 \leqslant t \leqslant T$, the upper bound of online-to-offline cumulative regrets between this learner and an offline expert can be written as:
\begin{equation*} 
\begin{split}
\sum\limits_{t = 1}^T {{L_{l,t}}(\beta _{l,t}^r)}  - \mathop {min}\limits_{\beta _{l, \cdot }^r} ({\Delta _{U_{l,0}^r}}(\beta _{l, \cdot }^r,\beta _{l,0}^r) + {L_{l,1..T}}(\beta _{l, \cdot }^r)) \\
\leqslant 2Y_m^2b(N + k)In(1 + \frac{{TD_m^2b}}{{{\lambda _l}s}})
\end{split}
\end{equation*}
where $\beta _{l, \cdot }^r$ is one solution of the offline expert, $s$ is the amplification factor of origin ${\lambda _l}$, $In$ denotes natural logarithm, ${D_m} = \mathop {\max }\limits_{1 \leqslant t \leqslant T} \{ ||{D_{l,t}}|{|_\infty }\} $, and ${Y_m} = \mathop {\max }\limits_{1 \leqslant t \leqslant T} \{ ||{Y_t}|{|_\infty },||{D_{l,t}}\beta _{l,t}^r|{|_\infty }\} $.
\end{theorem}
\noindent{\bf Proof.} Please see the Supplementary Materials D. The outcome is typical $log(T)$ style and dependent on some hyper-parameters, such as number of neuron nodes, batch capacity, regularization factor, and data dimensions.

\begin{lemma} {\bf Online-to-offline (relative) cumulative regrets for batch stream and forward regularization.} \label{lemma 3.3}
Assume that the $l$-th sub-leaner inside edRVFL incrementally evolves on extracted feature stream of $T$ batches using -F, and $\beta _{l, \cdot }^f$ belongs to the offline expert solution set $\mathcal{B}$. One possible solution can be given similar to the one in Lemma \ref{lemma 3.2} to ensure fair comparison. The following formula compares the regrets of IOL process with -F to those of the offline expert.
\begin{align}
&\sum\limits_{t = 1}^T {{L_{l,t}}(\beta _{l,t}^f)}  - ({\Delta _{U_{l,0}^f}}(\beta _{l, \cdot }^f,\beta _{l,0}^f) + {L_{l,1..T}}(\beta _{l, \cdot }^f)) \nonumber\\
=& \sum\limits_{t = 1}^T {({\Delta _{U_{l,t + 1}^f}}(\beta _{l,t}^f,\beta _{l,t + 1}^f)}  - {{\hat L}_{l,t + 1}}(\beta _{l,t}^f) + {{\hat L}_{l,t}}(\beta _{l,t}^f)) \nonumber\\ 
-& {\Delta _{U_{l,T + 1}^f}}(\beta _{l, \cdot }^f,\beta _{l,T + 1}^f) + {{\hat L}_{l,T + 1}}(\beta _{l, \cdot }^f) - {{\hat L}_{l,1}}(\beta _{l, \cdot }^f) \label{31}\\
+& {\Delta _{U_{l,1}^f}}(\beta _{l, \cdot }^f,\beta _{l,1}^f) - {\Delta _{U_{l,0}^f}}(\beta _{l, \cdot }^f,\beta _{l,0}^f) \nonumber 
\end{align}
\end{lemma}

The optimal status of the offline expert can be compared if set $\beta _{l, \cdot }^f = \beta _{l, * }^f$. $\beta _{l, * }^f$ denotes the GLOBAL optimal solution of expert. To derive the regret bounds of IOL process with -F, we extend Lemma 5.2 in \cite{20} focusing on a single stream to Lemma \ref{lemma 3.3} to enable adaptation to batch stream. (\ref{31}) transforms the relative cumulative regrets between IOL of sub-learner-F and the offline expert into the accumulative sum of mixed terms involving Bregman divergence and ${l_2}$ loss. In contrast to (\ref{22}), (\ref{31}) is more complex.

\begin{theorem}\label{theorem 5}
For the IOL process of the $l$-th online learner inside edRVFL-F working on batch stream, during $0 \leqslant t \leqslant T$, the upper bound of online-to-offline cumulative regrets between this learner and an offline expert can be written as:
\begin{align}
&\sum\limits_{t = 1}^T {{L_{l,t}}(\beta _{l,t}^f)}  - \mathop {min}\limits_{\beta _{l, \cdot }^f} (\frac{1}{2}{(\beta _{l, \cdot }^f)^T}{(\eta _{l,0}^f)^{ - 1}}\beta _{l, \cdot }^f + {L_{l,1..T}}(\beta _{l, \cdot }^f)) \hfill \nonumber\\
\leqslant& \frac{1}{2}Y_m^2b(N + k)(In(1 + \frac{{TD_m^2b}}{{{\lambda _l}s}}) - In(1 + \frac{{(T - 1)D_m^2b}}{{{\lambda _l}s + 2D_m^2b}})) \hfill \nonumber\\
\leqslant& \frac{1}{2}Y_m^2b(N + k)In(1 + \frac{{TD_m^2b}}{{{\lambda _l}s}}) \nonumber 
\end{align}
where $\beta _{l, \cdot }^f$ is one solution of the offline expert, $s$ is the amplification factor of origin ${\lambda _l}$, $In$ denotes natural logarithm, ${D_m} = \mathop {\max }\limits_{1 \leqslant t \leqslant T} \{ ||{D_{l,t}}|{|_\infty }\} $, and ${Y_m} = \mathop {\max }\limits_{1 \leqslant t \leqslant T} \{ ||{Y_t}|{|_\infty },||{D_{l,t}}\beta _{l,t}^f|{|_\infty }\} $. This suggests IOL with -F achieves \textit{lower regrets}, and its upper cumulative regret bound is at least 4 times better than that of IOL with -R.
\end{theorem}
\noindent{\bf Proof.} Please see the Supplementary Materials E.

We promote the use of -F for the following reasons. Incorporating an estimated regularization into -R facilitates the enhancement of variable learning rates, which accelerates the ensemble IOL process by dynamically improving the learning effects across multiple sub-learners. Additionally, referring to the results of Theorem \ref{theorem 4} and Theorem \ref{theorem 5}, -F achieves better regrets to ridge style during IOL processes. 

In our experimental section, we will further illustrate -F's advantages, including the enhancement of model's robustness and performance. It is advocated to substitute -R with -F to obtain lower predictive loss on future data, faster learning progress, and better cumulative regrets in IOL.
\subsection{Corollary and remarks of regret bounds of IOL}
Assume that the ${\gamma ^{\{ r,f\} }}$ denotes the upper cumulative regret bound of IOL process between single learner with -R or -F and the offline expert, and ${\Gamma ^{\{ r,f\} }}$ denotes the upper cumulative regret bound of IOL process between edRVFL-R/F and the offline expert respectively. This subsection contains some further studies on the regret bounds.
\begin{corollary}\label{coro1}
Theorem \ref{theorem 4} and Theorem \ref{theorem 5} are proved for the batch stream conditions. It is easy to derive the bounds of single data stream based on them, and obtain similar results as Lemma \ref{lemma 2.4} if set $b=1$. This means our proposed methods are more general. If the number of batch-sample is not constant but given by variable series $\{ b|{b_i} \in {\mathbb{N}^ + },1 \leqslant i \leqslant T\} $, then the regret bounds can be:
\begin{equation*}
\begin{split}
{\gamma ^r} &\leqslant 2Y_m^2{b_m}(N + k)In(1 + \frac{{TD_m^2{b_m}}}{{{\lambda _l}s}})\\
{\gamma ^f} &\leqslant \frac{1}{2}Y_m^2{b_m}(N + k)(In(1 + \frac{{TD_m^2{b_m}}}{{{\lambda _l}s}}) \\
&- In(1 + \frac{{(T - 1)D_m^2{b_m}}}{{{\lambda _l}s + 2D_m^2{b_m}}}))
\end{split}
\end{equation*}
where ${b_m} = \mathop {\max }\limits_{1 \leqslant i \leqslant T} {b_i}$, ${D_m} = \mathop {\max }\limits_{1 \leqslant t \leqslant T} \{ ||{D_{l,t}}|{|_\infty }\} $, and ${Y_m} = \mathop {\max }\limits_{1 \leqslant t \leqslant T} \{ ||{Y_t}|{|_\infty },||{D_{l,t}}\beta _{l,t}^{\{ r,f\} }|{|_\infty }\} $.
\end{corollary}
\begin{corollary}\label{coro2}
Theorems on regrets can indicate and guide configurations of hyper-parameters. For Theorem \ref{theorem 4} and Theorem \ref{theorem 5}, it is natural to set ${\lambda _l}s \leftarrow {\lambda _l}b$  when data volume expands $b$ times. Therefore,
\begin{align}
{\gamma ^r} &\leqslant 2Y_m^2b(N + k)In(1 + \frac{{TD_m^2}}{{{\lambda _l}}})\hfill \nonumber\\
{\gamma ^f} &\leqslant \frac{1}{2}Y_m^2b(N + k)(In(1 + \frac{{TD_m^2}}{{{\lambda _l}}}) - In(1 + \frac{{(T - 1)D_m^2}}{{{\lambda _l} + 2D_m^2}})) \hfill \nonumber\\
&\leqslant \frac{1}{2}Y_m^2b(N + k)In(1 + \frac{{TD_m^2}}{{{\lambda _l}}})\nonumber
\end{align}
where $\beta _{l, \cdot }^{\{ r,f\} }$ is the solution of the offline expert, ${D_m} = \mathop {\max }\limits_{1 \leqslant t \leqslant T} \{ ||{D_{l,t}}|{|_\infty }\} $, ${Y_m} = \mathop {\max }\limits_{1 \leqslant t \leqslant T} \{ ||{Y_t}|{|_\infty },||{D_{l,t}}\beta _{l,t}^{\{ r,f\} }|{|_\infty }\} $. Compared to Lemma \ref{lemma 2.4}, cumulative upper regret bounds expand $b$ times while time $T$ contracts to $T/b$ when applied to the same sequence.
\end{corollary}
\begin{corollary}\label{coro3}
As the sub-learners are aggregated and expected to produce a more accurate consequence inside edRVFL, the upper cumulative regret bounds of edRVFL will not be higher than the maximum of regret bounds of ensemble learners during IOL.
\begin{equation*}
\begin{split}
{\Gamma ^r} &\leqslant 2Y_M^2b(N + k)In(1 + \frac{{TD_M^2b}}{{{\lambda _M}s}})\\
{\Gamma ^f} &\leqslant \frac{1}{2}Y_M^2b(N + k)(In(1 + \frac{{TD_M^2b}}{{{\lambda _M}s}}) \\
&- In(1 + \frac{{(T - 1)D_M^2b}}{{{\lambda _M}s + 2D_M^2b}}))
\end{split}
\end{equation*}
where ${\lambda _M} = \mathop {\min }\limits_{1 \leqslant l \leqslant L} \{ {\lambda _l}\} $, ${D_M} = \mathop {\max }\limits_{1 \leqslant l \leqslant L} \{ {D_m}\} $, and ${Y_M} = \mathop {\max }\limits_{1 \leqslant l \leqslant L} \{ {Y_m}\} $. ${D_m}$ and ${Y_M}$ are values of $L$ sub-learners.

\end{corollary}
\begin{corollary}\label{coro4}
Increasing rates of cumulative regret bounds can be compared to illustrate the accumulation process of online error. Take the derivatives of ${\gamma ^{\{ r,f\} }}$ w.r.t time $t$:
\begin{equation*}
\begin{split}
\left[ {\begin{array}{*{20}{c}}
  {\frac{{\partial {\gamma ^r}(t)}}{{\partial t}}} \\ 
  {\frac{{\partial {\gamma ^f}(t)}}{{\partial t}}} 
\end{array}} \right] = \left[ {\begin{array}{*{20}{c}}
  {{a_1}\frac{{{a_2}}}{{1 + {a_2}t}}} \\ 
  {\frac{{{a_1}}}{4}(\frac{{{a_2}}}{{1 + {a_2}t}} - \frac{{{a_2}}}{{1 + {a_2} + {a_2}t}})} 
\end{array}} \right]
\end{split}
\end{equation*}
where ${a_1} = 2Y_m^2b(N + k)$, ${a_2} = \frac{{D_m^2b}}{{{\lambda _l}s}}$. It shows that: (1) both increasing rates tend to flatten out over time due to the usage of variable gradients and more data; (2) the derivative value of ${\gamma ^{ f }}$ is smaller than that of ${\gamma ^{ r }}$, and sub-learner of edRVFL-F has slower growth speed on cumulative regret bound compared to that of edRVFL-R during IOL, which validates better performance and inspires us to advocate the usage of -F again. This phenomenon can be observed in experiments.
\end{corollary}
\begin{rem}\label{remark 1}
It is assumed that the value of $||{D_{l,t}}\beta _{l,t}^{}|{|_\infty }$ lie in $[ - {Y_m},{Y_m}]$. If the assumption is not satisfied, clipping operations can restrict ranges but this requires some knowledge of $Y$. During IOL processes, $||{Y_t}|{|_\infty }$ should be refreshed, or one needs to know some prior knowledge of $Y$.
\end{rem}
\begin{rem}\label{remark 2}
Assumptions and expansions are made under adversarial setup in the above derivations, e.g. (\ref{26})-(\ref{28}), (\ref{36})-(\ref{37}). $log(T)$-style regret bounds can be tightened or adversarial setup removed if conditions related but not limited to activation functions and feature distribution can be provided.
\end{rem}
\begin{rem}\label{remark 3}
One can define regret bounds for further comparisons by shifting regularization terms. For example, ${\bar \gamma ^r}: = \sum\limits_{t = 1}^T {{L_{l,t}}(\beta _{l,t}^r)}  - {L_{l,1..T}}(\beta _{l, \cdot }^r)$, then ${\bar \gamma ^r} \leqslant {\gamma ^r} + {\Delta _{U_{l,0}^r}}(\beta _{l, \cdot }^r,\beta _{l,0}^r)$; or ${\hat \gamma ^r}: = {\Delta _{U_{l,0}^r}}(\beta _{l,T + 1}^r,\beta _{l,0}^r) + {\gamma ^r}$, then ${\hat \gamma ^r} \leqslant {\gamma ^r} + {\Delta _{U_{l,0}^r}}(\beta _{l,T + 1}^r,\beta _{l,0}^r)$. Above Bregman terms are easy to compute as it only includes initial and final values. 
\end{rem}
\section{Experiments}
This section documents experimental results on numerical simulations, regression tasks, classification tasks, huge dataset classification tasks, and challenge studies. We explored the efficacy of IOL processes with -R/-F and advantages of -F.
\subsection{Numerical simulations on -R/-F within IOL}
Refer to Supplementary Materials F for the numerical simulations, and detailed comparisons of IOL processes using -R/-F algorithms on single and batch streams. 
\subsection{Comparisons of performance on regression tasks}
To verify our proposed IOL frameworks of edRVFL and study the efficacy of employing forward learning in batch stream scenarios, in this subsection, we conducted performance comparisons of online learning versions of selected popular neural networks on regression tasks of a total 15 types of public UCI datasets. These datasets, as shown in Table \ref{table 1}, are usually used to test network properties in previous research. Additionally, ablation experiments were carried out for in-depth analysis of IOL for edRVFL-R/F.
\begin{table}[!htbp]
\renewcommand\arraystretch{1}
\scriptsize
\centering
\caption[]{\\ }
\caption*{{Description of regression datasets}}
\begin{tabular}{ccc}
\toprule
$\text{Dataset}$  &
$\text{Size}$&  
$\text{Features}$
 \\
\midrule
        abalone &  4177 &  8 \\
 air foil noise &  1503 &  5 \\
          auto mpg &   392 &  7 \\
          concrete &  1030 &  8 \\
      daily demand &    60 & 12 \\
     eating habits &  2111 & 16 \\
       forestfires &   517 & 12 \\
           machine &   209 &  7 \\
          mortgage &  1049 & 15 \\
          nox 2011 &  7411 &  7 \\
          nox 2012 &  7628 &  7 \\
          nox 2015 &  7384 &  7 \\
     weather Izmir &  1461 &  9 \\
  wine quality red &  1599 & 11 \\
wine quality white &  4898 & 11 \\
 \bottomrule
\end{tabular}
\label{table 1}
\end{table}

Main methods and models involved in this experiment were as follows:
(1)	SMAC3 \cite{23}: sequential model-based algorithm configuration offered a robust framework for Bayesian optimization to impartially help in configuring well-performing hyper-parameters of various networks.
(2)	edRVFL-R/F: our derived algorithms in Section IV which realized IOL process of edRVFL using ridge or forward learning style on dynamic batch stream.
(3)	OLRVFL \cite{24}: the online version of prototype RVFL.
(4)	pRVFL \cite{25}: configured from scratch, pRVFL could be grown, pruned, and recalled on data streams adaptively.
(5)	OS-ELM \cite{26}: online sequential extreme learning machine (OS-ELM) learned data chunk-by-chunk with fixed or varying chunk sizes.
(6)	DOS-ELM \cite{27}: used forgetting factor in OS-ELM to adapt previous memory for future data stream in learning epoch.
(7)	OBLS \cite{16}: the online version of broad learning system.
(8)	OSNN \cite{28}: the best-ranked feed-forward neural network using SELU activation function and alpha-dropout techniques worked online and required retrospective retraining process.

For regression datasets, each was partitioned into 4-folds and the validation set was detached from the training and testing sets with proper percentage considering dataset size. Performance was tested via cross-validation to ensure consistent results. In one trial, the network with hyper-parameters configured by SMAC3 was trained on every fold separately, which was repeated on two different random seeds for each fold. Normalization followed the preceding methods in \cite{27,28}. Assessment metric on validations was set to be the cost of incumbents which was optimized in SMAC3. The configuration space of important hyper-parameters of Randomized NN variants in SMAC3 is enumerated as Table \ref{table 2}. In addition, the three controllable parameters of pRVFL were tuned in range [0, 1], the number of hidden neurons of ELM was in range [4, 2048], the total numbers of neurons inside the BLS first and second layers were in range [4, 2048] and [200, 10000] respectively. The optimization space of OSNN retained the same with \cite{28} and we imposed early stopping to restrict degradation in normal online learning scenarios. The early stopping was determined on smoothed learning curves of 70 epochs of the validation set. To evaluate the performance of networks, averaged RMSE was used for regression tasks.

\begin{table}[!htbp]
\renewcommand\arraystretch{1}
\scriptsize
\centering
\caption[]{\\ }
\caption*{Main hyperparameter configuration space of SMAC3}
\begin{tabular}{cccc}
\toprule
$\text{Hp}$  &
$\text{Description}$&  
$\text{Range (regress / class)}$&
$\text{Type}$
 \\
\midrule
trial&SMAC3 trial times&200 / 400 &integer\\
$N$ &neuron amount per layer&[4,1024] / [4,2048]&integer\\
$L$&number of layers&[2, 16] / [2, 20]&integer\\
${\log _2}({\lambda ^{ - 1}})$&regularization factor&$[ - 6,6]$ / $[ - 8,8]$&integer\\
$b$&batch data proportion&[0.001, 0.2] / [0.001, 0.25]&float\\
$g$&activation function&['sigmoid','ReLU'…]
&class\\
 \bottomrule
\end{tabular}
\label{table 2}
\end{table}
\begin{table*}[!htbp]
\renewcommand\arraystretch{1}
\scriptsize
\centering
\caption[]{\\ }
\caption*{RMSE performance of online networks on regression tasks}
\begin{tabular}
{*{17}{c@{\hspace{1.0em}}}}
\hline
\toprule
\multirow{2}{*} {Dataset}
&\multicolumn{2}{c@{\hspace{1.6em}}}{OLRVFL} &     \multicolumn{2}{c@{\hspace{1.6em}}}{pRVFL} &   \multicolumn{2}{c@{\hspace{1.6em}}}{DOS-ELM} &    \multicolumn{2}{c@{\hspace{1.6em}}}{OS-ELM} &      \multicolumn{2}{c@{\hspace{1.6em}}}{OBLS} &  
\multicolumn{2}{c@{\hspace{1.6em}}}{OSNN}&
\multicolumn{2}{c@{\hspace{1.6em}}}{edRVFL-R}&
\multicolumn{2}{c@{\hspace{1.6em}}}{edRVFL-F}   \\

 &       TR &         TE &       TR &         TE &      TR &         TE &      TR &         TE &      TR &          TE &      TR &          TE &      TR &          TE &      TR &          TE \\
\midrule               
abalone &       0.0805 &     0.0819 &  0.0995 &     0.0877 &  0.0833 &      0.0894 &  0.0751 &      0.0772 &  0.0742 &       0.0760 &0.0909 &      0.0925  &  0.0728 &     0.0745 &\textbf{0.0726} &     \textbf{0.0736} \\
           
air foil noise &       0.0541 &     0.0533 &  0.0446 &     0.0471 &   \textbf{0.0210} &      0.0257 &  0.0427 &      0.0543 &  0.0603 &      0.0632 &0.0296 &      0.0338   & 0.0257 &     0.0258 &\textbf{0.0210} &     \textbf{0.0226}  \\
    
auto mpg &       0.1059 &     0.1198 &  0.1107 &     0.1248 &  0.0985 &      0.1075 &  0.1579 &      0.1659 &  0.0714 &      \textbf{0.0616} &\textbf{0.0029} &      0.0916  &  0.0737 &     0.0831 &0.0726 &     0.0661 \\
          
concrete &       0.0922 &     0.0967 &  0.0981 &     0.1036 &   0.0910 &      0.0903 &  0.2204 &      0.2936 &  \textbf{0.0807} &      \textbf{0.0864} &  0.0994 &      0.0996  &0.0824 &     0.0919 &0.0808 &     0.0872 \\
          
daily demand &       0.0306 &     0.0335 &  0.0399 &     0.0544 &  0.0346 &      0.0435 &   0.0430 &      0.0477 &  0.0494 &       0.0570 & \textbf{0.0136} &      \textbf{0.0218}  &  0.0145 &     0.0239 &0.0235 &     0.0252\\
      
eating habits &       0.0902 &     0.1044 &  0.0655 &     0.0672 &  0.0706 &      0.0835 &  0.0622 &      0.0639 &  \textbf{0.0555} &      0.0578 &  0.0846 &      0.1192  &0.0561 &     \textbf{0.0564} &0.0571 &      0.0570 \\
     
forestfires &      0.0615 &     0.0657 &  0.6081 &     0.1619 &  0.0651 &       0.0550 &  0.0609 &      0.0618 &  0.0515 &      0.0523 & \textbf{0.0105} &      0.1056    &0.0559 &     0.0541 &0.051 &     \textbf{0.0515}  \\
       
machine &       0.1277 &     0.1492 &  0.1226 &     0.1492 &  0.0421 &       0.0550 &  0.0511 &      0.0528 &  \textbf{0.0028} &      0.0163 & 0.0035 &      \textbf{0.0151}  &  0.0111 &      0.0210 &0.0031 &     0.0155\\
           
mortgage &     0.0726 &     0.0765 &  0.0835 &     0.0993 &  0.0093 &      0.0101 &  0.0081 &      0.0107 &  \textbf{0.0048} &      0.0061 &  0.0106 &      0.0149   &0.0063 &     \textbf{0.0054} &0.0064 &     0.0056  \\
          
nox 2011 &         0.048 &     0.0519 &  0.0591 &     0.0607 &  0.0441 &      0.0519 &  0.0318 &      0.0343 &  0.0328 &       0.0430 &0.0528 &      0.0547  & \textbf{0.0310} &     0.0319 &   0.0311 &     \textbf{0.0317}\\
          
nox 2012 &       0.0368 &     0.0392 &  0.0391 &     0.0416 &  0.0328 &       0.0350 &  0.0412 &       0.0510 &  0.0348 &      0.0369 &0.0988 &      0.1021  &0.0332 &     0.0336 &  \textbf{0.0322} &     \textbf{0.0335} \\
          
nox 2015 &      0.0559 &     0.0603 &  0.0629 &     0.0776 &  0.0487 &      0.0519 &  0.0623 &      0.0707 &  0.0501 &      0.0528 &0.0724 &      0.0814   &0.0509 &      0.0510 &  \textbf{0.0404} &     \textbf{0.0413} \\
          
weather Izmir &       0.0516 &     0.0569 &  0.0474 &     0.0571 &   0.059 &      0.0953 &  0.0233 &      0.0232 &  0.0212 &      0.0221 & 0.0316 &      0.0433   &\textbf{0.0181} &     \textbf{0.0201} &0.0193 &     0.0204 \\
     
wine quality red &       0.0812 &     0.0872 &  0.3276 &     0.2932 &  0.0858 &      0.0865 &  0.1476 &      0.1492 &  0.1251 &      0.1415 &  0.1181 &      0.1819  &0.0744 &     0.0828 &\textbf{0.0738} &     \textbf{0.0811} \\
  
wine quality white &       0.1977 &     0.2041 &  0.1688 &     0.1936 &  0.1156 &      0.1191 &  0.1042 &       0.113 &  0.1043 &      0.1271 &0.4082 &      0.4199  &  0.0995 &     0.0995 &\textbf{0.0838} &     \textbf{0.0843} \\
           Average &       0.0791 &     0.0854 &  0.1318 &     0.1079 &  0.0601 &      0.0666 &  0.0755 &      0.0846 &  0.0546 &        0.0600 & 0.0752 &      0.0985  &0.0470 &     0.0503 &  \textbf{0.0446} &     \textbf{0.0464} \\
\bottomrule
\end{tabular}
\label{table 3}
\end{table*}
Experimental results of the above 8 methods on 15 regression tasks are listed in Table \ref{table 3}. Averaged training and testing RMSE of repeated fold trials are denoted as TR and TE respectively, whose highest accuracy values of each dataset are highlighted in \textbf{bold}. Based on the statistics, some conclusions can be drawn. IOL frameworks of edRVFL-R/F networks outperform others on 11 tasks and the average of all datasets. Because of the intervention with forward knowledge, edRVFL-F gets the advantages of prediction RMSE on 11 datasets even compared to edRVFL-R within IOL frameworks, which is especially more evident on dynamic streams from larger datasets. This accords with the results from numerical simulations, where -R tends to yield more regrets compared to -F in the online learning process of continuous prediction tasks. The RVFL variants, OLRVFL and OS-ELM, perform similarly on online tasks while OBLS stands out as it adopted techniques of sparse reconstruction and multi-map extractions. The effectiveness gain of DOS-ELM to OS-ELM comes from forgetting regularization. OSNN is not proficient enough for batch streams of large datasets and it consumes much resources on fine-tuning of parameters as well as memory revisits. Results suggest that our IOL frameworks for edRVFL-R/F are effective and IOL of edRVFL-F can generate immediate high-quality responses and decision-making to requests on online batch streams. 

\begin{table}[!htbp]
\renewcommand\arraystretch{1}
\scriptsize
\centering
\caption[]{\\ }
\caption*{Wilcoxon testing matrix of network performance on regression}
\begin{tabular}{*{9}{c@{\hspace{0.2em}}}}
\toprule
$\text{Model}$  &
$\text{OLRVFL}$&
$\text{pRVFL}$&
$\text{DOS-ELM}$&
$\text{OS-ELM}$&
$\text{OBLS}$&
$\text{OSNN}$&
$\text{edRVFL-R}$&
$\text{edRVFL-F}$
 \\
\midrule                
OLRVFL &- &    0.0084 &   0.0591 &    0.5897 &  0.0035 &0.2808 &4.58e-6 &3.07e-6 \\
pRVFL &0.0084 &         - &   0.0009 &    0.0144 &  0.0004 &0.4186 &3.07e-6 &3.07e-6 \\
DOS-ELM &0.0591 &    0.0009 &        - &    0.2058 &  0.2058 &0.0591 &0.0003 &9.99e-6 \\
OS-ELM &0.5897 &    0.0144 &   0.2058 &         - &  0.0202 &0.2808 &2.12e-5 &3.75e-6 \\
OBLS &0.0035 &    0.0004 &   0.2058 &    0.0202 &       - &0.0213 &0.0191 &0.0004 \\
OSNN & 0.2808 &  0.4186 &  0.0591 &  0.2808 &  0.0213 &- &0.0005 &0.0002 \\
edRVFL-R &4.58e-6 &  3.07e-6 &   0.0003 &  2.12e-5 &  0.0191 &0.0005 &- &0.0381 \\
edRVFL-F &3.07e-6 &    3.07e-6 &   9.99e-6 &    3.75e-6 &  0.0004 &0.0002 &0.0381 &- \\
 \bottomrule
\end{tabular}
\label{table 4}
\end{table}

To validate the result's credibility, p-value of paired Wilcoxon test was conducted to identify the statistical significance of network outcomes based on their ranking arrays on all tasks in Table \ref{table 3} and ranksums function. The Wilcoxon symmetric matrix is listed in Table \ref{table 4}, which shows that edRVFL-based variants surpass others and using edRVFL-F enables better performance than using edRVFL-R in IOL processes as the relevant p-value ${\text{ = 0}}{\text{.0381\textless 0}}{\text{.05}}$.

\begin{table}[!htbp]
\renewcommand\arraystretch{1}
\scriptsize
\centering
\caption[]{\\ }
\caption*{Baseline configurations and online testset RMSE of \textit{Weather Izmir}}
\begin{tabular}{*{7}{c@{\hspace{0.6em}}}}
\toprule
$\text{Hp of}$  &
\multirow{2}{*} {$N$}&  
\multirow{2}{*} {$L$}&
\multirow{2}{*} {${\log _2}({\lambda ^{ - 1}})$}&
\multirow{2}{*} {$b$}&
\multirow{2}{*} {$g$}&
$\text{Random}$
 \\
 $\text{baseline}$  &
&  
&
&
&
&\text{weights}\\
\midrule
 $\text{Value}$  &
85&  
16&
-4&
0.045&
\text{sigmoid}&$ N(0,1)$\\
\toprule
Time Points &      0 &      5 &     10 &     15 &     20 &     22 \\
\midrule
edRVFL-R RMSE (ensem.) & 0.5784 & 0.1230 & 0.0527 & 0.0340 & 0.0247 & 0.0228 \\
edRVFL-F RMSE (ensem.)  & 0.5784 & 0.0418 & 0.0304 & 0.0262 & 0.0231 & 0.0213 \\
edRVFL-R RMSE (avr.) & 0.5784 & 0.1294 & 0.0603 & 0.0414 & 0.0317 & 0.0295 \\
edRVFL-F RMSE (avr.) & 0.5784 & 0.0529 & 0.0388 & 0.0337 & 0.0290 & 0.0283 \\
 \bottomrule
\end{tabular}
\captionsetup{font=scriptsize} \captionsetup{singlelinecheck=false,justification=raggedright}
\caption*{*Note the ensem. means ensemble strategy, avr. means average of multiple learners.}
\label{table 5}
\end{table}

\begin{figure}[!htbp]
  \centering
  \includegraphics[width=\columnwidth,trim=0mm 0mm 10mm 7mm, clip]{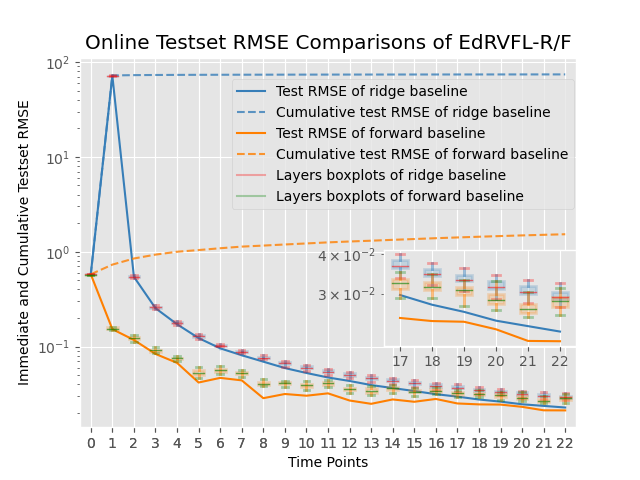}
  \caption{Testset RMSE variation curves of edRVFL-R/F baselines on regression task during IOL processes. Baselines' setup and values can be referred to Table \ref{table 5}, and performance is displayed by testset RMSE over time (data batches). Solid lines: immediate ensemble RMSE of baselines. Dashed lines: cumulative ensemble RMSE. Boxplots: immediate RMSE statistics of interior multiple sub-learners (layers). X-axis is locally enlarged in the inlaid subfigure.}
  \label{fig 4}
\end{figure}

To demonstrate the superiority of -F, SMAC3 configurations and detailed comparative tests of IOL processes of edRVFL-R/F were presented here based on the online stream of \textit{Weather Izmir}. To fairly contrast, edRVFL-R/F were designed and adjusted synchronously to the same network structure optimum, denoted by baselines as shown in Table \ref{table 5}, with the same SMAC3 space shared and incumbent loss set to the average of validation RMSE of their two repeated trials. We collected baseline trainable weights $\beta _{l,t}^r$ and $\beta _{l,t}^f$, and calculated immediate testset RMSE for all time points, as it could reflect performance changes. Note that online immediate RMSE trends of testset here included types of entire ensemble network and respective learners (each layer) for careful studies from two views, where the values are recorded in Table \ref{table 5} and displayed by several solid curves and boxplots in Fig. \ref{fig 4}. 

From Table \ref{table 5} and Fig. \ref{fig 4}, it can be concluded that testset RMSE of two baseline IOL processes declines with more stream batches being trained while IOL of edRVFL-F is accelerated by -F and outperforms the edRVFL-R because regrets are released by reasonable learning rate and precognition learning of using -F during IOL processes, regardless of ensemble network or respective learners angle. Likewise, the final RMSE results of edRVFL-F are better than the edRVFL-R's. The immediate regrets of edRVFL-R/F in IOL processes exhibit a gap, which diminishes over time. If the learning iterations were limited or terminated prematurely, there might be more differences in performance between them. RMSE is notably reduced by the ensemble strategy and clustered learners inside edRVFL-F are generally superior to the learners of edRVFL-R as the boxplots show. IOL process with -F has a smaller fluctuation range and stabilizes faster. The cumulative RMSE regret bound of edRVFL-R is more than 4 times higher than that of edRVFL-F because of initial slower adaptation to the online task and more prediction regrets of using -R in IOL.

edRVFL-R/F were incrementally updated on non-stationary streams during IOL, and had \textit{no retrospective retraining} of previous data and delivered rapid responses to requests. The immediate RMSE curves observed in Fig. \ref{fig 4} show an overall decreasing trend, indicating there was no obvious \textit{catastrophic forgetting}, and had good suppression of  \textit{distribution drift}.

\begin{figure*}[t]
\centering
\subfigure[]{\includegraphics[width=0.32\textwidth]{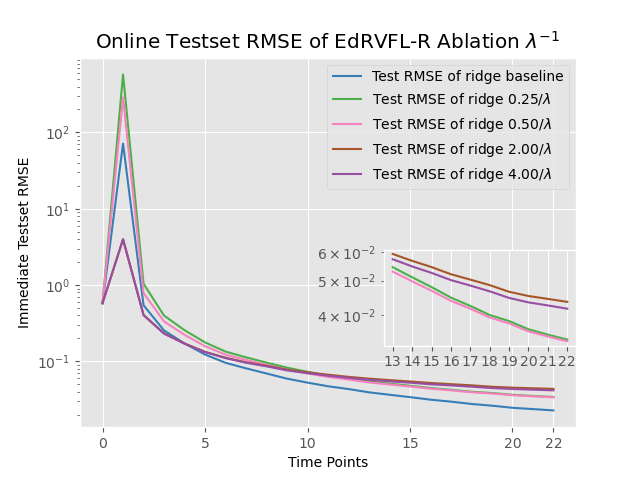}}
  \hfil
\subfigure[]{\includegraphics[width=0.32\textwidth]{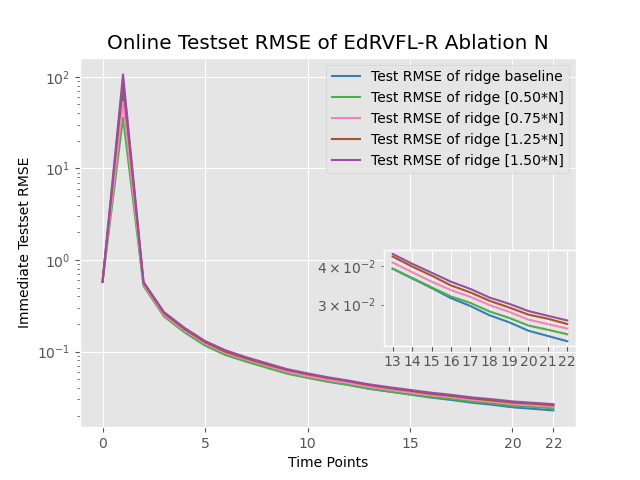}}
  \hfil
\subfigure[]{\includegraphics[width=0.32\textwidth]{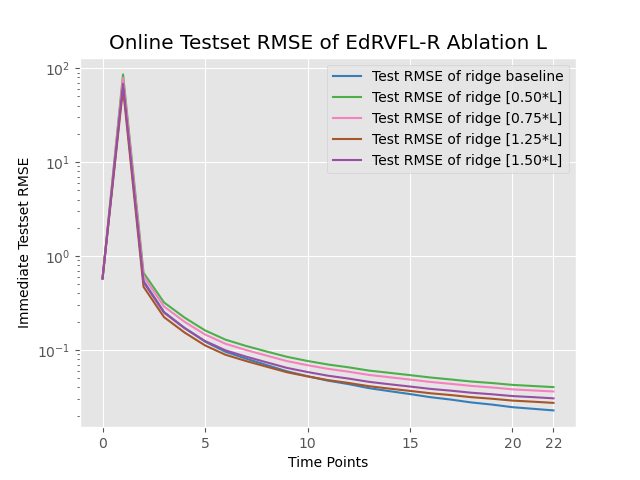}}
  \hfil

\vspace{-1.0em}
\subfigure[]{\includegraphics[width=0.32\textwidth]{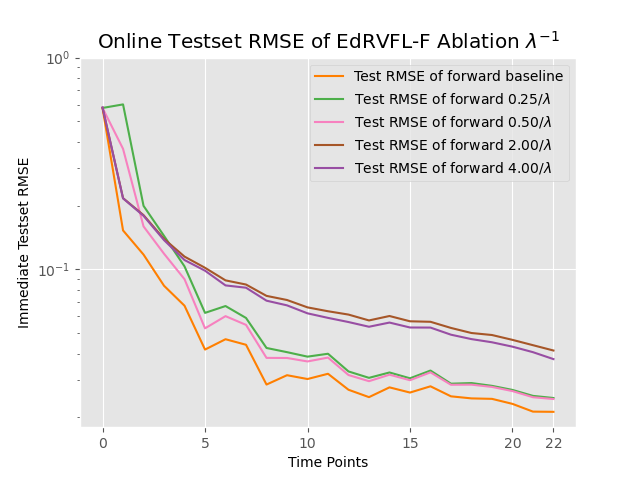}}
  \hfil
\subfigure[]{\includegraphics[width=0.32\textwidth]{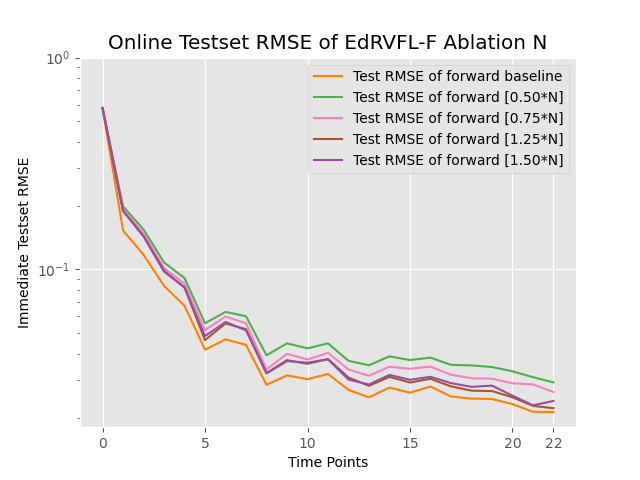}}
  \hfil
\subfigure[]{\includegraphics[width=0.32\textwidth]{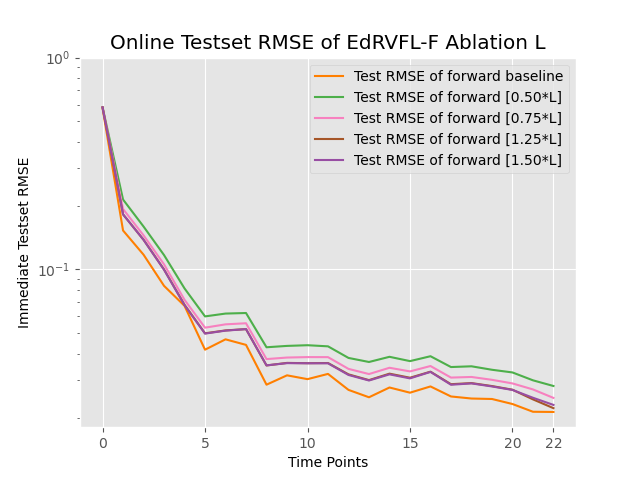}}
  \hfil

\centering
\caption{Ablation comparative experiments of IOL processes for edRVFL-R/F on regression task. Baselines' setups are in Table \ref{table 5}, and $[ \cdot ]$ denotes rounding operation. For edRVFL-R/F with varied setups, performance variations during IOL processes can be displayed by immediate testset RMSE over time. Some figures enlarge local X-axis in the inlaid subfigures.}
\label{fig 5}
\end{figure*}

For the guidance of structural fine-tuning, ablation experiments in Fig. \ref{fig 5} reveal the influence of main hyper-parameters, namely the $N$, $L$, $\lambda $, on the performance of IOL processes of edRVFL-R/F. Additional four configuration setups were tested in each subplot to contrast to the baselines. It can be concluded that the performance of edRVFL-R/F is greatly affected by $\lambda $ and $L$. Initial outlier peaks of edRVFL-R always linger in any case configuration, and oscillations exist in the process of edRVFL-F IOL because of the extra -F penalty.

\subsection{Comparisons of performance on classification tasks}
To further verify the IOL frameworks and the superiority of employing forward in batch stream scenarios, experiments based on online networks were performed on 25 classification tasks of public UCI datasets. These classification tasks are listed in Table \ref{table 6}. As before, main steps and experimental environment were still consistent here. Each dataset was separated into 5-folds and networks adjusted by SMAC3 from configuration space (see Table \ref{table 2}) were cross-validated on folds using two random seeds in one trial. 1.00 minus averaged classification accuracy of repeated validations was used to represent the incumbent loss, which reflected the accuracy gap to the full score. Early stopping was determined on smoothed learning curves of 80 epochs of validation set. Result analysis and additional ablation testing were followed afterward.
\begin{table}[!htbp]
\renewcommand\arraystretch{1}
\scriptsize
\centering
\caption[]{\\ }
\caption*{{Description of classification datasets}}
\begin{tabular}{cccc}
\toprule
$\text{Dataset}$  &
$\text{Size}$&  
$\text{Features}$&
$\text{Classes}$
 \\
\midrule
         abalone &  4177 &  8 &   3 \\
     balance scale &   625 &  4 &   3 \\
     breast cancer &   683 &  9 &   2 \\
   credit approval &   690 & 14 &   2 \\
             glass &   214 &  9 &   7 \\
 haberman survival &   306 &  3 &   2 \\
image segmentation &  2310 &  9 &   7 \\
        ionosphere &   351 & 34 &   2 \\
              iris &   150 &  4 &   3 \\
           letters & 20000 & 16 &  26 \\
       page blocks &  5473 & 10 &   5 \\
         pendigits & 10992 & 16 &  10 \\
              pima &   768 &  8 &   2 \\
      plant margin &  1600 & 64 & 100 \\
            raisin &   900 &  7 &   2 \\
             seeds &   210 &  7 &   3 \\
             sonar &   208 & 60 &   2 \\
          spambase &  4601 & 57 &   2 \\
       transfusion &   748 &  4 &   2 \\
          waveform &  5000 &  4 &   3 \\
              wine &   178 &  4 &   3 \\
  wine quality red &  1599 & 11 &   6 \\
wine quality white &  4898 & 11 &   7 \\
          wireless &  2000 &  7 &   4 \\
             yeast &  1484 &  8 &  10 \\
 \bottomrule
\end{tabular}
\label{table 6}
\end{table}

Experimental results of the above 8 methods on 25 classification tasks are listed in Table \ref{table 7}, and for each dataset, the highest accuracy values of TR and TE are highlighted in \textbf{bold}. Based on the statistics, it can be summarized that edRVFL-R/F outperforms other networks on 18 tasks and the average of all datasets, and edRVFL-F surpasses edRVFL-R in the accuracy of 19 datasets in IOL scenarios. The OBLS shows slight advantage over DOS-ELM referring to the average accuracy and, OSNN obtains the best scores on some online streams from smaller datasets. As before, the Wilcoxon testing is carried out for the classification tasks in Table \ref{table 8} and reveals that adopting -F enhances performance on classifications contrasted to -R in IOL with relevant p-value ${\text{ = 0}}{\text{.0091\textless 0}}{\text{.05}}$. As a result, edRVFL-F is suggested to be employed instead of using -R in IOL on classification tasks.

\begin{table}[h]
\renewcommand\arraystretch{1}
\scriptsize
\centering
\caption[]{\\ }
\caption*{Wilcoxon testing matrix of network performance on classification}
\begin{tabular}{*{9}{c@{\hspace{0.2em}}}}
\toprule
$\text{Model}$  &
$\text{OLRVFL}$&
$\text{pRVFL}$&
$\text{DOS-ELM}$&
$\text{OS-ELM}$&
$\text{OBLS}$&
$\text{OSNN}$&
$\text{edRVFL-R}$&
$\text{edRVFL-F}$
 \\
\midrule
OLRVFL&-     & 7.55e-5 & 0.3467    & 0.0161   & 0.0257   &0.0952   &1.23e-6 &1.64e-8 \\
pRVFL&7.55e-5 & -         & 2.92e-6 & 0.0204   & 4.22e-8   &2.00e-6 &2.29e-9 &1.33e-9 \\
DOS-ELM&0.3467    & 2.92e-6 & -     & 0.0006 & 0.2003   &0.4263    &2.55e-5 &5.86e-8 \\
OS-ELM&0.0161   & 0.0204  & 0.0006 & -      & 3.70e-6 & 0.0003 &8.28e-9 &1.60e-9 \\
OBLS&0.0257   & 4.22e-8   & 0.2003    & 3.70e-6 & -   &0.8084    &0.0002 &4.32e-7  \\
OSNN&0.0952   & 2.00e-6 & 0.4263    & 0.0003 & 0.8084    &-    &0.0021  &1.45e-5 \\
edRVFL-R&1.23e-6 & 2.29e-9 & 2.55e-5 & 8.28e-9 & 0.0002 &0.0021  &-      &0.0091 \\
edRVFL-F&1.64e-8 & 1.33e-9 & 5.86e-8 & 1.60e-9 & 4.32e-7  &1.45e-5 &0.0091  &- \\
 \bottomrule
\end{tabular}
\label{table 8}
\end{table}

\indent To further study the advantages of -F, SMAC3 configurations and detailed comparative tests of edRVFL-R/F in IOL processes were presented here based on online batch stream from \textit{Letters}. We still maintained the same structures of edRVFL-R/F to be optimized synchronously by SMAC3 for fair contrasts, and incumbent loss was set to 1.00 minus the average validation accuracy of two models in repeated trials. The baseline parameters and partial immediate accuracy during IOL are listed in Table \ref{table 9}. Note here that online immediate accuracy of testset included types of entire ensemble network and respective learners (each layer) for further studies from two views, where the values are recorded in Table \ref{table 9} and displayed by several solid curves and boxplots in Fig. \ref{fig 6}.

\begin{table}[!htbp]
\renewcommand\arraystretch{1}
\scriptsize
\centering
\caption[]{\\ }
\caption*{Baseline configurations and online testset accuracy of \textit{Letters}}
\begin{tabular}{*{7}{c@{\hspace{0.6em}}}}
\toprule
$\text{Hp of}$  &
\multirow{2}{*} {$N$}&  
\multirow{2}{*} {$L$}&
\multirow{2}{*} {${\log _2}({\lambda ^{ - 1}})$}&
\multirow{2}{*} {$b$}&
\multirow{2}{*} {$g$}&
$\text{Random}$
 \\
 $\text{baseline}$  &
&  
&
&
&
&\text{weights}\\
\midrule
 $\text{Value}$  &
720&  
3&
5&
0.030&
\text{sigmoid}&$ N(0,1)$\\
\toprule
Time Points &      0 &      8 &     16 &     24 &     32 &     34 \\
\midrule
edRVFL-R ACC (ensem.) & 0.0394 &  0.4772 &  0.7326 &  0.8422 &  0.8926 &  0.9008 \\
edRVFL-F ACC (ensem.)  & 0.0394 &  0.8866 &  0.9120 &  0.9180 & 0.9256 & 0.9262 \\
edRVFL-R ACC (avr.) & 0.0394 &  0.5413 &  0.7437 &  0.8207 & 0.8651 & 0.8708 \\
edRVFL-F ACC (avr.) & 0.0394 &  0.8788 &  0.9055 &  0.9135 & 0.91873 & 0.91867 \\
 \bottomrule
\end{tabular}
\captionsetup{font=scriptsize} \captionsetup{singlelinecheck=false,justification=raggedright}
\caption*{Note the ensem. means ensemble strategy, avr. means average of multiple learners.}
\label{table 9}
\end{table}

\begin{table*}[!htbp]
\renewcommand\arraystretch{1}
\scriptsize
\centering
\caption[]{\\ }
\caption*{Accuracy performance of online networks on classification tasks}
\begin{tabular}
{*{17}{c@{\hspace{1.0em}}}}
\hline
\toprule
\multirow{2}{*} {Dataset}&
\multicolumn{2}{c@{\hspace{1.6em}}}{OLRVFL} &     \multicolumn{2}{c@{\hspace{1.6em}}}{pRVFL} &   \multicolumn{2}{c@{\hspace{1.6em}}}{DOS-ELM} &    \multicolumn{2}{c@{\hspace{1.6em}}}{OS-ELM} &      \multicolumn{2}{c@{\hspace{1.8em}}}{OBLS} &
\multicolumn{2}{c@{\hspace{1.6em}}}{OSNN}&
\multicolumn{2}{c@{\hspace{1.6em}}}{edRVFL-R}&
\multicolumn{2}{c@{\hspace{1.8em}}}{edRVFL-F}   \\

 &       TR &         TE &       TR &         TE &      TR &         TE &      TR &         TE &      TR &          TE &      TR &          TE &      TR &          TE &      TR &          TE \\
\midrule               
abalone &     0.6822 &  0.6465 &  0.7503 &  0.5909 &  \textbf{0.8805} &  0.6363 &  0.7484 &  0.6297 &  0.6455 &  0.6336 &  0.6735 &  0.6492  &0.6781 &  0.6715 & 0.6817 &  \textbf{0.6733}\\

balance scale &     0.9484 &  0.8505 &  0.8187 &  0.6603 &  0.8462 &  0.8333 &  0.9310 &  0.7372 &  \textbf{1.0000} &  0.8525 &0.9829 &  0.8862  &0.9808 &  0.8846 & \textbf{1.0000} &  \textbf{0.9071}  \\

breast cancer &     0.9752 &  0.9638 &  0.9371 &  0.8938 &  0.9824 &  0.9307 &  0.9763 &  0.9220 &  0.9849 &  0.9689 &   \textbf{0.9912} &  0.9667  &0.9845 &  \textbf{0.9714} &0.9845 &  0.9710\\

credit approval &     0.8320 &  0.7863 &  0.8546 &  0.7848 &  0.9058 &  0.8459 &  0.8183 &  0.7965 &  \textbf{0.9435} &  \textbf{0.8775} &  0.8629 &  0.8546 &  0.8639 &  0.8648 &0.8667 &  0.8750\\

glass &     0.9699 &  0.6603 &  0.8633 &  0.6226 &  0.8222 &  0.6179 &  0.7577 &  0.4481 &  0.8015 &  0.6368 &0.8590 &  0.5823  &\textbf{1.0000} &  0.6986 &   \textbf{1.0000} &  \textbf{0.7033}\\

haberman survival &     0.8826 &  0.6711 &  \textbf{0.9869} &  0.7171 &  0.7513 &  0.7269 &  0.8474 &  0.6513 &  0.8826 &  0.6908 &0.9783 &  \textbf{0.7540}  &   0.9104 &  0.7368 &0.9044 &  0.7532\\

image segmentation &     0.8381 &  0.6567 &  0.6993 &  0.6715 &  0.8635 &  0.8023 &  0.6552 &  0.4856 &  0.9519 &  0.8457 &0.9356 &  0.8855  &0.9540 &  0.9008 &   \textbf{0.9597} &  \textbf{0.9014}\\

ionosphere &     0.8807 &  0.8920 &  0.9429 &  0.9147 &  \textbf{0.9582} &  0.9176 &  0.8772 &  0.8722 &  0.9543 &  0.9261 &0.9306 &  0.9118  &0.9429 &  0.9261 &   0.9467 &  \textbf{0.9432}\\

iris &     \textbf{1.0000} &  0.8784 &  0.9292 &  0.8986 &  0.9073 &  0.9054 &  0.9888 &  0.9459 &  \textbf{1.0000} &  0.9729 &   \textbf{1.0000} &  \textbf{0.9972}  &\textbf{1.0000} &  0.9932 &0.9977 &  0.9918\\

letters &     0.8847 &  0.8351 &  0.6948 &  0.6943 &  0.8026 &  0.7893 &  0.8154 &  0.7986 &  0.9410 &  0.9157 &  0.9284 &  0.9059  &0.9389 &  0.9104 &\textbf{0.9436} &  \textbf{0.9288} \\

page blocks &      0.9258 &  0.9012 &  0.6485 &  0.5758 &  0.8275 &  0.7662 &  0.6670 &  0.6692 &  0.8996 &  0.8623 &0.9418 &  0.9401 &0.9734 &  0.9638 &   \textbf{0.9815} &  \textbf{0.9756}\\

pendigits &     0.9870 &  0.9481 &  0.9379 &  0.8790 &  0.9808 &  0.9518 &  0.9496 &  0.9206 &  0.9674 &  0.9213 &0.9346 &  0.9371  &0.9837 &  0.9510 &   \textbf{0.9981} &  \textbf{0.9702}\\

pima &      \textbf{0.9913} &  0.7408 &  0.7093 &  0.6809 &  0.7982 &  0.7604 &  0.8413 &  0.7292 &  0.7846 &  0.7552 &0.7938 &  0.7253 &0.8386 &  0.7369 &   0.7899 &  \textbf{0.7695}\\

plant margin &     0.9132 &  0.5831 &  0.6573 &  0.5892 &  \textbf{1.0000} &  0.7550 &  \textbf{1.0000} &  0.7581 &  0.9503 &  0.6685 &0.8934 &  0.7390  &   \textbf{1.0000} &  \textbf{0.7743} &0.9350 &  0.7606\\

raisin &     0.8519 &  0.8527 &  0.6032 &  0.6277 &  \textbf{0.8857} &  0.8742 &  0.7758 &  0.7893 &  0.8794 &  0.8337 &0.8525 &  0.8126  &0.8703 &  \textbf{0.8834} &   0.8672 &  0.8711\\

seeds &     0.9423 &  0.9086 &  0.9231 &  0.8865 &  0.9916 &  0.8654 &  0.9493 &  0.6827 &  0.9746 &  0.9013 &   0.9523 &  \textbf{0.9338} & \textbf{1.0000} &  0.9134 &\textbf{1.0000} &  0.9286\\

sonar &      0.9007 &  0.7296 &  0.7900 &  0.6432 &  0.8933 &  0.7592 &  0.8541 &  0.6825 &  \textbf{1.0000} &  \textbf{0.7701} &0.7978 &  0.7462 &0.8735 &  0.7456 &   0.8652 &  0.7636\\

spambase &      0.9316 &  0.9130 &  0.7322 &  0.4139 &  \textbf{0.9833} &  0.8930 &  \textbf{0.9833} &  0.8935 &  0.9392 &  0.9195 &0.9137 &  0.9076 &   0.9293 &  0.9236 &0.9336 &  \textbf{0.9245}\\

transfusion &      0.8076 &  \textbf{0.7634} &  0.8344 &  0.7276 &  0.8323 &  0.7567 &  0.8201 &  0.7007 &  \textbf{0.8517} &  0.7407 &0.7728 &  0.6972 &   0.8341 &  0.7386 &0.8228 &  0.7410\\

waveform &     0.8872 &  0.7774 &  0.5024 &  0.4672 &  0.8487 &  0.8304 &  0.8606 &  0.8256 &  \textbf{0.9565} &  0.8498 &0.7941 &  0.7306  &0.8891 &  0.8678 &   0.8901 &  \textbf{0.8784}\\

wine &     \textbf{1.0000} &  0.9375 &  0.9776 &  0.9659 &  0.9925 &  0.9886 &  \textbf{1.0000} &  0.9829 &  \textbf{1.0000} &  0.9899 & 0.9998 &  0.9998  &  \textbf{1.0000} &  \textbf{1.0000} &\textbf{1.0000} &  \textbf{1.0000}\\

wine quality red &     0.7061 &  0.6168 &  0.3920 &  0.3843 &  0.6605 &  0.6106 &  0.5947 &  0.5825 &  0.6538 &  0.6124 &0.6074 &  0.5925 &0.9691 &  0.6406 &    \textbf{0.9708} &  \textbf{0.6419}\\

wine quality white &     0.5499 &  0.5226 &  0.7923 &  0.4590 &  0.5605 &  0.4350 &  0.7150 &  0.5460 &  0.5441 &  0.5261 &  0.6343 &  \textbf{0.6074}  & \textbf{0.8965} &  0.5856 &0.8806 &  0.5963\\

wireless &      0.9804 &  0.9613 &  0.9032 &  0.8743 &  0.9915 &  0.9800 &  0.9716 &  0.9736 &  0.9946 &  0.9735 &0.9550 &  0.9314 &0.9979 &  0.9833 &   \textbf{0.9998} &  \textbf{0.9851}\\

yeast &      0.7492 &  0.5696 &  0.3025 &  0.2736 &  \textbf{0.8393} &  0.5856 &  0.8140 &  0.4522 &  0.8221 &  0.5249 &0.6033 &  0.6063 &0.6803 &  \textbf{0.6352} &   0.6774 &  0.6312\\

Average &     0.8807 &  0.7827 &  0.7673 &  0.6759 &  0.8722 &  0.7927 &  0.8485 &  0.7390 &  0.8929 &  0.8068 & 0.8636 &0.8120  & \textbf{0.9196} &  0.8361 & 0.9159 &  \textbf{0.8434}\\
\bottomrule
\end{tabular}
\label{table 7}
\end{table*}

\begin{figure}[ht]
  \centering
  \includegraphics[width=\columnwidth, trim=25mm 0mm 55mm 5mm, clip]{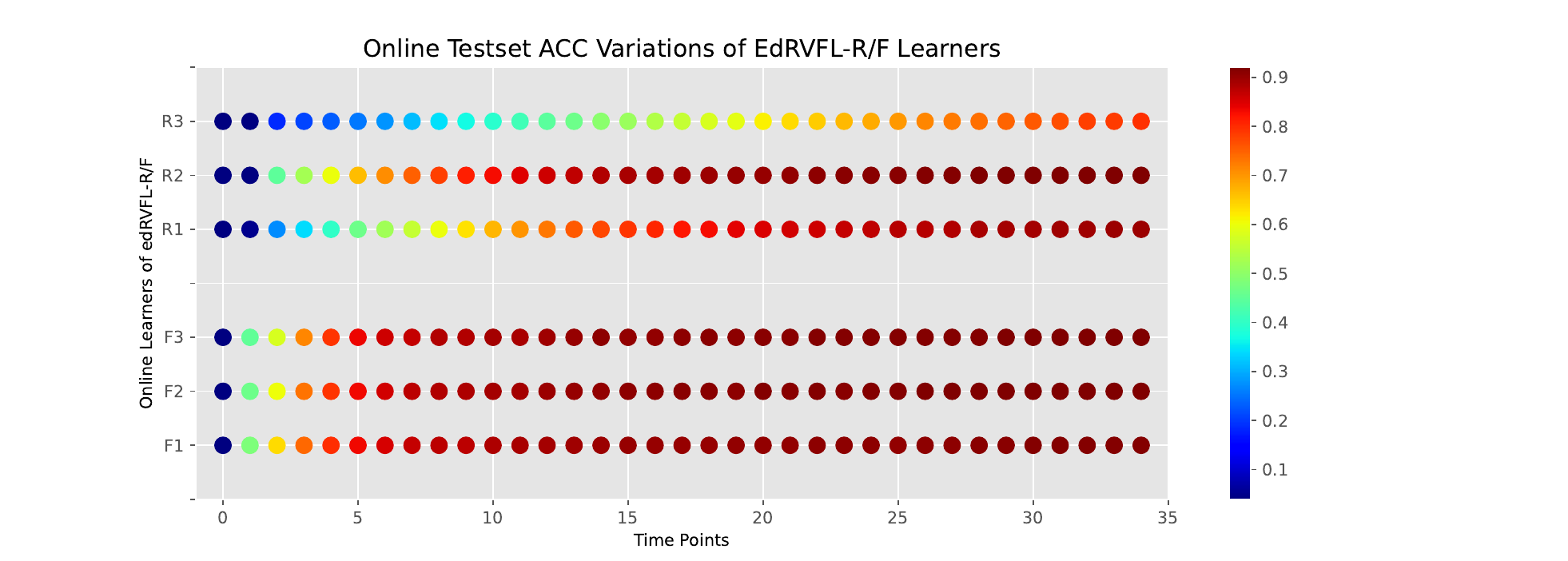}
  \caption{Status variations of online sub-learners of edRVFL-R/F in IOL process. Color bar denotes accuracy, and filled circles denote online learners at varied time points. X-axis: time points; Y-axis: learner index of edRVFL-R/F. Configurations and values are provided in Fig. \ref{fig 6} and Table \ref{table 9}. It can be found that the accuracy of edRVFL-F learners rises faster and higher than the learners of edRVFL-R during IOL. The delay in learning speed and ability leads to efficiency degradation in follow-on learners, ultimately resulting in the collapse of overall performance.}
  \label{fig 8}
\end{figure}

From Table \ref{table 9} and Fig. \ref{fig 6}, it can be concluded that testset accuracy of two IOL processes increases with more stream batches being trained while -F promotes edRVFL online updating faster and outperforms -R, regardless of ensemble or clustered learners angle, which also implies the edRVFL-F has lower cumulative and immediate regrets in IOL process. The -F utilizes proactive learning to eliminate some regrets, which can also be explained as the guiding role in IOL framework. Moreover, immediate regrets of edRVFL-R/F in IOL processes exhibit a gap that diminishes over time, and it is necessary to ensure enough updating times for networks with -R to prevent unqualified performance. The accuracy of edRVFL-R grows sluggishly at the start and even is lower than the median accuracy of interior multiple learners in the first 30 time steps, which is affected by some poor, slow, and even wrongly trained sub-learners inside it. The accuracy of edRVFL-F in IOL rises much faster and is always higher than the median accuracy of interior multiple learners at all time points. This erratic phenomenon of using -R arises from inadequate learning efficiency of some certain learners, leading the performance of subsequent learners or even the entire system of edRVFL-R to collapse. We illustrate this phenomenon in Fig. \ref{fig 8}. Each sub-learner inside edRVFL-F benefits from precognition of -F and is guided by future prophetic estimates in IOL but not every one inside edRVFL-R get the updated knowledge right according to the box shapes in Fig. \ref{fig 6}. From this view, edRVFL-F is more robust during IOL process. The accuracy of edRVFL-R/F is notably increased by the ensemble strategy in the end and edRVFL-F is still superior to -R as the boxplots and numerical results show.
\begin{figure}[!htbp]
\centering
\includegraphics[width=\columnwidth,trim=0mm 0mm 10mm 1mm, clip]{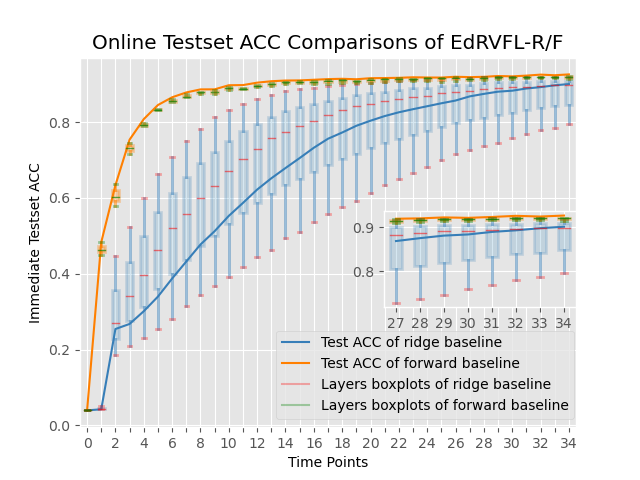}
  \caption{Testset accuracy variation curves of edRVFL-R/F baselines on classification task during IOL processes. Baselines' setup and values can be referred to Table \ref{table 9}, and performance can be compared by testset accuracy over time. Solid lines: immediate ensemble accuracy of baselines. Boxplots: immediate accuracy statistics of interior multiple layers (sub-learners). X-axis is locally enlarged in the inlaid subfigure.}
  \label{fig 6}
\end{figure}

edRVFL-R/F were incrementally updated on non-stationary streams during IOL, and had \textit{no retrospective retraining} of previous data and responded rapidly to requests. The immediate accuracy curves observed in Fig. \ref{fig 6} show an overall increasing trend, indicating there was no obvious \textit{catastrophic forgetting}, and had good suppression of  \textit{distribution drift}.

\begin{figure*}[!htbp]
\centering
\subfigure[]{\includegraphics[width=0.32\textwidth]{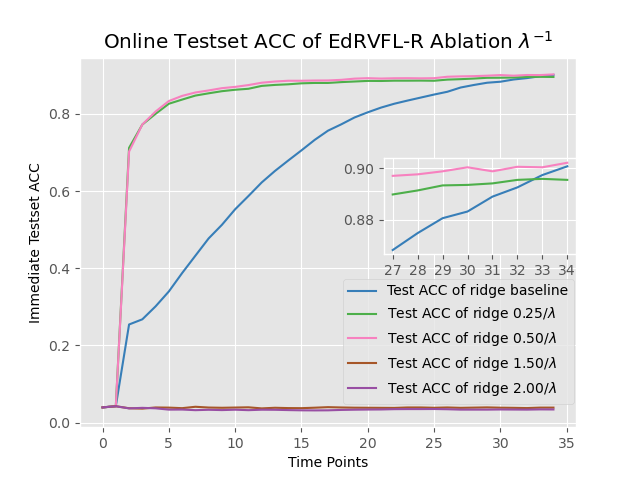}}
  \hfil
\subfigure[]{\includegraphics[width=0.32\textwidth]{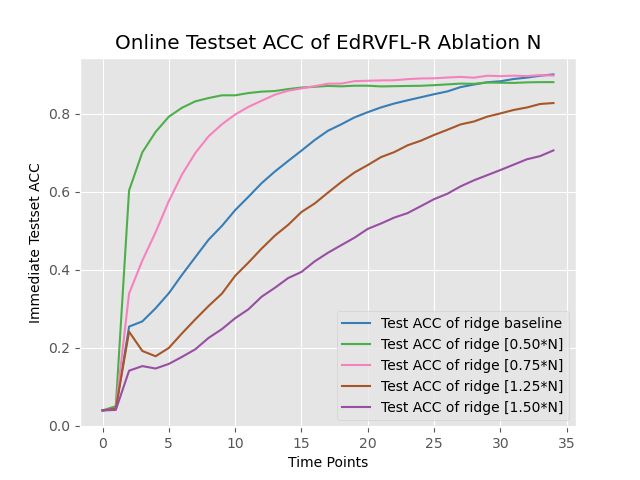}}
  \hfil
\subfigure[]{\includegraphics[width=0.32\textwidth]{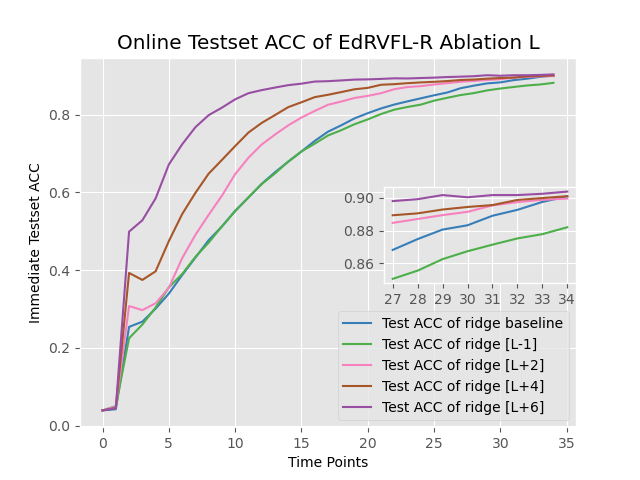}}
  \hfil

\vspace{-1.0em}
\subfigure[]{\includegraphics[width=0.32\textwidth]{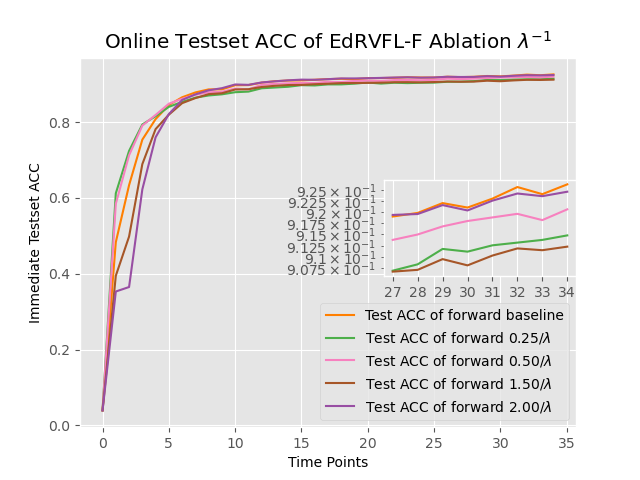}}
  \hfil
\subfigure[]{\includegraphics[width=0.32\textwidth]{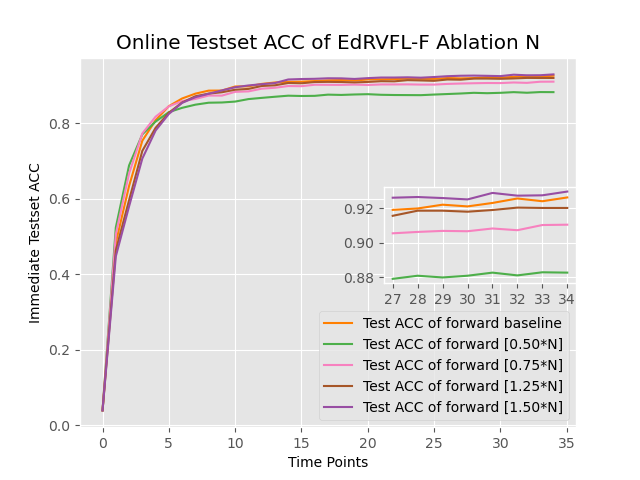}}
  \hfil
\subfigure[]{\includegraphics[width=0.32\textwidth]{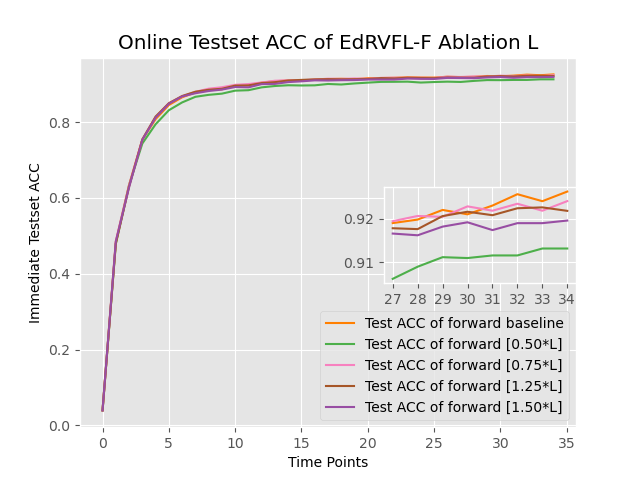}}
  \hfil

\centering
\caption{Ablation comparative experiments of  IOL processes for edRVFL-R/F on classification task. Baselines' setups are in Table \ref{table 9}. For edRVFL-R/F with varied setups, performance variations during IOL processes can be displayed by immediate testset accuracy over online batches. Some figures enlarge local X-axis in the inlaid subfigures.}
\label{fig 7}
\end{figure*}

For ablation experiments in Fig. \ref{fig 7}, it can be concluded that the performance of edRVFL-R is highly susceptible to inappropriate parameters. Similar to the regression task, there are initial local delays in the IOL process if using unsuitable setups, which can cause stunting in network online learning. On the contrary, IOL process of edRVFL-F is relatively stable and more robust on varying parameters in classification tasks.
\subsection{Studies of performance on large classification datasets}
To further simulate big data environments in real online learning scenarios and explain our advocacy of using -F in IOL process, we conducted experiments on 6 large \textit{openml} and \textit{kaggle} classification datasets. The datasets are described in Table \ref{table 10}. Each dataset was partitioned 20 percent randomly into the testset, with the remaining portion divided 10-fold, one of which was designated as the validation set for validating incumbent scores in SMAC3 optimization and early-stop decision of SNN. The early stopping was determined on smoothed learning curves of 100 epochs of the validation set. Classification accuracy was used as the evaluation criterion. For huge data volume, we restricted some SMAC3 configuration spaces to Table \ref{table 11}, and 1.00 minus averaged classification accuracy of repeated validations was used to represent the incumbent loss. Result analysis and ablation testing were followed afterward.
\begin{table}[!htbp]
\renewcommand\arraystretch{1}
\scriptsize
\centering
\caption[]{\\ }
\caption*{{Description of large \textit{openml} datasets}}
\begin{tabular}{ccccc}
\toprule
$\text{Dataset}$ &
$\text{Size}$&  
$\text{Features}$&
$\text{Classes}$&
$\text{ID}$
 \\
\midrule
BNG(page-blocks) &  295,245 & 10& 5 &259\\
BNG(solar-flare) &  1,000,000 & 12& 3 &1179\\
click prediction &  399,482 & 11& 2 &1219\\     creditcard &  284,807 & 30& 2 &1597\\  
miniboone &  130,064 & 50& 2 &41150\\ 
poker hand &  1,025,010 & 10& 10 &1595\\
 \bottomrule
\end{tabular}
\label{table 10}
\end{table}

The experimental results of the above 8 methods on 6 classification tasks are listed in Table \ref{table 12}, and the highest accuracy values of testing accuracy are highlighted in \textbf{bold} for each dataset. The standard deviations reflect result accuracy fluctuations in cross-validations, and the average running time of each model's single IOL process in SMAC3 configuration is recorded in Table \ref{table 13}. From the results, edRVFL-F surpasses other networks in testing accuracy on 4 datasets and the average value. The edRVFL-R gets similar results to edRVFL-F on some large datasets because multiple online updates brought the solutions of these two methods close. The SNN also demonstrates good performance while it consumes much time and memory for old data retrievals. If the dataset for training was reduced by half, testing accuracy of SNN would quickly drop to 0.71 while the edRVFL-R/F still realized around 0.85. Other RVFL or ELM-based networks tend to have more hidden nodes or maps in SMAC3 optimization, however, this would also cause resource consumption of large matrix and late performance degradation in IOL process.

\begin{table}[!htbp]
\renewcommand\arraystretch{1}
\scriptsize
\centering
\caption[]{\\ }
\caption*{SMAC3 configuration space for large datasets}
\begin{tabular}{cccc}
\toprule
$\text{Hp}$  &
$\text{Description}$&  
$\text{Range}$&
$\text{Type}$
 \\
\midrule
trial&SMAC3 trial times&100 &integer\\
$N$ &neuron amount per layer&[4,1024]&integer\\
$L$&number of layers&[2, 8]&integer\\
${\log _2}({\lambda ^{ - 1}})$&regularization factor&$[ - 6,8]$&integer\\
$b$&batch data proportion&[0.001, 0.01]&float\\
$g$&activation function&['sigmoid','ReLU'…]
&class\\
 \bottomrule
\end{tabular}
\label{table 11}
\end{table}

\begin{figure}[!htbp]
\centering
\includegraphics[width=\columnwidth,trim=0mm 0mm 10mm 1mm, clip]{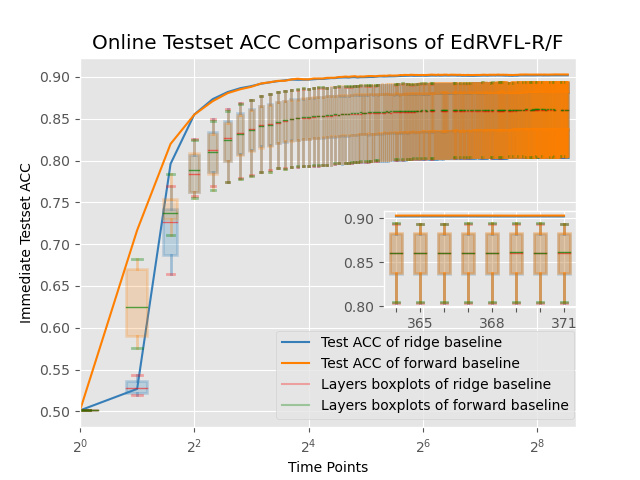}
  \caption{Testset accuracy variation curves of edRVFL-R/F baselines on \textit{poker hand} task in IOL processes. Baselines' setups and values can be referred to Table \ref{table 14}, and performance is shown by accuracy over time. Solid lines: immediate ensemble accuracy of baselines. Boxplots: immediate accuracy statistics of clustered learners. X-axis is locally enlarged in the inlaid subfigure.}
  \label{fig 9}
\end{figure}

To further show the advantages of employing -F in IOL, we conducted comparative experiments on challenging \textit{poker hand} dataset. Note the dataset features were sorted for better results. Structures of edRVFL-R/F were maintained the same and optimized by SMAC3 synchronously for equal comparisons. Incumbent loss was set to 1.00 minus the average validation accuracy of repeated trials of two models. The baseline parameters and partial immediate accuracy during IOL are listed in Table \ref{table 14}. Note online immediate accuracy of testset included two views: ensemble results of the entire network and clustered learners' accuracy. Their values are recorded in Table \ref{table 14} and displayed by several solid lines and boxplots in Fig. \ref{fig 9}. 

\begin{table}[!htbp]
\renewcommand\arraystretch{1}
\scriptsize
\centering
\caption[]{\\ }
\caption*{Baseline configurations and online testset accuracy of \textit{poker hand}}
\begin{tabular}{*{7}{c@{\hspace{0.6em}}}}
\toprule
$\text{Hp of}$  &
\multirow{2}{*} {$N$}&  
\multirow{2}{*} {$L$}&
\multirow{2}{*} {${\log _2}({\lambda ^{ - 1}})$}&
\multirow{2}{*} {$b$}&
\multirow{2}{*} {$g$}&
$\text{Random}$
 \\
 $\text{baseline}$  &
&  
&
&
&
&\text{weights}\\
\midrule
 $\text{Value}$  &
989&  
5&
-2&
0.0027&
\text{ReLU}&Xavier\\
\toprule
Time Points &   ${2^0}$ &${2^2}$&${2^4}$ &    ${2^6}$&${2^8}$ &371 \\
\midrule
edRVFL-R ACC (ensem.) & 0.5012 &  0.8549 &  0.8971 & 0.9025  & 0.9027  & 0.9028 \\
edRVFL-F ACC (ensem.)  & 0.5012 &  0.8558 &  0.8977 & 0.9024 & 0.9029 &0.9030  \\
edRVFL-R ACC (avr.) & 0.5012 &0.7881 & 0.8463  &0.85405   & 0.8560 & 0.8581 \\
edRVFL-F ACC (avr.) & 0.5012 & 0.7883  &0.8470   & 0.85407  &0.8561  & 0.8582 \\
 \bottomrule
\end{tabular}
\captionsetup{font=scriptsize} \captionsetup{singlelinecheck=false,justification=raggedright}
\caption*{Note the ensem. means ensemble method, avr. means average of multiple learners.}
\label{table 14}
\end{table}

Table \ref{table 14} and Fig. \ref{fig 9} show that networks' performance and immediate testset accuracy improve as more stream batches are trained in IOL, which demonstrates the availability of our proposed IOL frameworks of edRVFL-R/F. From the ensemble and clustered learners' view, -F style enhances network updating speed at the start of IOL process, and accuracy curves of edRVFL-R/F gradually converge to similar values as enough batches are given, but the cumulative regrets of -F are still smaller compared to using -R.

\begin{figure}[!htbp]
\centering
\includegraphics[width=\columnwidth,trim=0mm 0mm 10mm 1mm, clip]{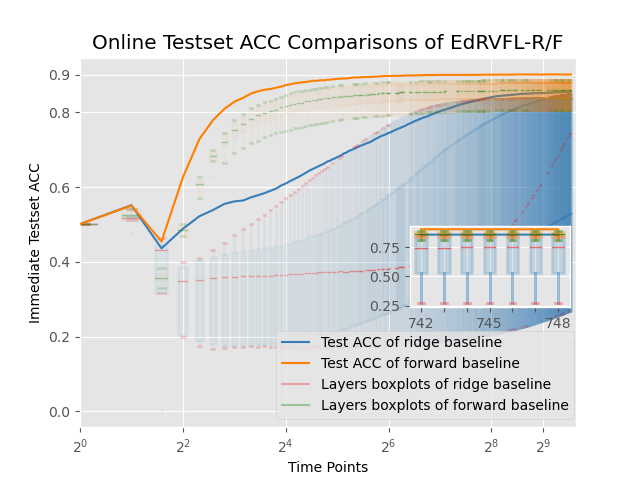}
  \caption{Testset accuracy variation curves of edRVFL-R/F with reduced $b$. The total update times are doubled. edRVFL-R is configured by SMAC3 and edRVFL-F uses the same parameters, but edRVFL-R performs worse than edRVFL-F. Solid lines: immediate ensemble accuracy of baselines. Boxplots: immediate accuracy statistics of clustered learners. X-axis is locally enlarged in the inlaid subfigure.}
  \label{fig 10}
\end{figure}

During SMAC3 optimization, we observed that edRVFL-F could be more stable than edRVFL-R given the same configurations. With the batch samples $b$ of Table \ref{table 14} halved, the accuracy variation curves of edRVFL-R/F in IOL are shown in Fig. \ref{fig 10}. The structure of edRVFL-R was configured by SMAC3 and edRVFL-F used the same one. In contrast to Fig. \ref{fig 9}, for edRVFL-F, the updating rate becomes slow and achieves expected results in ${2^6}$ steps because the samples were reduced in batches. However, the edRVFL-R evolves even tardily and gets worse final results. From the boxplots in Fig. \ref{fig 10}, we can find there are some sub-learners of low efficiency inside edRVFL-R which hamper the learning process and even result in system collapse in IOL.

\begin{figure}[!htbp]
\centering
\includegraphics[width=\columnwidth,trim=0mm 0mm 10mm 1mm, clip]{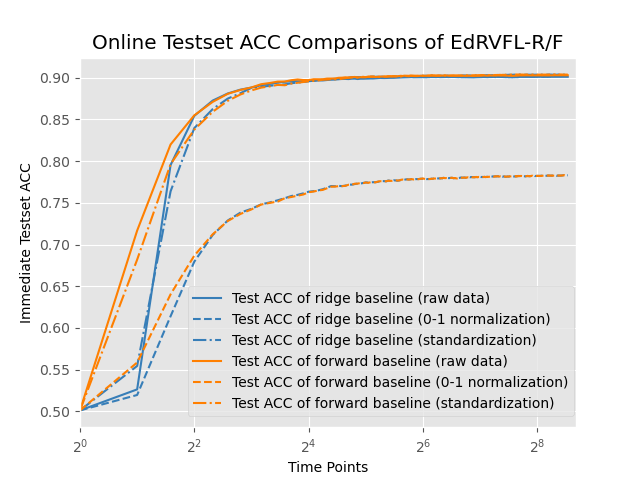}
  \caption{Testset accuracy variation curves of edRVFL-R/F with different normalization methods. It shows that the 0-1 normalization causes an adverse influence on accuracy.}
  \label{fig 11}
\end{figure}

\begin{figure*}[!htbp]
\centering
\subfigure[]{\includegraphics[width=0.32\textwidth]{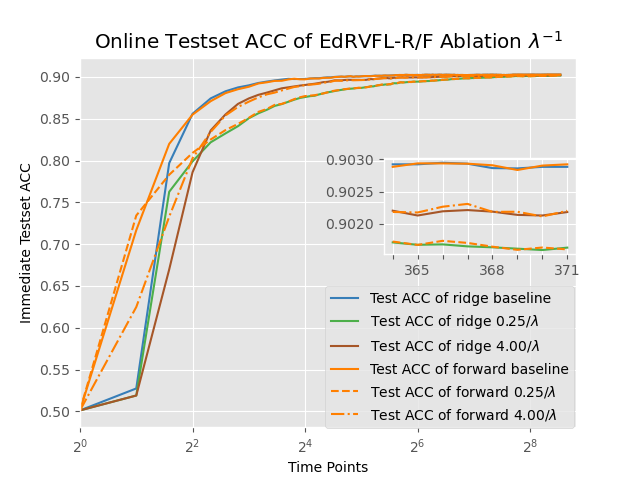}}
  \hfil
\subfigure[]{\includegraphics[width=0.32\textwidth]{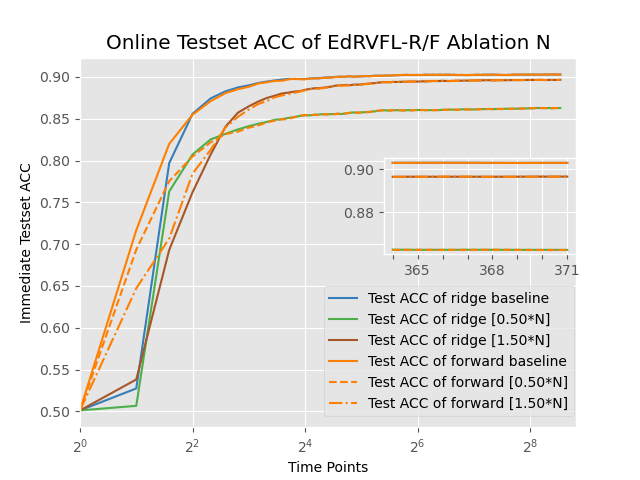}}
  \hfil
\subfigure[]{\includegraphics[width=0.32\textwidth]{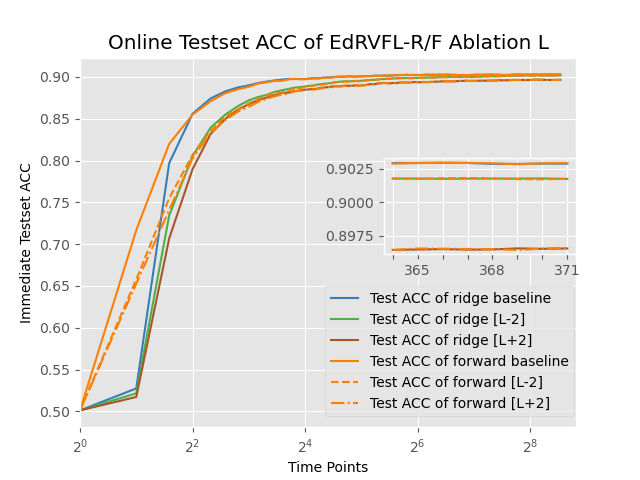}}
  \hfil

\centering
\caption{Ablation comparative experiments of edRVFL-R/F on \textit{poker hand} task.  Baselines' setups and values can be referred to Table \ref{table 14}. Different parameters affect the convergence rate and final accuracy of IOL processes of edRVFL-R/F, and they can converge to similar results given long enough steps. Some figures enlarge local X-axis in inlaid subfigures.}
\label{fig 12}
\end{figure*}

The ablation experiment on the normalization method is shown in Fig. \ref{fig 11}. The use of 0-1 normalization had a larger negative impact on the accuracy of edRVFL-R/F in IOL. The ablation experiments on the main hyper-parameter are shown in Fig. \ref{fig 12}. The baseline configurations are still the same as those of Table \ref{table 14}. In this task, three hyper-parameters have a great influence on the early IOL processes of edRVFL-R/F especially the rate of convergence, and $N$ would significantly affect their final performance. Given enough long time steps, the effects of inappropriate parameters on both models are similar, therefore, edRVFL-R/F maintain close accuracy under the same parameters during IOL processes.

\begin{table*}[!htbp]
\renewcommand\arraystretch{1}
\scriptsize
\centering
\caption[]{\\ }
\caption*{Accuracy performance of online networks on \textit{openml} classification tasks}
\begin{tabular}
{*{17}{c@{\hspace{1.0em}}}}
\hline
\toprule
\multirow{2}{*} {Dataset}
&
\multicolumn{2}{c@{\hspace{1.6em}}}{OLRVFL} &     \multicolumn{2}{c@{\hspace{1.6em}}}{pRVFL} &   \multicolumn{2}{c@{\hspace{1.6em}}}{DOS-ELM} &    \multicolumn{2}{c@{\hspace{1.6em}}}{OS-ELM} &      \multicolumn{2}{c@{\hspace{1.8em}}}{OBLS} & 
\multicolumn{2}{c@{\hspace{1.6em}}}{OSNN}&
\multicolumn{2}{c@{\hspace{1.6em}}}{edRVFL-R}&
\multicolumn{2}{c@{\hspace{1.8em}}}{edRVFL-F}   \\

 &       TE &         std.\%&       TE &         std.\%&       TE &         std.\%&       TE &         std.\%&       TE &         std.\%&       TE &         std.\%&       TE &         std.\%&       TE &         std.\%  \\
\midrule               
BNG(page-blocks) &     0.8703& 1.0358 &  0.8220 &2.5183  &   0.8592  &1.3395  &   0.8269 & 2.3701 & 0.8671 & 1.3024  &0.8769 & 2.5948&0.9118 &  0.2235   &\textbf{0.9152}& 0.0337 \\

BNG(solar-flare)&    0.9937 & 0.0059 & 0.9817 &0.0101 &    0.9947  &0.0062 &    0.9813&  0.0160 & 0.9895 & 0.0089  & 0.9954 & 0.0041   &0.9956&   0.0053 &\textbf{0.9969}& 0.0059\\     

click prediction&   0.8320&  0.2138 & 0.7643 &1.2384  &   0.8204  &0.5306  &   0.8204 & 2.0258 & 0.8219 & 0.5724  &0.8323 & 0.2019   & \textbf{0.8341}  & 0.2368 &0.8340 &0.1986 \\ 

creditcard&  0.9974&  0.0324&  0.9924&  0.0387 &   0.9857 & 0.1431 &    0.9869 & 0.1881 & 0.9803 & 0.1735&0.9989 & 0.0454      &0.9993 &  0.0059 &  \textbf{0.9995} &0.0014 \\

miniboone& 0.8754 & 0.5785 & 0.8663 & 1.5721  &  0.8239 & 2.0820 &    0.7858 & 1.9191 & 0.8759 & 1.3441 &0.8893 & 0.2341    &0.9131 &  0.2314  &  \textbf{0.9243}& 0.1959\\

poker hand&       0.5876&  5.2027 & 0.5595 & 2.0016 &   0.7368 & 1.4225 &    0.6800 & 3.0124&  0.7859&  1.8637  &\textbf{0.9660} & 2.0951   & 0.9056 &  0.2457  &0.9352 &0.1156\\

Average&0.8594&1.1782&0.8310&1.2299&0.8701&0.9207&0.8469&1.5886&0.8868&0.8775&0.9265&0.8626&0.9266&0.1581&\textbf{0.9342}&0.0919\\
\bottomrule
\end{tabular}
\label{table 12}
\end{table*}

\begin{table*}[!htbp]
\renewcommand\arraystretch{1}
\scriptsize
\centering
\caption[]{\\ }
\caption*{Average IOL computing time in SMAC3 configurations}
\begin{tabular}{ccccccccc}
\toprule
$\text{Dataset}$  &
$\text{OLRVFL}$&
$\text{pRVFL}$&
$\text{DOS-ELM}$&
$\text{OS-ELM}$&
$\text{OBLS}$&
$\text{OSNN}$&
$\text{edRVFL-R}$&
$\text{edRVFL-F}$
 \\
\midrule
BNG(page-blocks)&					1.7318&		1.6305&			1.5024&		1.1103&		1.5832&	3.3360&1.2380&1.3860 \\

BNG(solar-flare)& 			3.6927&		3.3023&			3.8930&        	1.7294&		4.1048&	12.2656		&4.0815&4.0269	\\

click prediction&		1.2104	&	0.5420	&		2.0034	&	0.5304	&	2.3041	&	3.9328		&0.7378	&0.6079 \\

creditcard&			2.4727	&	2.2727		&	1.3029	&	1.3173	&	0.6319	&2.4285	&0.5008	&0.4138	 \\

miniboone& 		0.2047	&	0.5981		&	1.0023	&	1.1658	&	1.2221&	0.9036		&	0.1968	&0.2134\\

poker hand&		2.1849	&	3.7501		&	3.7392	&	2.1440	&	7.3194	&13.9533		&5.0625	&4.9104	 \\

 \bottomrule
\end{tabular}
\captionsetup{font=scriptsize,margin=2.5cm} \captionsetup{singlelinecheck=false,justification=raggedright}
\caption*{Note values are counted in minutes. Server: AMD Ryzen 3990X with 4x NVIDIA RTX 3090.}
\label{table 13}
\end{table*}

\subsection{Analysis on existing challenges and online restrictions}
In this subsection, we provided a brief analysis of the availability of proposed -R/-F algorithms within IOL frameworks for the existing challenges and online restrictions, combining experimental results with theoretical proofs. 

\textit{Hysteretic non-incremental updating.} We presented studies on computational complexity. During the IOL processes of edRVFL-R/F, for $0 \leqslant t \leqslant T$ and learner cluster, complexity primarily depended on the recursive updates of learning rate $\eta _{l,t}^{\{ r,f\} } \in {\mathbb{R}^{(N+k) \cdot (N+k)}}$ using Sherman's equations, which resulted in time complexity $\mathcal{O}(L \cdot {b^3})$ at each data chunk in our algorithms. If batch had an excess of samples, it could be split into several sub-batches to feed, under the acceptable margin of error. In some typical implementations, the inversion calculation was based on matrix size of either $b \cdot b$ or $(NL + k) \cdot (NL + k)$, potentially causing resource-intensive or even out-of-memory under large data and limited resources. This problem was alleviated through batch decomposition and ensemble strategy in edRVFL. Moreover, the IOL processes did not involve data retrieval and iterative training (see Sec. IV B), and edRVFL-R/F achieved faster updates compared to fully-connected models with GD (Table \ref{table 13}). This could incrementally maintain an up-to-date online learner to rapidly answer queries with growing quality and less delays over time. 

\textit{Memory usage.} From the perspective of data access, for $0 \leqslant t \leqslant T$ and learner cluster, edRVFL-R only stored the current batch $({X_t},{Y_t})$ while edRVFL-F required more memory of $\mathcal{O}(Lb \cdot {(N + k)})$ complexity to accommodate $({X_t},{Y_t})$ and $X_{t +1}$ during IOL. We also studied space complexity. During the IOL processes of edRVFL-R/F, for $0 \leqslant t \leqslant T$ and learner cluster, complexity primarily depended on the recursive updates of learning rate $\eta _{l,t}^{\{ r,f\} } \in {\mathbb{R}^{(N+k) \cdot (N+k)}}$, which resulted in space complexity $\mathcal{O}(L \cdot {b^2})$ in IOL frameworks. Moreover, for online learning with past revisiting, memory consumption increased significantly over time while IOL processes allowed past data to be discarded and remained necessary memory capacity unchanged. 

\textit{Retrospective retraining.} Refer to Theorem \ref{theorem 2} and Theorem \ref{theorem 3}, the IOL processes of edRVFL-R/F did not need retrospective retraining.

\begin{figure}[!htbp]
\centering
\includegraphics[width=\columnwidth,trim=0mm 0mm 10mm 1mm, clip]{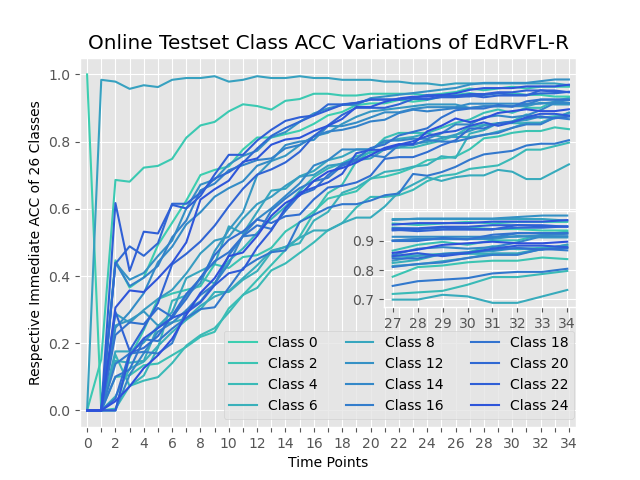}
  \caption{Testset accuracy variation curves of edRVFL-R on 26 classes. Gradient blue indicates different categories.}
  \label{fig 13}
\end{figure}

\begin{figure}[!htbp]
\centering
\includegraphics[width=\columnwidth,trim=0mm 0mm 10mm 1mm, clip]{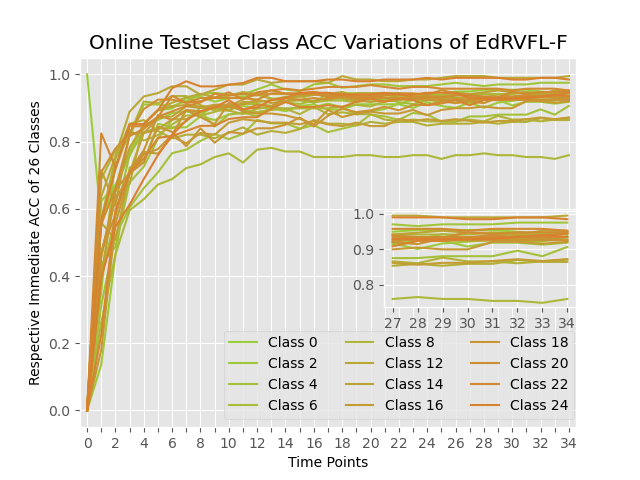}
  \caption{Testset accuracy variation curves of edRVFL-F on 26 classes. Gradient yellow indicates different categories.}
  \label{fig 14}
\end{figure}

\textit{Catastrophic forgetting and distribution drifting.} Referring to Theorem \ref{theorem 1}, the IOL processes remained resistant to these two challenges as they had no relative weight distribution drifting compared to offline experts. The \textit{Letter} classification was a task worth studying. It had 20000 samples from 26 classes. During IOL processes of edRVFL-R/F, the overall accuracy and performance of sub-learners are continuously increasing in Fig. \ref{fig 6}. In Fig. \ref{fig 13} and Fig. \ref{fig 14}, learnable weights are significantly trained in IOL initial phases as the model learns from scratch, resulting in possible local drops of some class accuracy. Minor fluctuations can also occur in IOL processes as the network considers all data, which may sacrifice the accuracy of certain categories to enhance overall accuracy. Generally, the accuracy of each category and the entire dataset shows an upward trend over time, without apparent signs of forgetting or being affected by drift, which means a reasonable trade-off between plasticity and stability. This test also shows -F overperforms -R in IOL processes. 

\section{Conclusion}
Focusing on the dilemmas of online learning such as non-incremental updating, retrospective training, and catastrophic forgetting, this article proposed and analyzed IOL frameworks of Randomized NN on dynamic batch stream in detail, and -F was advocated for the replacement of -R in IOL based on our theoretical and experimental results. Some contributions included: (1) to achieve progressive immediate decision-making on stream, IOL framework that enhanced incremental learning was proposed to facilitate continuous evolvement of deep Randomized NN. The IOL framework alleviated the existing \textit{challenges} and \textit{restrictions} in online learning. 
(2)Within the IOL framework, algorithms of edRVFL-R/-F were derived with variable learning rates and recursive weight update policy. -F style further propelled the IOL process.
(3) we conducted theoretical analysis and derived cumulative regret bounds for IOL processes of edRVFL-R/F on batch streams by a novel methodology, which demonstrated the advantages of online learning acceleration and lower regrets of employing -F during IOL. Several corollaries were presented for further studies on the regret process.
(4) extensive experiments on regression, classification, and huge dataset tasks were conducted for the efficacy of proposed IOL frameworks of edRVFL-R/F and the advantages of using -F. In the future, we would like to explore online learning improvements by using different regularization, and methodologies of lifelong class-IL tasks.





\bibliographystyle{ieeetr}

\bibliography{Incremental_Online_Learning_of_Randomized_Neural_Network_with_Forward_Regularization}



\clearpage
\section*{Supplemental Materials}
\subsection{Theorem 1}
\begin{mytheorem}\textbf{\textit{1 }}
By using -R/-F within IOL frameworks of edRVFL, for $0 \leqslant t \leqslant T$, trainable weights in edRVFL-R/F can hold approximate optimal solutions as the ones of an up-to-date synchronous offline expert, which revisits and globally learns all previously observed data at each time point. This indicates no relative weight distribution drifting between synchronous offline experts and the IOL processes for edRVFL-R/F.
\end{mytheorem}
\noindent\textbf{\textit{Proof. }} 
We first study the learnable weight distribution changes and stability of previous knowledge memory during IOL. Directly analyze the clustered sub-learners inside edRVFL based on Lemma \ref{lemma 2.3again}. For IOL process of edRVFL-R,
\begin{equation} \label{11}
\begin{split}
  \beta _{l,t + 1}^r &= argmi{n_\beta }{\text{ }}{\Delta _{U_{l,t}^r}}(\beta ,\beta _{l,t}^r) + {L_{l,t}}(\beta ) \\ 
   &= argmi{n_\beta }{\text{ }}U_{l,t}^r(\beta ) - U_{l,t}^r(\beta _{l,t}^r)\\
   &- (\beta  - \beta _{l,t}^r)\nabla U_{l,t}^r(\beta _{l,t}^r) + {L_{l,t}}(\beta ). 
\end{split}
\end{equation}
Note $\nabla U_{l,t}^r(\beta _{l,t}^r) = 0$ and $U_{l,t + 1}^r = U_{l,t}^r + {L_{l,t}}$ considering of the latest optimization solution, rewrite (\ref{11}) as:
\begin{equation} \label{12}
\begin{split}
\beta _{l,t + 1}^r &= argmi{n_\beta }{\text{ }}U_{l,t + 1}^r(\beta ) - U_{l,t}^r(\beta _{l,t}^r) \\ 
&= argmi{n_\beta }{\text{ }}{\Delta _{U_{l,0}^r}}(\beta ,\beta _{l,0}^r) + {L_{l,1..t}}(\beta ) - const. \\ 
\end{split}
\end{equation}
This result shows that, at each time point, effect of incrementally updating edRVFL-R within IOL framework on batch stream is equal to employing an offline expert to be trained on global data observed so far. This means the solutions to both models (i.e. online learner and offline expert) remain similar as IOL process progresses. 

And for IOL process of edRVFL-F,
\begin{equation} \label{13}
\begin{aligned}
\beta _{l,t + 1}^f&= argmi{n_\beta }{\text{ }}{\Delta _{U_{l,t}^f}}(\beta ,\beta _{l,t}^f) + {L_{l,t}}(\beta ) \\
&\phantom{{}= \hspace{1.52cm}}+ {{\hat L}_{l,t + 1}}(\beta ) - {{\hat L}_{l,t}}(\beta )\\
&= argmi{n_\beta }{\text{ }}U_{l,t}^f(\beta ) - U_{l,t}^f(\beta _{l,t}^f) \\
&\phantom{{}= \hspace{1.52cm}}+ {L_{l,t}}(\beta ) + {{\hat L}_{l,t + 1}}(\beta ) - {{\hat L}_{l,t}}(\beta ) \\ 
&= argmi{n_\beta }{\text{ }}U_{l,t + 1}^f(\beta ) - U_{l,t}^f(\beta _{l,t}^f) \\ 
&= argmi{n_\beta }{\text{ }}{\Delta _{U_{l,0}^f}}(\beta ,\beta _{l,0}^f) + {L_{l,1..t}}(\beta ) \\
&\phantom{{}= \hspace{1.52cm}}+ {{\hat L}_{l,t + 1}}(\beta ) - const.\\ 
\end{aligned}
\end{equation}
where one can config $U_{l,0}^r = U_{l,0}^f$, $\beta _{l,0}^r = \beta _{l,0}^f$ to the same initialization. This result shows that, at each time point, effect of adopting IOL process to update edRVFL-F on batch stream is equal to employing a synchronous offline expert with an extra penalized estimate to be trained on global data observed so far. Estimate is expected to guide learning direction and outperform edRVFL-R in IOL. As before, learners with -F remain similar solutions with offline experts at all times. The IOL frameworks collect foregone knowledge in ${\Delta _{U_{l,t}^{\{ r,f\} }}}$ and remove past rehearsals ostensibly while the experts cannot.

The experts are often well-performed models and periodically retrained on previous global datasets when new batch comes, which effectively suppresses \textit{catastrophic forgetting} and the \textit{distribution drifting} on data and concepts. The working logic of the experts is analogous to Lemma \ref{lemma 2.2}, and experts can accommodate pre-trained deep layer extensions to address challenging tasks. In the Theorem \ref{theorem 2} and Theorem \ref{theorem 3}, we derive the stepwise recursive updates to IOL frameworks of edRVFL-R/F, thereby satisfying the restrictions of \textit{no retrospective retraining} and \textit{low memory usage}. Consequently, the advantage of IOL framework lies in achieving approximate performance as offline experts without the need for \textit{retrospective retraining}. It is feasible to adopt IOL strategy.
\hfill$\blacksquare$
\subsection{Theorem 2}
\begin{mytheorem}\textbf{\textit{2 }}
For the IOL process of edRVFL-R on batch stream, as shown in the Fig. \ref{fig 1}, recursive updates of trainable weights follow $\beta _{l,t + 1}^r = \beta _{l,t}^r - \eta _{l,t}^r(D_{l,t}^T{D_{l,t}}\beta _{l,t}^r - D_{l,t}^T{Y_t})$ with variable learning rate $\eta _{l,t}^r = {({(\eta _{l,0}^r)^{ - 1}} + \sum\limits_{q = 1}^t {D_{l,q}^T{D_{l,q}}} )^{ - 1}}$, where $\eta _{l,0}^r = {({\lambda _l} \cdot I)^{ - 1}}$. This updating policy indicates that incremental updates to weights do not require the involvement of \textit{retrospective retraining} and \textit{high memory usage} of past data.
\end{mytheorem}
\noindent{\textbf{\textit{Proof.}}}
Denote the $t$-th batch loss of $\ell $-th edRVFL layer as ${L_{l,t}}(\beta ) = \frac{1}{2}||{D_{l,t}} \cdot \beta  - {Y_t}||_2^2$, initial Bregman function $U_{l,0}^r(\beta ) = \frac{1}{2}{\beta ^T}{(\eta _{l,0}^r)^{ - 1}}\beta $ where ${(\eta _{l,0}^r)^{ - 1}}$ is a symmetric positive definite matrix. Initial Bregman divergence is defined as ${\Delta _{U_{l,0}^r}}(\beta ,\beta _{l,0}^r) = \frac{1}{2}{(\beta  - \beta _{l,0}^r)^T}{(\eta _{l,0}^r)^{ - 1}}(\beta  - \beta _{l,0}^r)$.
Based on similar logic of Lemma \ref{lemma 2.2}, for $0 \leqslant t \leqslant T$ we have:
\begin{equation} \label{14}
\begin{aligned}
\beta _{l,t + 1}^r &= argmi{n_\beta }{\text{ }}U_{l,t + 1}^r(\beta )\\
U_{l,t + 1}^r(\beta ) &= {\Delta _{U_{l,0}^r}}(\beta ,\beta _{l,0}^r) + {L_{l,1..t}}(\beta ) \\ 
&= \frac{1}{2}{(\beta  - \beta _{l,0}^r)^T}{(\eta _{l,0}^r)^{ - 1}}(\beta  - \beta _{l,0}^r) \\
&+ \sum\limits_{q = 1}^t {\frac{1}{2}||{D_{l,q}} \cdot \beta  - {Y_q}||_2^2}  \\ 
\end{aligned}
\end{equation}

Firstly we introduce two properties of Bregman term here:
\begin{lemma} {\bf Bregman divergence property.} \label{lemma 3.1}
\begin{align*}
&\text{(1): } {\Delta _{{G_1} + \mu {G_2}}}(\tilde \theta ,\theta ) = {\Delta _{{G_1}}}(\tilde \theta ,\theta ) + \mu {\Delta _{{G_2}}}(\tilde \theta ,\theta ), \forall \mu  \geqslant 0\\
&\text{(2): If} \hspace{0.16cm} {G_1}(\theta ) - {G_2}(\theta ) = \omega \theta  + \upsilon, \omega  \in {\mathbb{R}^k}, \upsilon  \in \mathbb{R},\\
&\hspace{0.66cm}\text{then } {\Delta _{{G_1}}}(\tilde \theta ,\theta ) = {\Delta _{{G_2}}}(\tilde \theta ,\theta )
\end{align*}
\end{lemma}

Convert (\ref{14}) into the form described by Lemma \ref{lemma 2.3again}, and derive variable learning rate via Lemma \ref{lemma 3.1}:
\begin{align}
&\phantom{{}= \hspace{0.1cm}}U_{l,t}^r(\beta ) + \frac{1}{2}||{D_{l,t}} \cdot \beta  - {Y_t}||_2^2\nonumber\\
&= U_{l,0}^r(\beta ) - U_{l,0}^r(\beta _{l,0}^r) \nonumber\\
&- {(\beta  - \beta _{l,0}^r)^T}\nabla U_{l,0}^r(\beta _{l,0}^r) + \sum\limits_{q = 1}^t {\frac{1}{2}||{D_{l,q}} \cdot \beta  - {Y_q}||_2^2}   \nonumber\\
&\Rightarrow U_{l,t}^r(\beta ) - U_{l,0}^r(\beta ) \nonumber\\
&= \sum\limits_{q = 1}^{t - 1} {\frac{1}{2}||{D_{l,q}} \cdot \beta  - {Y_q}||_2^2}  \nonumber\\
&- \frac{1}{2}{(\beta _{l,0}^r)^T}{(\eta _{l,0}^r)^{ - 1}}\beta _{l,0}^r - {(\beta  - \beta _{l,0}^r)^T}{(\eta _{l,0}^r)^{ - 1}}\beta _{l,0}^r  \nonumber
\end{align}
\begin{align}
&\Rightarrow U_{l,t}^r(\beta ) - U_{l,0}^r(\beta )\nonumber\\
&= \frac{1}{2}{\beta ^T}(\sum\limits_{q = 1}^{t - 1} {D_{l,q}^T{D_{l,q}})\beta }  - \sum\limits_{q = 1}^{t - 1} {Y_q^T{D_{l,q}}\beta }\nonumber\\
&+ \frac{1}{2}\sum\limits_{q = 1}^{t - 1} {Y_q^T{Y_q}}  + \frac{1}{2}{(\beta _{l,0}^r)^T}{(\eta _{l,0}^r)^{ - 1}}\beta _{l,0}^r \!-\! {\beta ^T}{(\eta _{l,0}^r)^{ - 1}}\beta _{l,0}^r\nonumber  \\
&U_{l,t}^r(\beta ) - (U_{l,0}^r(\beta ) + \frac{1}{2}{\beta ^T}(\sum\limits_{q = 1}^{t - 1} {D_{l,q}^T{D_{l,q}})\beta } ) = \omega \beta  + \upsilon  \label{15}
\end{align}

Based on Lemma \ref{lemma 3.1} and let $V_{l,t - 1}^r = \frac{1}{2}{\beta ^T}(\sum\limits_{q = 1}^{t - 1} {D_{l,q}^T{D_{l,q}})\beta } $, relation can be obtained from (\ref{15}):
\begin{equation}
\begin{aligned} \label{15.1}
{\Delta _{U_{l,t}^r}}(\beta ,\tilde \beta ) &= {\Delta _{U_{l,0}^r + V_{l,t - 1}^r}}(\beta ,\tilde \beta ) \\
&= {\Delta _{U_{l,0}^r}}(\beta ,\tilde \beta ) + {\Delta _{V_{l,t - 1}^r}}(\beta ,\tilde \beta )
\end{aligned}
\end{equation}

To obtain the analogical form of (\ref{8again}), we can set $\tilde \beta  = \beta _{l,t}^r$ to (\ref{15.1}) and have:\\
${\Delta _{U_{l,t}^r}}(\beta ,\beta _{l,t}^r) \!=\! \frac{1}{2}{(\beta  - \beta _{l,t}^r)^T}[{(\eta _{l,0}^r)^{ - 1}} + \sum\limits_{q = 1}^{t - 1} {D_{l,q}^T{D_{l,q}}]} (\beta  - \beta _{l,t}^r)$

As the IOL process collects all past knowledge into Bregman divergence, the main change of optimization equation of IOL process at different times occurs in the divergence terms. The optimization solution to that IOL process will change with the varying $(\eta _{l,0}^r)$. So that the variable learning rate can be $\eta _{l,t - 1}^r = {({(\eta _{l,0}^r)^{ - 1}} + \sum\limits_{q = 1}^{t - 1} {D_{l,q}^T{D_{l,q}}} )^{ - 1}}$, namely $\eta _{l,t}^r = {({(\eta _{l,0}^r)^{ - 1}} + \sum\limits_{q = 1}^t {D_{l,q}^T{D_{l,q}}} )^{ - 1}}$ and specifically $\eta _{l,0}^r = {({\lambda _l} \cdot I)^{ - 1}}$ when $t = 0$. The recursive updates to the learning rate can be solved by Sherman's equation. 

The incremental IOL process of edRVFL-R can be represented as:
\begin{equation} \label{16}
\begin{split}
\beta _{l,t + 1}^r &= argmi{n_\beta }{\text{ }}{\Delta _{U_{l,t}^r}}(\beta ,\beta _{l,t}^r) + {L_{l,t}}(\beta ) \\
&= argmi{n_\beta }{\text{ }}\frac{1}{2}{(\beta  - \beta _{l,t}^r)^T}{(\eta _{l,t - 1}^r)^{ - 1}}(\beta  - \beta _{l,t}^r) \\
&\phantom{{}= \hspace{1.52cm}}+ \frac{1}{2}||{D_{l,t}} \cdot \beta  - {Y_t}||_2^2
\end{split}
\end{equation}

Finally, the relations of stepwise updating between $\beta _{l,t}^r$ and $\beta _{l,t + 1}^r$ can be derived. (\ref{16}) can be transformed to the following constrained optimization problem:
\begin{align}\label{16.0}
&\mathop {min}\limits_{\beta  \in \beta _{l,t + 1}^r} \;\frac{1}{2}{(\beta  - \beta _{l,t}^r)^T}{(\eta _{l,t - 1}^r)^{ - 1}}(\beta  - \beta _{l,t}^r) + \frac{1}{2}||\xi ||_2^2  \\
&\phantom{{}= \hspace{0.32cm}} s.t.\;{D_{l,t}} \cdot \beta  - {Y_t} = \xi ,\forall t \nonumber
\end{align}
where the $\xi $ denotes the gap between ground truth and prediction. The Lagrangian function of problem (\ref{16.0}) is:
\begin{align}\label{16.1}
\mathcal{L}(\beta ,\xi ,\mu ) &= \frac{1}{2}{(\beta  - \beta _{l,t}^r)^T}{(\eta _{l,t - 1}^r)^{ - 1}}(\beta  - \beta _{l,t}^r) \\
&+ \frac{1}{2}||\xi ||_2^2 - {\mu ^T}({D_{l,t}} \cdot \beta  - {Y_t} - \xi ) \nonumber
\end{align}
where $\mu $ denotes Lagrangian multiplier. 

The Lagrangian function (\ref{16.1}) can be tackled through Karush-Kuhn-Tucker (KKT) conditions, which can be built into the following formulation:
\begin{align}\label{16.2}
&\frac{{\partial \mathcal{L}(\beta ,\xi ,\mu )}}{{\partial \beta }} = 0 \Rightarrow {(\eta _{l,t - 1}^r)^{ - 1}}(\beta  - \beta _{l,t}^r) - D_{l,t}^T \cdot \mu  = 0 \nonumber \\
&\frac{{\partial \mathcal{L}(\beta ,\xi ,\mu )}}{{\partial \xi }} = 0 \Rightarrow \xi  + \mu  = 0\\
&\frac{{\partial \mathcal{L}(\beta ,\xi ,\mu )}}{{\partial \mu }} = 0 \Rightarrow {D_{l,t}} \cdot \beta  - {Y_t} - \xi  = 0 \nonumber 
\end{align}

Based on (\ref{16.2}), we can obtain the recursive updating policy between $\beta _{l,t}^r$ and $\beta _{l,t+1}^r$ as follows:
\begin{align}
{(\eta _{l,t - 1}^r)^{ - 1}}(\beta _{l,t + 1}^r &- \beta _{l,t}^r) + D_{l,t}^T({D_{l,t}} \cdot \beta _{l,t + 1}^r - {Y_t}) = 0 \nonumber\\ 
{(\eta _{l,t}^r)^{ - 1}}\beta _{l,t + 1}^r &= ({(\eta _{l,t}^r)^{ - 1}} - D_{l,t}^T{D_{l,t}})\beta _{l,t}^r + D_{l,t}^T{Y_t} \label{17}\\ 
\beta _{l,t + 1}^r &= \beta _{l,t}^r - \eta _{l,t}^r(D_{l,t}^T{D_{l,t}}\beta _{l,t}^r - D_{l,t}^T{Y_t})  \nonumber
\end{align}
Using this methodology, we can obtain the variable learning rate and recursive weight updating policy. The (\ref{17}) endows the edRVFL-R with the ability to get rid of \textit{retrospective retraining}, because the values are only relevant to the chunk at the current time. Specific algorithmic flows of IOL for edRVFL-R could refer to \textbf{Algorithm 1}.
\hfill$\blacksquare$
\subsection{Theorem 3}
\begin{mytheorem}\textbf{\textit{3 }}
For the IOL process of edRVFL-F on batch stream, as shown in the Fig. \ref{fig 1}, stepwise updates of learnable weights follow $\beta _{l,t + 1}^f = \beta _{l,t}^f - \eta _{l,t + 1}^f(D_{l,t + 1}^T{D_{l,t + 1}}\beta _{l,t}^f - D_{l,t}^T{Y_t}) + \eta _{l,t + 1}^f(D_{l,t + 1}^T{D_{l,t + 1}} - D_{l,t}^T{D_{l,t}})\beta _{l,0}^f$ with variable learning rate $\eta _{l,t + 1}^f = {({(\eta _{l,0}^f)^{ - 1}} + \sum\limits_{q = 1}^{t + 1} {D_{l,q}^T{D_{l,q}}} )^{ - 1}}$, where $\eta _{l,0}^f =$ $ {({\lambda _l} \cdot I)^{ - 1}}$. Specially, $\beta _{l,t + 1}^f = \beta _{l,t}^f - \eta _{l,t + 1}^f(D_{l,t + 1}^T{D_{l,t + 1}}\beta _{l,t}^f - D_{l,t}^T{Y_t})$ given $\beta _{l,0}^f = 0$. This recursive policy indicates that incremental weight updates do not require the involvement of \textit{retrospective retraining} and \textit{high memory usage} of past data at $t$ time. The optimization processes of IOL with -R/-F can be demonstrated in Fig. \ref{fig 2}.
\end{mytheorem}
\noindent{\textbf{\textit{Proof.}}}
For the forward learning, define the extra predictive loss term ${\hat L_{l,t + 1}}(\beta ) = \frac{1}{2}||{D_{l,t + 1}}(\beta  - \beta _{l,0}^f)||_2^2$ (here the baseline $\beta _{l,0}^f$ can be substituted of 0 or other values containing prior knowledge for different performance), initial Bregman function $U_{l,0}^f(\beta ) = \frac{1}{2}{\beta ^T}{(\eta _{l,0}^f)^{ - 1}}\beta $, where ${(\eta _{l,0}^f)^{ - 1}} = {\lambda _l} \cdot I$ is a symmetric positive definite matrix, and initial Bregman divergence term can be written as ${\Delta _{U_{l,0}^f}}(\beta ,\beta _{l,0}^f) = \frac{1}{2}{(\beta  - \beta _{l,0}^f)^T}{(\eta _{l,0}^f)^{ - 1}}(\beta  - \beta _{l,0}^f)$.
Based on similar logic of Lemma \ref{lemma 2.2}, for $0 \leqslant t \leqslant T$, we have:
\begin{align}
\beta _{l,t + 1}^f &= argmi{n_\beta }{\text{ }}U_{l,t + 1}^f(\beta )\nonumber\\
U_{l,t + 1}^f(\beta ) &= {\Delta _{U_{l,0}^f}}(\beta ,\beta _{l,0}^f) + {L_{l,1..t}}(\beta ) + {\hat L_{l,t + 1}}(\beta ) \nonumber\\
&= \frac{1}{2}{(\beta  - \beta _{l,0}^f)^T}{(\eta _{l,0}^f)^{ - 1}}(\beta  - \beta _{l,0}^f) \label{18}\\
&+ \sum\limits_{q = 1}^t {\frac{1}{2}||{D_{l,q}} \cdot \beta  - {Y_q}||_2^2}  + \frac{1}{2}||{D_{l,t + 1}}(\beta  - \beta _{l,0}^f)||_2^2\nonumber
\end{align}

Firstly convert (\ref{18}) into the form described by Lemma \ref{lemma 2.3again} and compute the variable learning rate using Lemma \ref{lemma 3.1}:
\begin{align}
&\phantom{{}= \hspace{0.1cm}}U_{l,t}^f(\beta ) + \frac{1}{2}||{D_{l,t}} \cdot \beta  - {Y_t}||_2^2  - \frac{1}{2}||{D_{l,t}}(\beta  - \beta _{l,0}^f)||_2^2 \nonumber\\
&= U_{l,0}^f(\beta )  + \sum\limits_{q = 1}^t {\frac{1}{2}||{D_{l,q}} \cdot \beta  - {Y_q}||_2^2} \nonumber\\
&  - {(\beta  - \beta _{l,0}^f)^T}\nabla U_{l,0}^f(\beta _{l,0}^f) - U_{l,0}^f(\beta _{l,0}^f)\hfill \nonumber\\
& \Rightarrow U_{l,t}^f(\beta ) - U_{l,0}^f(\beta ) \nonumber\\
&= \sum\limits_{q = 1}^{t - 1} {\frac{1}{2}||{D_{l,q}} \cdot \beta  - {Y_q}||_2^2}  - {\beta ^T}{(\eta _{l,0}^f)^{ - 1}}\beta _{l,0}^f \hfill \nonumber\\
&+ \frac{1}{2}||{D_{l,t}}(\beta  - \beta _{l,0}^f)||_2^2 + \frac{1}{2}{(\beta _{l,0}^f)^T}{(\eta _{l,0}^f)^{ - 1}}\beta _{l,0}^f \nonumber\\
&\Rightarrow U_{l,t}^f(\beta ) - U_{l,0}^f(\beta )\nonumber\\ 
&= \frac{1}{2}{\beta ^T}(\sum\limits_{q = 1}^{t - 1} {D_{l,q}^T{D_{l,q}})\beta }  - \sum\limits_{q = 1}^{t - 1} {Y_q^T{D_{l,q}}\beta }  \nonumber\\
&+ \frac{1}{2}\sum\limits_{q = 1}^{t - 1} {Y_q^T{Y_q}}  + \frac{1}{2}{(\beta _{l,0}^f)^T}{(\eta _{l,0}^f)^{ - 1}}\beta _{l,0}^f \nonumber\\
&+ \frac{1}{2}||{D_{l,t}}(\beta  - \beta _{l,0}^f)||_2^2 - {\beta ^T}{(\eta _{l,0}^f)^{ - 1}}\beta _{l,0}^f \hfill \nonumber\\
&U_{l,t}^f(\beta ) - (U_{l,0}^f(\beta ) + \frac{1}{2}{\beta ^T}(\sum\limits_{q = 1}^t {D_{l,q}^T{D_{l,q}})\beta } ) = \omega \beta  + \upsilon   \label{19}
\end{align}

Based on Lemma \ref{lemma 3.1} and Lemma \ref{lemma 2.3again}, set $V_{l,t }^f = \frac{1}{2}{\beta ^T}(\sum\limits_{q = 1}^{t } {D_{l,q}^T{D_{l,q}})\beta } $, and relation can be obtained from (\ref{19}):

\begin{equation}
\begin{aligned} \label{19.5}
{\Delta _{U_{l,t}^f}}(\beta ,\tilde \beta ) &= {\Delta _{U_{l,0}^f + V_{l,t }^f}}(\beta ,\tilde \beta ) \\
&= {\Delta _{U_{l,0}^f}}(\beta ,\tilde \beta ) + {\Delta _{V_{l,t }^f}}(\beta ,\tilde \beta )
\end{aligned}
\end{equation}

The ${\Delta _{U_{l,t}^f}}(\beta ,\beta _{l,t}^f)$ can be obtained from (\ref{19})-(\ref{19.5}):\\
\begin{equation*}
\begin{aligned}
{\Delta _{U_{l,t}^f}}(\beta ,\beta _{l,t}^f) &= \frac{1}{2}{(\beta  - \beta _{l,t}^f)^T}[{(\eta _{l,0}^f)^{ - 1}} + \sum\limits_{q = 1}^t {D_{l,q}^T{D_{l,q}}]} (\beta  - \beta _{l,t}^f) \\ 
   &= \frac{1}{2}{(\beta  - \beta _{l,t}^f)^T}{(\eta _{l,t}^f)^{ - 1}}(\beta  - \beta _{l,t}^f) \\ 
\end{aligned}
\end{equation*}

Same as Theorem \ref{theorem 2}, the solution to the equation that needs stepwise optimization during IOL process will change with the varying $\eta _{l,t}^f$. It is easy to obtain the variable learning rate $\eta _{l,t}^f = {({(\eta _{l,0}^f)^{ - 1}} + \sum\limits_{q = 1}^t {D_{l,q}^T{D_{l,q}}} )^{ - 1}}$. Thus, the batch IOL process of edRVFL-F can be rewritten as follows. Sherman’s equation can solve the recursive updates to the learning rate. One can compare it with (\ref{16}) to see the advancement in learning rate.
\begin{equation} \label{20}
\begin{aligned}
\beta _{l,t + 1}^f &= argmi{n_\beta }{\text{ }}{\Delta _{U_{l,t}^f}}(\beta ,\beta _{l,t}^f) + {L_{l,t}}(\beta ) \\
&\phantom{{}= \hspace{15.2mm}}+ {{\hat L}_{l,t + 1}}(\beta ) - {{\hat L}_{l,t}}(\beta ) \\ 
&= argmi{n_\beta }{\text{ }}\frac{1}{2}{(\beta  - \beta _{l,t}^f)^T}{(\eta _{l,t}^f)^{ - 1}}(\beta  - \beta _{l,t}^f) \\
&\phantom{{}= \hspace{15.2mm}}+ \frac{1}{2}||{D_{l,t}} \cdot \beta  - {Y_t}||_2^2 \\
&\phantom{{}= \hspace{15.2mm}}+ \frac{1}{2}||{D_{l,t + 1}}(\beta  - \beta _{l,0}^f)||_2^2 \\
&\phantom{{}= \hspace{15.2mm}}- \frac{1}{2}||{D_{l,t}}(\beta  - \beta _{l,0}^f)||_2^2 \\ 
\end{aligned}
\end{equation}

(\ref{20}) can be transformed to the following constrained optimization problem:
\begin{align}\label{20.0}
&\mathop {min}\limits_{\beta  \in \beta _{l,t + 1}^f} \frac{1}{2}{(\beta  - \beta _{l,t}^f)^T}{(\eta _{l,t}^f)^{ - 1}}(\beta  - \beta _{l,t}^f) \nonumber\\
&\phantom{{}= \hspace{0.42cm}}+ \frac{1}{2}||{\xi _1}||_2^2 + \frac{1}{2}||{\xi _2}||_2^2 - \frac{1}{2}||{\xi _3}||_2^2 \\
&\phantom{{}= \hspace{0.32cm}} s.t.\;{D_{l,t}} \cdot \beta  - {Y_t} = {\xi _1},\forall t \nonumber\\
&\phantom{{}= \hspace{0.92cm}}{D_{l,t + 1}}(\beta  - \beta _{l,0}^f) = {\xi _2},\forall t\nonumber\\
&\phantom{{}= \hspace{0.92cm}}{D_{l,t}}(\beta  - \beta _{l,0}^f) = {\xi _3},\forall t\nonumber
\end{align}
where the ${\xi _{\{ 1,2,3\} }}$ denotes the gap between ground truth and prediction. The Lagrangian function of problem (\ref{20.0}) is:
\begin{align}\label{20.1}
\mathcal{L}(\beta ,{\xi _{\{ 1,2,3\} }},{\mu _{\{ 1,2,3\} }}) &= \frac{1}{2}{(\beta  - \beta _{l,t}^f)^T}{(\eta _{l,t}^f)^{ - 1}}(\beta  - \beta _{l,t}^f)\nonumber\\
&+ \frac{1}{2}||{\xi _1}||_2^2 + \frac{1}{2}||{\xi _2}||_2^2 - \frac{1}{2}||{\xi _3}||_2^2\nonumber\\
&+ \mu _1^T({D_{l,t}} \cdot \beta  - {Y_t} - {\xi _1})\\
&+ \mu _2^T({D_{l,t + 1}}(\beta  - \beta _{l,0}^f) - {\xi _2})\nonumber\\
&+ \mu _3^T({D_{l,t}}(\beta  - \beta _{l,0}^f) - {\xi _3})\nonumber
\end{align}
where ${\mu _{\{ 1,2,3\} }}$ denotes Lagrangian multiplier. 

The Lagrangian function (\ref{20.1}) can be tackled through Karush-Kuhn-Tucker (KKT) conditions, which can be built into the following formulation:
\begin{align}\label{20.2}
&\frac{{\partial \mathcal{L}(\beta ,\xi ,\mu )}}{{\partial \beta }} = 0 \Rightarrow \nonumber\\
&{(\eta _{l,t}^f)^{ - 1}}(\beta  - \beta _{l,t}^f) + D_{l,t}^T{\mu _1} + D_{l,t + 1}^T{\mu _2} + D_{l,t}^T{\mu _3} = 0\nonumber\\
&\frac{{\partial \mathcal{L}(\beta ,\xi ,\mu )}}{{\partial {\xi _1}}} = 0 \Rightarrow {\xi _1} = {\mu _1}\nonumber\\
&\frac{{\partial \mathcal{L}(\beta ,\xi ,\mu )}}{{\partial {\xi _2}}} = 0 \Rightarrow {\xi _2} = {\mu _2}\nonumber\\
&\frac{{\partial \mathcal{L}(\beta ,\xi ,\mu )}}{{\partial {\xi _3}}} = 0 \Rightarrow  - {\xi _3} = {\mu _3}\\
&\frac{{\partial \mathcal{L}(\beta ,\xi ,\mu )}}{{\partial {\mu _1}}} = 0 \Rightarrow {D_{l,t}} \cdot \beta  - {Y_t} = {\xi _1}\nonumber\\
&\frac{{\partial \mathcal{L}(\beta ,\xi ,\mu )}}{{\partial {\mu _2}}} = 0 \Rightarrow {D_{l,t + 1}}(\beta  - \beta _{l,0}^f) = {\xi _2}\nonumber\\
&\frac{{\partial \mathcal{L}(\beta ,\xi ,\mu )}}{{\partial {\mu _3}}} = 0 \Rightarrow {D_{l,t}}(\beta  - \beta _{l,0}^f) = {\xi _3}\nonumber
\end{align}

Based on (\ref{20.2}), we can obtain the recursive updating policy between $\beta _{l,t}^f$ and $\beta _{l,t+1}^f$ as follows:
\begin{align}
&{(\eta _{l,t}^f)^{ - 1}}(\beta _{l,t + 1}^f - \beta _{l,t}^f) + D_{l,t}^T({D_{l,t}} \cdot \beta _{l,t + 1}^f - {Y_t}) \nonumber\\
&+ D_{l,t + 1}^T{D_{l,t + 1}}(\beta _{l,t + 1}^f - \beta _{l,0}^f) - D_{l,t}^T{D_{l,t}}(\beta _{l,t + 1}^f - \beta _{l,0}^f) = 0 \nonumber\\ 
& \Rightarrow {(\eta _{l,t + 1}^f)^{ - 1}}\beta _{l,t + 1}^f \nonumber\\
&= ({(\eta _{l,t + 1}^f)^{ - 1}} - D_{l,t + 1}^T{D_{l,t + 1}})\beta _{l,t}^f \label{21}\\
&+ D_{l,t}^T{Y_t} + (D_{l,t + 1}^T{D_{l,t + 1}} - D_{l,t}^T{D_{l,t}})\beta _{l,0}^f \nonumber\\ 
&\Rightarrow \beta _{l,t + 1}^f \nonumber\\
&= \beta _{l,t}^f - \eta _{l,t + 1}^f(D_{l,t + 1}^T{D_{l,t + 1}}\beta _{l,t}^f - D_{l,t}^T{Y_t}) \nonumber\\
&+ \eta _{l,t + 1}^f(D_{l,t + 1}^T{D_{l,t + 1}} - D_{l,t}^T{D_{l,t}})\beta _{l,0}^f \nonumber 
\end{align}

Without loss of generality, given $\beta _{l,0}^f = 0$ to allow the learner to start from no prior knowledge, we have:
\begin{equation}\label{21.1}
\beta _{l,t + 1}^f = \beta _{l,t}^f - \eta _{l,t + 1}^f(D_{l,t + 1}^T{D_{l,t + 1}}\beta _{l,t}^f - D_{l,t}^T{Y_t})
\end{equation}

Using this methodology, we can obtain the variable learning rate and weight updating policy. The (\ref{21.1}) endows the edRVFL-R with the ability to get rid of \textit{retrospective retraining}, because the values are only relevant to the chunks of current and next data batches. The specific algorithmic flows of IOL for edRVFL-F could refer to \textbf{Algorithm 2}.
\hfill$\blacksquare$
\subsection{Theorem 4}
\begin{mytheorem}\textbf{\textit{4 }}
For the IOL process of the $l$-th online learner inside edRVFL-R working on batch stream, during $0 \leqslant t \leqslant T$, the upper bound of online-to-offline cumulative regrets between this learner and an offline expert can be written as:
\begin{equation*} 
\begin{split}
\sum\limits_{t = 1}^T {{L_{l,t}}(\beta _{l,t}^r)}  - \mathop {min}\limits_{\beta _{l, \cdot }^r} ({\Delta _{U_{l,0}^r}}(\beta _{l, \cdot }^r,\beta _{l,0}^r) + {L_{l,1..T}}(\beta _{l, \cdot }^r)) \\
\leqslant 2Y_m^2b(N + k)In(1 + \frac{{TD_m^2b}}{{{\lambda _l}s}})
\end{split}
\end{equation*}
where $\beta _{l, \cdot }^r$ is one solution of the offline expert, $s$ is the amplification factor of origin ${\lambda _l}$, $In$ denotes natural logarithm, ${D_m} = \mathop {\max }\limits_{1 \leqslant t \leqslant T} \{ ||{D_{l,t}}|{|_\infty }\} $, and ${Y_m} = \mathop {\max }\limits_{1 \leqslant t \leqslant T} \{ ||{Y_t}|{|_\infty },||{D_{l,t}}\beta _{l,t}^r|{|_\infty }\} $.
\end{mytheorem}
\noindent{\textbf{\textit{Proof.}}}
Based on Lemma \ref{lemma 3.2} and (\ref{15})-(\ref{16}) results, the right part of (\ref{22}) can be rewritten as following:
\begin{equation} \label{23}
\begin{split}
&\sum\limits_{t = 1}^T {{\Delta _{U_{l,t + 1}^r}}(\beta _{l,t}^r,\beta _{l,t + 1}^r)}  - {\Delta _{U_{l,T + 1}^r}}(\beta _{l, \cdot }^r,\beta _{l,T + 1}^r) \hfill \\
=& \sum\limits_{t = 1}^T {\frac{1}{2}{{(\beta _{l,t}^r - \beta _{l,t + 1}^r)}^T}{{(\eta _{l,t}^r)}^{ - 1}}(\beta _{l,t}^r - \beta _{l,t + 1}^r)}  \\
-& {\Delta _{U_{l,T + 1}^r}}(\beta _{l, \cdot }^r,\beta _{l,T + 1}^r) \hfill \\ 
\end{split}
\end{equation}

It’s interesting to study the difference of cumulative regrets between online learners and the optimal offline expert, as it gives the upper bound of performance gap under adversarial setting. The relative cumulative regret gap between online learners and other experts which perform more poorly will be below this threshold. Assume there always a solution set to the offline expert, based on Lemma \ref{lemma 2.1} and Lemma \ref{lemma 2.3again} and Theorem \ref{theorem 1}, let $\beta _{l, \cdot }^r = \beta _{l, * }^r$ (i.e. the optimal solution), the updated $\beta _{l,t + 1}^r$ in IOL process will converge to the optimal solution of offline expert in the end, namely $\beta _{l, * }^r = \beta _{l,T + 1}^r$. Use (\ref{17}) twice and $\eta _{l,t}^r$ is positive definite so that (\ref{23}) can be restricted by:
\begin{equation} \label{24}
\begin{split}
&\leqslant \sum\limits_{t = 1}^T {\frac{1}{2}{{(\beta _{l,t}^r - \beta _{l,t + 1}^r)}^T}{{(\eta _{l,t}^r)}^{ - 1}}(\beta _{l,t}^r - \beta _{l,t + 1}^r)}  \hfill \\
&= \sum\limits_{t = 1}^T {\frac{1}{2}} {({D_{l,t}}\beta _{l,t}^r - {Y_t})^T}{D_{l,t}}\eta _{l,t}^rD_{l,t}^T({D_{l,t}}\beta _{l,t}^r - {Y_t}) \hfill \\
&= \sum\limits_{t = 1}^T {\frac{1}{2}||{{({D_{l,t}}\beta _{l,t}^r - {Y_t})}^T}{D_{l,t}}||_2^2}  \\
&\times{\frac{{{{({D_{l,t}}\beta _{l,t}^r - {Y_t})}^T}{D_{l,t}}}}{{||{{({D_{l,t}}\beta _{l,t}^r - {Y_t})}^T}{D_{l,t}}|{|_2}}}\eta _{l,t}^r\frac{{D_{l,t}^T({D_{l,t}}\beta _{l,t}^r - {Y_t})}}{{||D_{l,t}^T({D_{l,t}}\beta _{l,t}^r - {Y_t})|{|_2}}}}  \hfill \\ 
\end{split}
\end{equation}

According to the Matrix Spectral Theorem, it can be proved that ${e_{max}}(A) = \sup ({p^T}Ap|{\kern 1pt} \,||p|{|_2} = 1)$ is a convex function of $A$, where ${e_{max}}$ serves as the maximum eigenvalue of real-valued symmetric matrix $A$. (\ref{24}) can be simplified to:
\begin{equation} \label{25}
\leqslant \frac{1}{2}||{({D_{l,t}}\beta _{l,t}^r - {Y_t})^T}{D_{l,t}}||_2^2\sum\limits_{t = 1}^T {{e_{max}}(\eta _{l,t}^r)}. 
\end{equation}

To prevent the model from potential attacks, in the adversarial setting, we assume ${D_{l,t}}\beta _{l,t}^r$ lies in vector $[ - {Y_m},{Y_m}] \cdot {{\textbf{1}}}$.
The first term of (\ref{25}) can be bounded to:
\vspace{-2mm}
\begin{align}
&\phantom{{}= \hspace{1mm}}||{({D_{l,t}}\beta _{l,t}^r - {Y_t})^T}{D_{l,t}}||_2^2 \hfill \nonumber\\
&= \sum\limits_{i,j} {[(({D_{l,t}}\beta _{l,t}^r - {Y_t}){{({D_{l,t}}\beta _{l,t}^r - {Y_t})}^T}) \odot } ({D_{l,t}}D_{l,t}^T){]_{i,j}} \hfill \nonumber\\
&\leqslant 4Y_m^2{b^2}D_m^2(N + k),  \label{26}
\end{align}
 ${D_m} \!=\! \mathop {\max }\limits_{1 \leqslant t \leqslant T} \{ ||{D_{l,t}}|{|_\infty }\} $, ${Y_m} \!=\!
 \mathop {\max }\limits_{1 \leqslant t \leqslant T} \{ ||{Y_t}|{|_\infty },||{D_{l,t}}\beta _{l,t}^r|{|_\infty }\} $ here, and $(N + k)$ is dimensions of $\eta _{l,t}^r$, and $ \odot $ is Hadamard product operation.
The second term of (\ref{25}) can be bounded to:
\begin{equation} \label{27}
\begin{split}
\sum\limits_{t = 1}^T {{e_{max}}(\eta _{l,t}^r)}  = \sum\limits_{t = 1}^T {\frac{1}{{{e_{min}}({{(\eta _{l,t}^r)}^{ - 1}})}}} 
\end{split}
\end{equation}
while $\sup ({e_{min}}({(\eta _{l,t}^r)^{ - 1}} - {(\eta _{l,0}^r)^{ - 1}})) \leqslant D_m^2bt$, because $\sum\limits_i {{e_i}({{(\eta _{l,t}^r)}^{ - 1}} - {{(\eta _{l,0}^r)}^{ - 1}})}  = trace({(\eta _{l,t}^r)^{ - 1}} - {(\eta _{l,0}^r)^{ - 1}})$.
Consider initial conditions in Theorem \ref{theorem 2}, (\ref{27}) can be computed via integration of series and bounded to:
\begin{equation} \label{28}
\begin{split}
\sum\limits_{t = 0}^T {\inf \frac{1}{{{e_{min}}({{(\eta _{l,t}^r)}^{ - 1}})}}}  &= \int_0^T {\frac{1}{{{\lambda _l} + D_m^2bt}}dt} \\
&= {\frac{1}{{D_m^2b}}In(1 + \frac{{TD_m^2b}}{{{\lambda _l}}})} 
\end{split}
\end{equation}

Summarize (\ref{25})-(\ref{28}), (\ref{24}) can be bounded to:
\begin{equation} \label{29}
\begin{split}
\sum\limits_{t = 1}^T {\frac{1}{2}{{(\beta _{l,t}^r - \beta _{l,t + 1}^r)}^T}{{(\eta _{l,t}^r)}^{ - 1}}(\beta _{l,t}^r - \beta _{l,t + 1}^r)}  \\
\leqslant 2Y_m^2b(N + k)In(1 + \frac{{TD_m^2b}}{{{\lambda _l}}})
\end{split}
\end{equation}

Actually, ${\lambda _l}$ should be fine-tuned to fit the model learning when batched volume changes. For consistency, here ${\lambda _l}$ remains as before and a more general theorem is given by:
\begin{equation} \label{30}
\begin{split}
\sum\limits_{t = 1}^T {\frac{1}{2}{{(\beta _{l,t}^r - \beta _{l,t + 1}^r)}^T}{{(\eta _{l,t}^r)}^{ - 1}}(\beta _{l,t}^r - \beta _{l,t + 1}^r)}  \\
\leqslant 2Y_m^2b(N + k)In(1 + \frac{{TD_m^2b}}{{{\lambda _l}s}})
\end{split}
\end{equation}
where $s$ is the amplification factor of ${\lambda _l}$.

Look back to the (\ref{23}), we obtain the cumulative regret bound for IOL with -R by the following equation:
\begin{equation*} 
\begin{split}
\sum\limits_{t = 1}^T {{L_{l,t}}(\beta _{l,t}^r)}  - \mathop {min}\limits_{\beta _{l, \cdot }^r} ({\Delta _{U_{l,0}^r}}(\beta _{l, \cdot }^r,\beta _{l,0}^r) + {L_{l,1..T}}(\beta _{l, \cdot }^r)) \\
\leqslant 2Y_m^2b(N + k)In(1 + \frac{{TD_m^2b}}{{{\lambda _l}s}})
\end{split}
\end{equation*}
where $\beta _{l, \cdot }^r$ is one solution of the offline expert, $s$ is the amplification factor of origin ${\lambda _l}$, ${D_m} = \mathop {\max }\limits_{1 \leqslant t \leqslant T} \{ ||{D_{l,t}}|{|_\infty }\} $, and ${Y_m} = \mathop {\max }\limits_{1 \leqslant t \leqslant T} \{ ||{Y_t}|{|_\infty },||{D_{l,t}}\beta _{l,t}^r|{|_\infty }\} $.
\hfill $\blacksquare$

\subsection{Theorem 5}
\begin{mytheorem}\textbf{\textit{5 }}
For the IOL process of the $l$-th online learner inside edRVFL-F working on batch stream, during $0 \leqslant t \leqslant T$, the upper bound of online-to-offline cumulative regrets between this learner and an offline expert can be written as:
\begin{align}
&\sum\limits_{t = 1}^T {{L_{l,t}}(\beta _{l,t}^f)}  - \mathop {min}\limits_{\beta _{l, \cdot }^f} (\frac{1}{2}{(\beta _{l, \cdot }^f)^T}{(\eta _{l,0}^f)^{ - 1}}\beta _{l, \cdot }^f + {L_{l,1..T}}(\beta _{l, \cdot }^f)) \hfill \nonumber\\
\leqslant& \frac{1}{2}Y_m^2b(N + k)(In(1 + \frac{{TD_m^2b}}{{{\lambda _l}s}}) - In(1 + \frac{{(T - 1)D_m^2b}}{{{\lambda _l}s + 2D_m^2b}})) \hfill \nonumber\\
\leqslant& \frac{1}{2}Y_m^2b(N + k)In(1 + \frac{{TD_m^2b}}{{{\lambda _l}s}}) \nonumber 
\end{align}
where $\beta _{l, \cdot }^f$ is one solution of the offline expert, $s$ is the amplification factor of origin ${\lambda _l}$, $In$ denotes natural logarithm, ${D_m} = \mathop {\max }\limits_{1 \leqslant t \leqslant T} \{ ||{D_{l,t}}|{|_\infty }\} $, and ${Y_m} = \mathop {\max }\limits_{1 \leqslant t \leqslant T} \{ ||{Y_t}|{|_\infty },||{D_{l,t}}\beta _{l,t}^f|{|_\infty }\} $. This suggests IOL with -F achieves \textit{lower regrets}, and its upper cumulative regret bound is at least 4 times better than that of IOL with -R.
\end{mytheorem}
\noindent{\textbf{\textit{Proof.}}}
To start IOL process without prior knowledge, $\beta _{l,0}^f = 0$ is retained as the same setup in Theorem \ref{theorem 2} and Theorem \ref{theorem 3}. So that it has:
\begin{equation} \label{32}
\beta _{l,t + 1}^f = \beta _{l,t}^f - \eta _{l,t}^f(D_{l,t + 1}^T{D_{l,t + 1}}\beta _{l,t + 1}^f - D_{l,t}^T{Y_t}).
\end{equation}

Note the data chunk comes at $t=1$ and ${\beta _{l,0}^f}={\beta _{l,1}^f}$ here according to the same logic in Lemma \ref{lemma 2.2}, so that we have ${\Delta _{U_{l,1}^f}}(\beta _{l, \cdot }^f,\beta _{l,1}^f) - {\hat L_{l,1}}(\beta _{l, \cdot }^f) - {\Delta _{U_{l,0}^f}}(\beta _{l, \cdot }^f,\beta _{l,0}^f) = 0$. Based on the Lemma \ref{lemma 3.3} and (\ref{18})-(\ref{20}), the right part of (\ref{31}) can be rewritten as: 
\begin{align}
&= \sum\limits_{t = 1}^T {({\Delta _{U_{l,t + 1}^f}}(\beta _{l,t}^f,\beta _{l,t + 1}^f)}  - {{\hat L}_{l,t + 1}}(\beta _{l,t}^f) + {{\hat L}_{l,t}}(\beta _{l,t}^f)) \nonumber\\
&- {\Delta _{U_{l,T + 1}^f}}(\beta _{l, \cdot }^f,\beta _{l,T + 1}^f) + {{\hat L}_{l,T + 1}}(\beta _{l, \cdot }^f) \hfill \nonumber\\
&= \sum\limits_{t = 1}^T {(\frac{1}{2}{{(\beta _{l,t}^f - \beta _{l,t + 1}^f)}^T}{{(\eta _{l,t + 1}^f)}^{ - 1}}(\beta _{l,t}^f - \beta _{l,t + 1}^f)} \nonumber\\
&- {\frac{1}{2}||{D_{l,t + 1}}\beta _{l,t}^f||_2^2 + \frac{1}{2}||{D_{l,t}}\beta _{l,t}^f||_2^2)}  \hfill \label{33}\\
&- \frac{1}{2}{(\beta _{l, \cdot }^f - \beta _{l,T + 1}^f)^T}{(\eta _{l,T + 1}^f)^{ - 1}}(\beta _{l, \cdot }^f - \beta _{l,T + 1}^f) \nonumber\\
&+ \frac{1}{2}||{D_{l,T + 1}}\beta _{l, \cdot }^f||_2^2 \hfill \nonumber 
\end{align}

Here we still focus on the difference of cumulative regrets
between online learners and the optimal offline expert. The expert model can maintain the same as the one in Theorem \ref{theorem 4} because the performance of the two IOL processes with -R/-F can be visually compared. Assume there always a solution set to the offline expert, based on Lemma \ref{lemma 2.1} and Lemma \ref{lemma 2.3again} and Theorem \ref{theorem 1}, let $\beta _{l, \cdot }^f = \beta _{l, * }^f$ (i.e. the optimal solution), the updated $\beta _{l,t + 1}^f$ in IOL process will converge to the optimal solution of offline expert in the end, namely $\beta _{l, * }^f = \beta _{l,T + 1}^f$. Use (\ref{21}) and (\ref{32}) to simplify (\ref{33}) while keeping the time synchronization of each element:
\begin{align}
&= \sum\limits_{t = 1}^T {(\frac{1}{2}{{(\beta _{l,t}^f - \beta _{l,t + 1}^f)}^T}{{(\eta _{l,t + 1}^f)}^{ - 1}}(\beta _{l,t}^f - \beta _{l,t + 1}^f)} \nonumber\\
&-{ \frac{1}{2}||{D_{l,t + 1}}\beta _{l,t}^f||_2^2 + \frac{1}{2}||{D_{l,t}}\beta _{l,t}^f||_2^2)}  \hfill \nonumber\\
&- \frac{1}{2}{(\beta _{l, \cdot }^f - \beta _{l,T + 1}^f)^T}{(\eta _{l,T + 1}^f)^{ - 1}}(\beta _{l, \cdot }^f - \beta _{l,T + 1}^f) \nonumber\\
&+ \frac{1}{2}||{D_{l,T + 1}}\beta _{l, \cdot }^f||_2^2 \hfill \nonumber\\
&= \sum\limits_{t = 1}^T {(\frac{1}{2}{{(D_{l,t}^T{Y_t})}^T}\eta _{l,t}^fD_{l,t}^T{Y_t}} \nonumber\\
&- \frac{1}{2}{{(D_{l,t + 1}^T{D_{l,t + 1}}\beta _{l,t + 1}^f)}^T}\eta _{l,t}^fD_{l,t + 1}^T{D_{l,t + 1}}\beta _{l,t + 1}^f \nonumber\\
&+{ \frac{1}{2}||{D_{l,t}}\beta _{l,t}^f||_2^2 - \frac{1}{2}||{D_{l,t + 1}}\beta _{l,t + 1}^f||_2^2)}  \hfill \label{34}\\
&- \frac{1}{2}{(\beta _{l, * }^f - \beta _{l,T + 1}^f)^T}{(\eta _{l,T + 1}^f)^{ - 1}}(\beta _{l, * }^f - \beta _{l,T + 1}^f) \nonumber\\
&+ \frac{1}{2}||{D_{l,T + 1}}\beta _{l, * }^f||_2^2 \hfill \nonumber\\
&= \sum\limits_{t = 1}^T {\frac{1}{2}{{(D_{l,t}^T{Y_t})}^T}\eta _{l,t}^fD_{l,t}^T{Y_t} }\nonumber\\
&-{ \sum\limits_{t = 1}^T {\frac{1}{2}{{(D_{l,t + 1}^T{D_{l,t + 1}}\beta _{l,t + 1}^f)}^T}\eta _{l,t}^fD_{l,t + 1}^T{D_{l,t + 1}}\beta _{l,t + 1}^f  } } \nonumber\\
&-\frac{1}{2}||{D_{l,T + 1}}\beta _{l,T + 1}^f||_2^2 \hfill \nonumber\\
&- \frac{1}{2}{(\beta _{l, * }^f - \beta _{l,T + 1}^f)^T}{(\eta _{l,T + 1}^f)^{ - 1}}(\beta _{l, * }^f - \beta _{l,T + 1}^f) \nonumber\\
&+ \frac{1}{2}||{D_{l,T + 1}}\beta _{l, * }^f||_2^2 \hfill \nonumber
\end{align}

Compares online learner with the optimal offline expert as it can give upper bounds for all offline experts. 
Theoretically, $||{D_{l,T + 1}}\beta _{l, * }^f|{|_2} \leqslant ||{D_{l,T + 1}}\beta _{l,T + 1}^f|{|_2}$ holds as the $\beta _{l, * }^f$ is the optimum of offline expert on global data. Remove the negative terms and (\ref{34}) can be restricted as follows:
\begin{equation}\label{35}
\begin{aligned}
&\leqslant \sum\limits_{t = 1}^T {\frac{1}{2}{{(D_{l,t}^T{Y_t})}^T}\eta _{l,t}^fD_{l,t}^T{Y_t}} \\
&- {\sum\limits_{t = 1}^T {\frac{1}{2}{{(D_{l,t + 1}^T{D_{l,t + 1}}\beta _{l,t + 1}^f)}^T}\eta _{l,t}^fD_{l,t + 1}^T{D_{l,t + 1}}\beta _{l,t + 1}^f} } 
\end{aligned}
\end{equation}

Based on the Matrix Spectral Theorem, the first summation term in (\ref{35}) can be bounded to:
\begin{equation}\label{36}
\begin{aligned}
&\phantom{{}= \hspace{0mm}}\sum\limits_{t = 1}^T {\frac{1}{2}{{(D_{l,t}^T{Y_t})}^T}\eta _{l,t}^fD_{l,t}^T{Y_t}}  \\
&= \sum\limits_{t = 1}^T {\frac{1}{2}||D_{l,t}^T{Y_t}||_2^2\frac{{Y_t^T{D_{l,t}}}}{{||D_{l,t}^T{Y_t}|{|_2}}}\eta _{l,t}^f\frac{{D_{l,t}^T{Y_t}}}{{||D_{l,t}^T{Y_t}|{|_2}}}}  \hfill \\
&\leqslant  \frac{1}{2}\sum\limits_{i,j} {{{[({D_{l,t}}D_{l,t}^T) \odot ({Y_t}Y_t^T)]}_{i,j}}} \sum\limits_{t = 1}^T {\frac{1}{{{e_{min}}({{(\eta _{l,t}^f)}^{ - 1}})}}} \\
&\leqslant \frac{1}{2}{({Y_m}{D_m}b)^2}(N + k)\sum\limits_{t = 1}^T {\frac{1}{{{e_{min}}({{(\eta _{l,t}^f)}^{ - 1}})}}}  \\
&\leqslant \frac{1}{2}{({Y_m}{D_m}b)^2}(N + k)\int_0^T {\frac{1}{{{\lambda _l} + D_m^2bt}}dt} \\
&\leqslant \frac{1}{2}{({Y_m}{D_m}b)^2}(N + k)\frac{1}{{D_m^2b}}In(1 + \frac{{TD_m^2b}}{{{\lambda _l}}}) \hfill \\
&= \frac{1}{2}Y_m^2b(N + k)In(1 + \frac{{TD_m^2b}}{{{\lambda _l}}}) \hfill \\ 
\end{aligned}
\end{equation}
where $(N + k)$ is dimensions of $\eta _{l,t}^f$, ${D_m} = \mathop {\max }\limits_{1 \leqslant t \leqslant T} \{ ||{D_{l,t}}|{|_\infty }\} $, and ${Y_m} = \mathop {\max }\limits_{1 \leqslant t \leqslant T} \{ ||{Y_t}|{|_\infty },||{D_{l,t}}\beta _{l,t}^f|{|_\infty }\} $ here.

The second summation term in (\ref{35}) can be bounded to:
\begin{align}
&\phantom{{}= \hspace{0mm}}\sum\limits_{t = 1}^T {\frac{1}{2}{{(D_{l,t + 1}^T{D_{l,t + 1}}\beta _{l,t + 1}^f)}^T}\eta _{l,t}^fD_{l,t + 1}^T{D_{l,t + 1}}\beta _{l,t + 1}^f}  \hfill \nonumber\\
&\leqslant \sum\limits_{t = 1}^T {\frac{1}{2}{{(D_{l,t + 1}^T{D_{l,t + 1}}\beta _{l,t + 1}^f)}^T}\eta _{l,t + 1}^fD_{l,t + 1}^T{D_{l,t + 1}}\beta _{l,t + 1}^f}  \hfill \nonumber\\
&\leqslant \frac{1}{2}{({Y_m}{D_m}b)^2}(N + k)\sum\limits_{t = 1}^T {\frac{1}{{{e_{min}}({{(\eta _{l,t + 1}^f)}^{ - 1}})}}}  \hfill \label{37}\\
&\leqslant \frac{1}{2}{({Y_m}{D_m}b)^2}(N + k)\int_1^T {\frac{1}{{{\lambda _l} + D_m^2b(t + 1)}}} dt \hfill \nonumber\\
&= \frac{1}{2}Y_m^2b(N + k)In(1 + \frac{{(T - 1)D_m^2b}}{{{\lambda _l} + 2D_m^2b}}) \hfill \nonumber 
\end{align}
Likewise ${\lambda _l}$ should be fine-tuned when data volume changes. For consistency and fair comparison, here ${\lambda _l}$ and $s$ remain as (\ref{30}). To summarize above, a more general theorem on cumulative regret bound of IOL process with forward regularization is given based on (\ref{33}):
\begin{align}
&\phantom{{}= \hspace{0mm}}\sum\limits_{t = 1}^T {{L_{l,t}}(\beta _{l,t}^f)}  - \mathop {min}\limits_{\beta _{l, \cdot }^f} (\frac{1}{2}{(\beta _{l, \cdot }^f)^T}{(\eta _{l,0}^f)^{ - 1}}\beta _{l, \cdot }^f + {L_{l,1..T}}(\beta _{l, \cdot }^f)) \hfill \nonumber\\
&\leqslant \frac{1}{2}Y_m^2b(N + k)(In(1 + \frac{{TD_m^2b}}{{{\lambda _l}s}}) \label{38}\\
&- In(1 + \frac{{(T - 1)D_m^2b}}{{{\lambda _l}s + 2D_m^2b}})) \hfill \nonumber 
\end{align}
where $\beta _{l, \cdot }^f$ is one solution of the offline expert, $s$ is the amplification factor of origin ${\lambda _l}$, $In$ denotes natural logarithm, ${D_m} = \mathop {\max }\limits_{1 \leqslant t \leqslant T} \{ ||{D_{l,t}}|{|_\infty }\} $, and ${Y_m} = \mathop {\max }\limits_{1 \leqslant t \leqslant T} \{ ||{Y_t}|{|_\infty },||{D_{l,t}}\beta _{l,t}^f|{|_\infty }\} $.
\hfill$\blacksquare$

\subsection{Numerical simulations on -R/-F within IOL}
This experimental part mainly focused on numerical simulations of IOL processes of sub-learner-R/-F. It was based on Section. IV theorems about regrets and included detailed performance comparisons of IOL processes on single and batch dynamic streams respectively, using -R and -F algorithms with variable regularization factors for sustaining model learning. It was also expected to explain the advantages of using -F within IOL framework under rigorous numerical situations.

Here followed the previous setup, and the edRVFL network working on online streams was investigated. For $0 \leqslant t \leqslant T$, trainable model of the $l$-th layer (sub-learner) inside edRVFL could be denoted as ${D_{l,t}} \cdot \beta _{l,t}^{\{ r,f\} } \to {Y_t}$ projection, where the $\beta _{l,t }^{\{ r,f\} }$ was deduced by Algorithm \ref{alg1} or Algorithm \ref{alg2} for ridge or forward learning respectively. For the sake of simplicity, following analysis and experiments were based on one sub-learner inside of edRVFL. It could be proved that the value of ${D_{l,t}}$ also obeyed normal distribution given real-valued ${X_t}$. Without loss of generality, randomized weights were generated by classical normal distribution to simulate the adversarial scenarios as the results could be spread to popular Kaiming or Xavier settings. This allowed us to examine learning performance under the possible conditions of extreme values being mixed. We still used $(D,Y) = \{ ({D_t},{Y_t})\} _{t = 1}^T = \{ \{ (d_t^i,y_t^i)\} _{i = 1}^b\} _{t = 1}^T$ to denote the online stream, and only one sample (resp. batch) at $t$ time was picked for single (resp. batch) data scenarios. The $Y = D \cdot {\beta ^o} + \varepsilon $ was used to generate targets for regression simulations where ${\beta ^o}$ signified oracle weights and $\varepsilon $ was sub-Gaussian noise. Other parameters included: $T=1000$, $b=10$, $d_t^i \sim N(0,I)$, ${\beta ^o} \sim N(5,1)$ for single and ${\beta ^o} \sim N(20,1)$ for batch stream, noise factor $= 0.15$, number of repetitions ${n_t}=200$, regularization factor $\lambda =[0.005, 0.01, 0.02, 1/T  ]$.

Some metrics were used to evaluate the algorithm performance in IOL:
Immediate regrets on single stream: $IR_{single}^{\{ r,f\} } = 2 \cdot ({\ell _t}(\beta _t^{\{ r,f\} }) - {\ell _t}(\beta _{}^o))$;
Immediate regrets on batch stream: $IR_{batch}^{\{ r,f\} } = 2 \cdot ({L_t}(\beta _t^{\{ r,f\} }) - {L_t}(\beta _{}^o))$.
They used the $\bar \gamma $ form in remarks to compute synchronous immediate regrets between online learners of IOL processes and oracles that could also be viewed as an offline oracle expert, and the averaged values in multiple trials were denoted by $\overline {IR} _{single}^{\{ r,f\} } = \frac{1}{{{n_t}}}\sum\nolimits_i {IR_{single}^{\{ r,f\} }} $ and $\overline {IR} _{batch}^{\{ r,f\} }$ respectively.
Immediate regret terms for single stream, namely $IRT_{single}^{\{ r,f\} }$, were defined to the values of the left part of equations in Lemma \ref{lemma 2.4} at time $t$, which measured the synchronous immediate regrets between online learners of IOL processes and offline experts. Similar definitions for $IRT_{batch}^{\{ r,f\} }$ were based on Theorem \ref{theorem 4} - Theorem \ref{theorem 5}: $IRT_{batch}^r = \underbrace {{{({D_t}\beta _t^r - {Y_t})}^T}{D_t}\eta _t^rD_t^T({D_t}\beta _t^r - {Y_t})}_{1^{st} \text{-term of ridge}}$,
$IRT_{batch}^f = \underbrace {{{(D_t^T{Y_t})}^T}\eta _t^fD_t^T{Y_t}}_{1^{st} \text{-term of forward}} - \underbrace {{{(D_{t + 1}^T{D_{t + 1}}\beta _{t + 1}^f)}^T}\eta _t^fD_{t + 1}^T{D_{t + 1}}\beta _{t + 1}^f}_{2^{nd} \text{-term of forward}}$.
Their averaged forms were denoted by $\overline {IRT} _{single}^{\{ r,f\} }$ and $\overline {IRT} _{batch}^{\{ r,f\} }$ respectively. Note that $IR{T^r}$ had only one term while $IR{T^f}$ was composed of two terms. It was necessary to investigate the decomposition of these error terms in detail in the experiment. Cumulative regrets assessed the total regrets between the online learners of IOL processes and the optimal offline experts (using ${\beta _ * }$) defined in our proposed Theorem \ref{theorem 4} - Theorem \ref{theorem 5}, and were denoted by $CR_{batch}^{\{ r,f\} }$ here.

\begin{figure*}[htbp]
\centering
\subfigure[]{\includegraphics[width=0.4\textwidth]{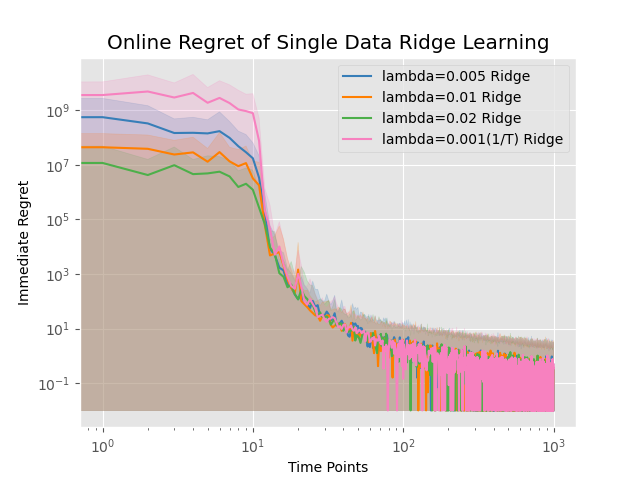}}
  \hfil
\subfigure[]{\includegraphics[width=0.4\textwidth]{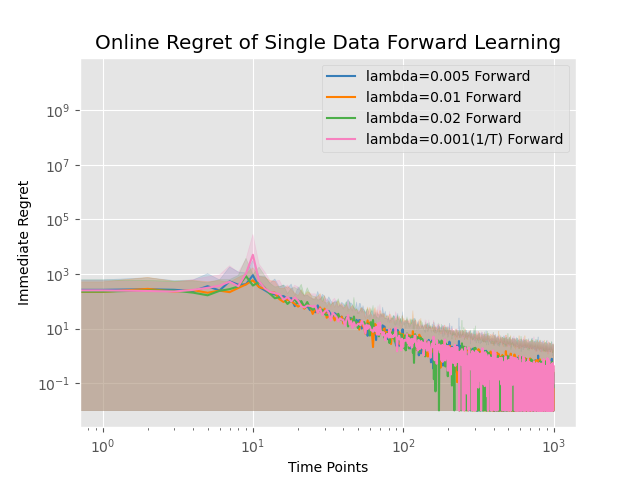}}
  \hfil
\\
\vspace{-1em}
\subfigure[]{\includegraphics[width=0.4\textwidth]{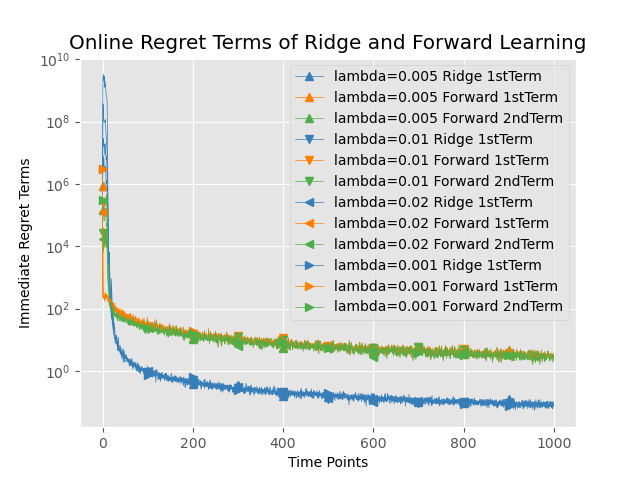}}
  \hfil
\subfigure[]{\includegraphics[width=0.4\textwidth]{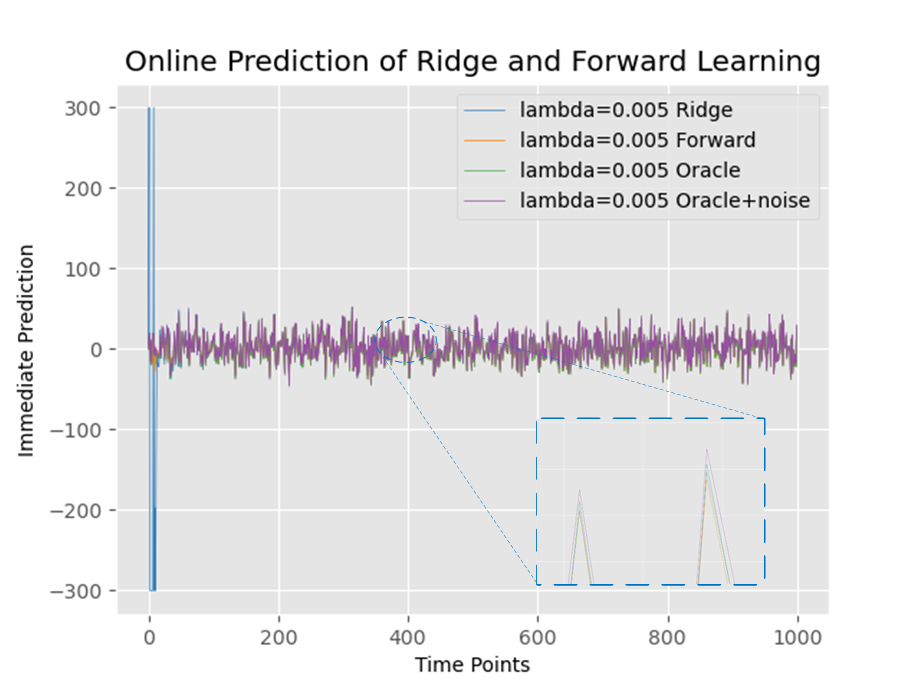}}
  \hfil

\centering
\caption{Comparisons of IOL processes using -R and -F on online single streams. (a) $\overline{I R}_{single}^r$: relative immediate regrets of ridge learner w.r.t. oracle expert on single stream. (b) $\overline{I R}_{single}^f$: relative immediate regrets of forward learner w.r.t. oracle expert on single stream. (c) $\overline {IRT} _{single}^{\{ r,f\} }$: respective immediate regret terms of online learners with -R and -F w.r.t. optimal offline expert (learned on global) on single stream. (d) prediction values of online learners with -R and -F.}
\label{fig 3.1}
\end{figure*}

\begin{figure*}[htbp]
\centering
\subfigure[]{\includegraphics[width=0.4\textwidth]{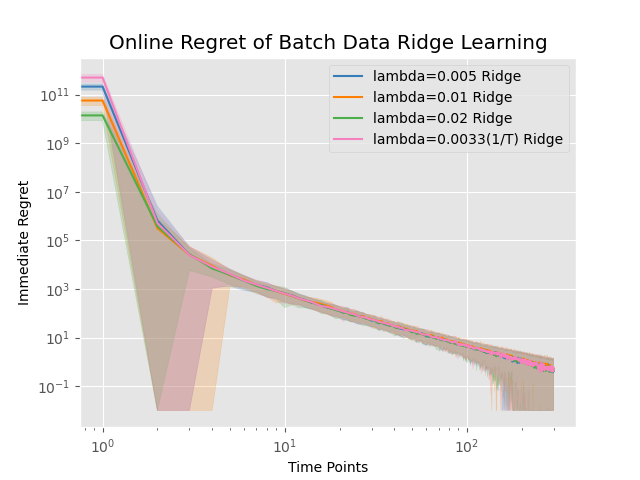}}
  \hfil
\subfigure[]{\includegraphics[width=0.4\textwidth]{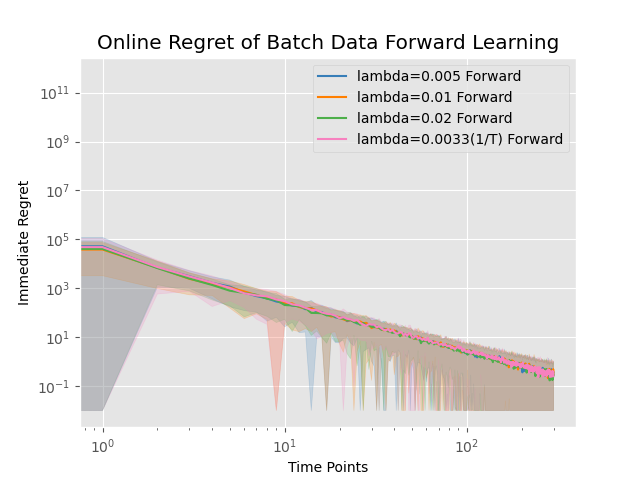}}
  \hfil
  \\
  \vspace{-1em}
\subfigure[]{\includegraphics[width=0.4\textwidth]{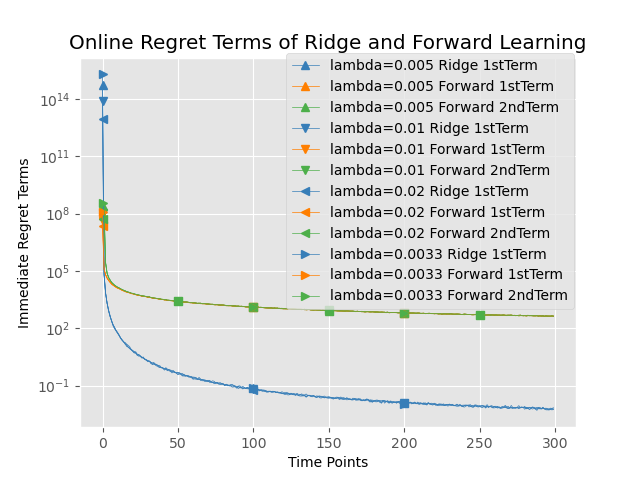}}
  \hfil
\subfigure[]{\includegraphics[width=0.4\textwidth]{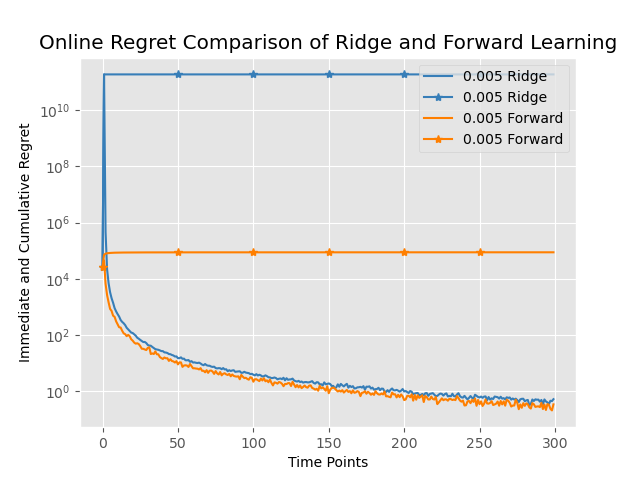}}
  \hfil
\centering
\caption{Comparisons of IOL processes using -R and -F on online batch streams. (a) $\overline{I R}_{batch}^r$: relative immediate regrets of ridge learner w.r.t. oracle expert on batch stream. (b) $\overline{I R}_{batch}^f$: relative immediate regrets of forward learner w.r.t. oracle expert on batch stream. (c) $\overline {IRT} _{batch}^{\{ r,f\} }$: respective immediate regret terms of online learners with -R and -F w.r.t. optimal offline expert on batch stream. (d) $\overline {IR} _{batch}^{\{ r,f\} }$: regrets of online learners with -R and -F on batch stream and $\overline{C R}_{batch}^{\{ r,f\} }$: respective cumulative regrets.}
\label{fig 3.2}
\end{figure*}

Results of IOL processes on single stream are shown in Fig. \ref{fig 3.1}(a)-(b). Contrasted to $\overline {IR} _{single}^r$ by given $\lambda $, $\overline {IR} _{single}^f$ shows a faster rate of convergence, smaller regret loss to expert with oracle parameter, and lower standard deviation (oscillation) in multi-trials. This suggests that using forward benefits: (1) faster approximation to oracle and lower predictive loss; (2) less cumulative regrets and more robustness in IOL process. Unlike the $\overline {IRT} _{single}^r$, $\overline {IRT} _{single}^f$ is calculated by the difference of two terms which faster approaches and cancels each other out in Fig. \ref{fig 3.1}(c). That difference of $\overline {IRT} _{single}^f$ declines faster than the $\overline {IRT} _{single}^r$, which shows lower immediate and also cumulative regret loss to the optimal offline expert by using -F. Please note the oracle expert and optimal offline expert are different. In fact, the $\overline {IR} _{single}^f$ is lower in Fig. \ref{fig 3.1}(a)-(b) and $||\beta _{T + 1}^r - {\beta ^o}||_2^2=0.205>||\beta _{T + 1}^f - {\beta ^o}||_2^2=0.196$ when $\lambda =0.005$, and in Fig. \ref{fig 3.1}(d) the immediate prediction by -F is closer to original results and has stronger anti-noise ability compared to -R. However, these effects may not be obvious in single stream scenarios. In the following tests of batch stream, we will show the advantages of -F still hold and can be observed clearly, and verify some conclusions based on the theorems of Section III.

Results of IOL processes with -R/-F on batch data stream are shown in Fig. \ref{fig 3.2}(a)-(b). Comparisons between $\overline {IR} _{batch}^r$ and $\overline {IR} _{batch}^f$ are more distinct. Conclusions are summarized as: (1)  $\overline {IR} _{batch}^f$ attenuates smoothly and faster arrives with lower regret loss in the end compared to $\overline {IR} _{batch}^r$. Here $||\beta _{T + 1}^r - {\beta ^o}||_2^2 = 0.521 > ||\beta _{T + 1}^f - {\beta ^o}||_2^2 = 0.430$ given $\lambda =0.005$. It means using -F makes the IOL process more robust and approaches closely to oracle parameters, and regrets of -R grow faster than -F's; (2) for the immediate regrets between learners of IOL processes and optimal offline expert, in Fig. \ref{fig 3.2}(c), the superiority of -F is maintained because the two regret terms of $\overline {IRT} _{batch}^f$ intersect rapidly with smaller loss to optimal offline expert while the regret term of $\overline {IRT} _{batch}^r$ descends slowly. To clearly show that -F gives lower regrets than -R, $\overline {IR} _{batch}^{\{ r,f\} }$ at $\lambda =0.005$ is drawn in Fig. \ref{fig 3.2}(d), and cumulative regret $\overline {CR} _{batch}^{\{ r,f\} }$ is also displayed by $\rlap{---} \Delta $. Even if the gap of immediate regrets of -R and -F sharply descends from the start, regrets accumulated during the processes vary much in Fig. \ref{fig 3.2}(d). Results also validate some corollaries and remarks in Section. IV, such as better regret bound of using -F during IOL, -F achieves slower growth speed on cumulative regret bound compared to -R during IOL. Furthermore, by comparing the regret curves of single and batch scenarios, we can find that batch learning scenario not only keeps the above advantages of using -F over -R, but also accelerates learning process, resists interference, and reduces vibrations of non-determinacy in IOL. Based on above analysis, it is suggested to substitute -R with -F in IOL frameworks.

During IOL, the learners with -R/-F successfully learned knowledge on non-stationary online streams. Experimental processes were implemented according to Algorithm \ref{alg1} and Algorithm \ref{alg2}, so there was \textit{no retrospective retraining} of old data. The immediate regret variations observed in Fig. \ref{fig 3.1}(a)-(c) and Fig. \ref{fig 3.2}(a)-(d) both smoothly decrease, indicating that the learners gradually improved performance, and there is no obvious \textit{catastrophic forgetting} and has good suppression of \textit{distribution drift}.

\end{document}